\documentclass[12pt]{article}

\usepackage{amsmath,amsfonts,amsthm,mathrsfs,xcolor,bm,bbm, mathtools, mathdots}
\usepackage{amssymb}
\usepackage{verbatim}
\usepackage{graphicx}
\usepackage{setspace}    
\usepackage{natbib}
\usepackage{fancyhdr}
\usepackage{comment}
\usepackage{soul}
\usepackage{cancel}
\usepackage{array}
\usepackage{tabularx}
\usepackage{multirow}
\usepackage[most]{tcolorbox}
\usepackage{makecell}
\usepackage[normalem]{ulem}
\usepackage{hyperref} 
\usepackage[left=1in, right=1in, top=1in, bottom=1in]{geometry}
\usepackage[title]{appendix}
\usepackage{authblk}

\newcommand{\highlight}[1]{{\color{black}#1}}
\newcommand{\second}[1]{{\color{black}#1}}

\newcommand{\BlackBox}{\rule{1.5ex}{1.5ex}}  
\ifdefined\proof
    \renewenvironment{proof}{\par\noindent{\bf Proof\ }}{\hfill\BlackBox\\[2mm]}
\else
    \newenvironment{proof}{\par\noindent{\bf Proof\ }}{\hfill\BlackBox\\[2mm]}
\fi

\def\NN{\mathbb N}
\def\ZZ{\mathbb Z}
\def\RR{\mathbb R}

\numberwithin{equation}{section}

\newtheorem{theorem}{Theorem}[section]
\newtheorem{lemma}[theorem]{Lemma}
\newtheorem{proposition}[theorem]{Proposition}
\newtheorem{corollary}[theorem]{Corollary}
\newtheorem{definition}[theorem]{Definition}
\newtheorem{remark}[theorem]{Remark}

\newtheorem{assumption}[theorem]{Assumption}
\newtheorem*{lemma*}{Lemma}

\title{Neural Operators for Nonlinear Functionals on RKHS}
\author[1]{Tian-Yi Zhou}
\author[2]{Namjoon Suh}
\author[2]{Guang Cheng}
\author[3]{Xiaoming Huo}
\affil[1]{Data Science Institute, Columbia University in the City of New York}
\affil[2]{Department of Statistics and Data Science,  University of California, Los Angeles}
\affil[3]{H. Milton Stewart School of Industrial and Systems Engineering, Georgia Institute of Technology}

\date{\vspace{-5ex}}

\begin{document}

\maketitle

\begin{abstract}
 \noindent  
Motivated by the abundance of functional data, such as time series and images, we study the approximation and statistical learning of nonlinear functionals defined on reproducing kernel Hilbert spaces (RKHSs) using neural networks. 
By leveraging interpolating orthogonal projections in RKHSs, we use finitely many point evaluations in place of integration-based basis function expansions. This leads to a simpler and more flexible neural-network architecture that remains applicable even when the data domain or the underlying kernel is not explicitly known.
We establish universal approximation results and derive explicit kernel-dependent approximation rates and parameter-complexity bounds for RKHSs induced by inverse multiquadric, Gaussian, and Sobolev kernels. 
We also apply our results to the regression maps arising in generalized functional linear models.
Finally, we analyze the generalization properties of the resulting neural-network classes and establish finite-sample guarantees for learning nonlinear functionals.
\end{abstract}

 \section{Introduction}
 This paper studies the approximation and statistical learning of nonlinear functionals defined on reproducing kernel Hilbert spaces (RKHSs) using neural networks. A functional maps an infinite-dimensional function space to $\RR$ and can therefore be viewed as a special type of operator. We study two related questions: how efficiently neural networks can approximate such functionals and how accurately they can be learned from finitely many observations.

 Neural networks have been known as universal approximators since \citep{cybenko1989approximation}, i.e., to approximate any continuous function, mapping a finite-dimensional input space into
another finite-dimensional output space, to an arbitrary accuracy. 
These days, many interesting tasks entail learning operators, i.e., mappings between an infinite-dimensional input Banach space and (possibly) an infinite-dimensional output space.
A prototypical example in scientific computing is to map the initial datum into the (time series of) solution of a nonlinear time-dependent partial differential equation (PDE).
A priori, it is unclear whether neural networks can approximate such operators efficiently and learn them accurately from finite data.

One of the first successful uses of neural networks in the context of operator learning was provided by \citep{chen1995universal}. A shallow neural network-based architecture, termed as \textit{neural operator}, is proposed and is proved to possess an universal approximation property for nonlinear operators from an infinite-dimensional space to a finite-dimensional Euclidean space.
Recently, there has been an active line of research developing new methods for learning operators by deep neural networks, such as the \textit{DeepONet} \citep{lu2021learning}, \textit{physics-informed neural networks} (PINNs) \citep{raissi2019physics, wang2021learning}, \textit{Fourier Neural Operator} (FNO) \citep{li2020fourier}, and \textit{DeepGreen} \citep{gin2021deepgreen}.
Although these methods have shown success through numerical experiments in approximating operators in many interesting examples, theoretical guarantees for approximation and finite-sample learning in this context remain limited. 
Some recent quantitative approximation results are obtained in \citep{kovachki2021universal, lanthaler2022error, liu2022deep}. 

This work focuses on the approximation and statistical learning of nonlinear functionals defined on RKHSs. A major advantage of the RKHS structure is that it allows an input function to be represented from its values at finitely many sampling points, rather than through integration-based basis expansions. This reduces the original infinite-dimensional problem to a finite-dimensional one while allowing us to quantify the error introduced by this reduction.

We consider neural networks equipped with \second{either the ReLU or the tanh activation function}, both of which play important roles in operator learning. Much of the recent theoretical literature on neural network approximation has focused on ReLU networks. In contrast, smooth activation functions such as tanh are often preferred in physics-informed neural networks (PINNs) and related operator-learning methods for solving forward and inverse problems governed by partial differential equations, since their smoothness is advantageous when evaluating higher-order derivatives through automatic differentiation; see \citep{raissi2019physics} and the references therein. Thus, studying both ReLU and tanh networks enables us to establish results for two activation functions of substantial theoretical and practical relevance in operator learning.

To provide theoretical guarantees for learning nonlinear functionals from finite data, approximation theory alone is not sufficient; the generalization error must also be controlled. We first establish explicit bounds showing that neural networks can approximate functionals on RKHSs to any prescribed accuracy. \second{We further develop a generalization analysis for our proposed functional-learning architecture, providing finite-sample guarantees for learning nonlinear functionals from data.}

\subsection{Related Works}\label{sec:relatedwork}
\subsubsection{Classic Funtional MLP}

In an earlier work \citep{mhaskar1997neural}, a shallow neural network with infinitely differentiable activation function is used to approximate nonlinear, continuous functionals on $L^p$ spaces, for $1 \leq p \leq \infty$. 
Though the approximation rate given there does not seem satisfactory --- when measuring the approximation error on the unit ball of a H\"{o}lder space, the number of network parameters grows exponentially
with a large exponent as the accuracy increases --- it is proven that such a rate is, in fact, optimal.
Later in \citep{stinchcombe1999neural}, a generalized neural network is introduced to approximate functionals whose domain is a locally convex topological vector space.   
Based on the aforementioned work, a novel functional multi-layer perception (MLP) is introduced by \citep{rossi2005functional}, which can be directly applied to functional data. Recently in \citep{thind2023deep}, an optimization algorithm is proposed to implement the functional MLP in practice.
The functional MLP is based on, but different from, a classical fully-connected neural network. 

To illustrate this special network's design, we first formulate a standard fully-connected network. 
For $\mathbf b= (b_1,\ldots, b_r)\in \RR^r$, let $\sigma_{\mathbf b}: \RR^r \rightarrow \RR^r$ be the shifted activation function as 
\begin{equation*}
    \sigma_{\mathbf b}\begin{pmatrix}
    x_1\\
    \vdots\\
    x_r
    \end{pmatrix}:= \begin{pmatrix}
    \sigma (x_1-b_1)\\
    \vdots\\
    \sigma (x_r-b_r)
    \end{pmatrix},
\end{equation*}
where $\sigma: \RR \to \RR$ is an activation function such as tanh: $\sigma(x) = \frac{e^x- e^{-x}}{e^x+e^{-x}}$ and ReLU: $\sigma(x) = \max \{x,0\}$. 
We use the vector $\mathbf d= (d_1,\ldots, d_\ell)\in \NN^\ell$ to indicate the width in each hidden layer.  
\begin{definition}\label{def:classic}
A fully-connected neural network $\widehat{G} : \RR^{N} \rightarrow \RR$ with $L$ hidden layers is a function that takes the form
\begin{equation*}
    \widehat{G}(x) = a^T\sigma_{\mathbf b_L}(W_L \sigma_{\mathbf b_{L-1}}(W_{L-1}\cdots  \sigma_{\mathbf b_1}(W_1 x))),
\end{equation*}
where $x \in \RR^{N}$ is the input, $a\in \RR^{d_L}$ is the outer weight, $d_0 = N$, $W_\ell \in \RR^{d_\ell \times d_{\ell-1}}$ denote the weight matrices, $b_\ell \in \RR^{d_\ell}$ denote the bias vectors over the layers $\ell =1,\ldots, L,$ in the network.
\end{definition}

The functional MLP introduced in \citep{rossi2005functional} is designed to approximate functionals defined on $L^p[-1,1]^s$ for $p\geq 1$ and $s \geq 1$, where the network input is a function covariate $f: L^p[-1,1]^s \rightarrow \RR$. We give the formulation of functional MLP below. 
\begin{definition}[Functional MLP]
A functional MLP $\Theta:L^p[-1,1]^s \rightarrow \RR$ with $L$ hidden layers is any functional that takes the form
\begin{equation*}
    \Theta(f) = a^T\sigma_{\mathbf b_L}W_L \sigma_{\mathbf b_{L-1}}W_{L-1}\cdots  \sigma_{\mathbf b_2}W_2 T(f),
\end{equation*}
where $f\in L^p[-1,1]^s$, $T:L^p[-1,1]^s \rightarrow \RR^{d_1}$ is a bounded linear operator with 
\begin{equation}\label{firstlayer}
    T(f) = \left(\int_{[-1,1]^s}f(x)g_1(x)dx, \cdots, \int_{[-1,1]^s}f(x)g_{d_1}(x)dx\right)^T,
\end{equation}
where $\{g_k\}_{k=1}^{d_1}\subset L^q[-1,1]^s$ is a set of pre-selected basis functions and  $\frac{1}{p}+ \frac{1}{q} = 1$.
\end{definition}
As we can see in (\ref{firstlayer}), 
the initial layer of this network converts the function $f$ into a vector form that can be fed directly into a standard neural network via basis function expansion. 
This is referred to as the \textit{discretization step}. 
One drawback of such a design is that a prior selection of basis functions is required, often without information about the task at hand. 
A universal approximation theorem of functional MLP is established in \citep{rossi2005functional},
where a scheme to learn $\{g_k\}_{k=1}^{d_1}$ is introduced by assuming them in a parametric function space.

Recently, two variants of functional MLP have been introduced in \citep{yao2021deep} and \citep{song2023approximationarxiv}.
In \citep{yao2021deep}, each basis function is parameterized with a micro neural network embedded in the MLP; 
while in \citep{song2023approximationarxiv}, the basis functions are specifically chosen to be some families of orthogonal polynomials.  
For approximating functionals on $L^p$ spaces ($p\geq 1$), the network proposed in \citep{song2023approximationarxiv} achieves a similar approximation rate as in \citep{mhaskar1997neural}. 

\second{More recently, \citep{shi2025nonlinear} proposed a kernel-structured variant of  functional MLP involving two kernel choices: an embedding kernel and a projection Mercer kernel. Let \(\mathcal X \subset\mathbb R^d\) denote the domain of the input functions. Instead of directly representing the input $f$ through a basis expansion as in \eqref{firstlayer}, their construction first maps \(f\) through a kernel integral transformation
$$
    L_Kf(x)=\int_{\mathcal X}K(x,t)f(t)\,dt,
$$
where \(K:\mathcal X \times \mathcal X \to \RR \) is an embedding kernel. The embedded function \(L_Kf\) is then projected onto the first \(d_1\) eigenfunctions \(\{\phi_j\}_{j=1}^{d_1}\) of the integral operator associated with a projection Mercer kernel \(K_0\) and a Borel measure $\mu$ on $\mathcal X$, yielding
$$
    T(L_Kf)
    =
    \left(
    \langle L_Kf,\phi_1\rangle_{L^2(\mu)},
    \ldots,
    \langle L_Kf,\phi_{d_1}\rangle_{L^2(\mu)}
    \right)^T.
$$
This finite-dimensional representation is subsequently fed into a standard ReLU network. Thus, their proposed method is similar to the classic functional MLP, except that the initial functional-to-vector transformation is carried out through a kernel embedding followed by spectral projection.}

At this point, a question arises: can we adopt a simpler network to approximate nonlinear functionals without integrations in the first layer?
All the above-mentioned functional learning methods are based on a set of basis functions defined on a domain in a Euclidean space. 
In this paper, we study the approximation of functionals on RKHS, and the discretization of functions can be simply done by point evaluations.
In other words, we adopt the standard fully-connected neural networks, with the input vector given by the discretely observed functional values. 
    Although each RKHS is uniquely induced by a specific reproducing kernel, the explicit form of the underlying \second{kernel section $K(\cdot, x)$} is often unknown in practice.
    This lack of accessibility to the \second{kernel section} poses a major limitation for integration-based discretization methods. In contrast, our approach leverages the reproducing property of RKHS directly through point evaluations, which can be computed without knowledge of the underlying kernel.

Our point-evaluation-based method is also consistent with the design philosophy of modern neural approaches for learning functions and operators, such as \textit{DeepONet}, PINNs, and \textit{DeepGreen}, which operate directly on discretized function values rather than requiring a basis expansion of the input function.

\subsubsection{Neural Operators}
The term \textit{neural operator} was first formalized  by \citep{novak2018reproducing} to describe a deep learning framework designed to learn operators. It generalizes neural networks to work directly on infinite-dimensional inputs and outputs. In \citep{novak2018reproducing}, a general theory was established, showing that functions in an infinite-dimensional Banach space can be discretized by evaluating them at finitely many points.  However, the distinct geometry of RKHSs introduces mathematical challenges not fully addressed by this general theory, making the problem substantially different from approximation in general Banach spaces.

The DeepONet architecture, introduced by \citep{lu2021learning}, is among the first neural operators, extending the shallow architecture proposed in \citep{chen1995universal} by incorporating an iterated, deep structure.
More recently, \citep{lanthaler2022error} provided an error estimate for DeepONet in the $L^2$ norm with respect to the Lebesgue measure. We note that for any continuous function $f\in C(\mathcal{X})$ and $1 \leq p <\infty$, 
\begin{eqnarray}
    \|f\|_\infty = \sup\left\{\|f\|_{L^p_{\rho_\mathcal{X}}} =\left(\int_{\mathcal{X}}|f(x)|^p \ d\rho_\mathcal{X}\right)^{\frac{1}{p}}:
  \rho_\mathcal{X} 
    \text{ is \second{a Borel probability} measure  on $\mathcal{X}$}\right\}.
\end{eqnarray}
In other words, the traditional $L^p$ norms are  relative to the Lebesgue measure of the Euclidean domain only, whereas the uniform norm maximizes over all possible measures. In contrast to \citep{lanthaler2022error}, our analysis focuses on uniform error bounds, which are useful for further studying other learning problems such as the classification of functional data.

Notably, the analysis of \citep{lanthaler2022error} relies on the assumption that the eigenbasis of the covariance operator is uniformly bounded, a restrictive condition verified only for periodic kernels, finite-dimensional RKHSs, and certain special kernels with the uniform measure on a Euclidean domain. Our results avoid this assumption and thus apply more broadly. We require neither a uniform marginal distribution nor a bounded density with respect to the uniform measure.

Finally, the bounds in \citep{lanthaler2022error} depend on unspecified eigenvalues of the covariance operator and therefore do not yield explicit rates in terms of network size, training sample size, and data dimension. Our results provide explicit rates in all three quantities.

\subsection{Problem Formulation}
Let $K$ be a  Mercer kernel on a compact metric space $(\mathcal{X}, d_\mathcal{X})$, which induces a unique RKHS \citep{aronszajn1950theory} defined as:
\begin{definition}[RKHS]
The RKHS $\mathcal{H}_K$ is the closure of the set of functions \\$\left\{\sum_{j=1}^p \alpha_j K(\cdot, x_j): p\in \NN, \alpha_j \in \RR, x_j \in \mathcal{X}\right\}$ under the metric induced by the inner product $ \langle \cdot, \cdot \rangle_{\mathcal{H}_K}$ 
 given by $$\left\langle \sum_{i=1}^q \beta_i K(\cdot, y_i), \sum_{j=1}^p \alpha_j K(\cdot, x_j) \right\rangle_{\mathcal{H}_K}  = \sum_{i=1}^q \sum_{j=1}^p \beta_i \alpha_j K(y_i, x_j).$$
\end{definition}
The RKHS  $\mathcal{H}_K$ is characterized by the property that the point evaluation functional $L_x$ on $\mathcal{H}_K$ given by $f \mapsto f(x)$ is continuous for every $ x \in \mathcal{X}$.

Denote by $\|\cdot\|_{\mathcal{H}_K}$ the RKHS norm induced by the inner product $ \langle \cdot, \cdot \rangle_{\mathcal{H}_K}$. 
We define the unit ball of $\mathcal{H}_K$ given by 
$$
    \mathcal{K} := \{f\in \mathcal{H}_K: \|f\|_{\mathcal{H}_K} \leq 1\}.
$$
We are interested in  approximating a functional $F:\mathcal{K} \rightarrow \RR$, where $\mathcal{K}$ is regarded as a compact subset of $C(\mathcal{X})$, the space of all continuous functions on $\mathcal{X}$ with norm $\|f\|_\infty = \sup_{x\in \mathcal{X}}|f(x)|$. 
Throughout this paper, we impose the following regularity assumption on $F$.
\begin{assumption}\label{assump:holder-F}
Assume that $F$ is $s$-H\"{o}lder continuous  for some $0<s \leq 1$. That is, there exists a constant $C_F \geq 0$ such that
\begin{equation}
    |F(f) - F(\widetilde{f})| \leq C_F \|f-\widetilde{f}\|^s_{\infty}, \qquad \forall f, \widetilde{f} \in \mathcal{K}.
\end{equation}
\end{assumption}

\second{
\begin{remark}
\label{rem:holder}
Let $\mathcal P(\mathcal X)$ denote the collection of Borel probability measures on $\mathcal X$. For any $f,\widetilde f\in\mathcal K$, 
\[ \sup_{\mu\in\mathcal P(\mathcal X)} \|f-\widetilde f\|_{L^2(\mu)} =\sup_{\mu\in\mathcal P(\mathcal X)} \left(\int_\mathcal X |f(x)-\widetilde f(x)|^2 d\mu \right)^{1/2}= \|f-\widetilde f\|_\infty. \] 
Thus, Assumption~\ref{assump:holder-F} can be interpreted as H\"older continuity with respect to the worst-case $L^2$ discrepancy over all Borel probability measures on $\mathcal X$. Moreover, since $$\|f-\widetilde f\|_{L^2(\mu)} \leq\|f-\widetilde f\|_\infty$$ for every $\mu\in\mathcal P(\mathcal X)$, H\"older continuity with respect to any fixed $L^2(\mu)$ norm implies Assumption~\ref{assump:holder-F}.
The supremum-norm formulation
therefore avoids the need to specify an additional reference measure
on $\mathcal X$. 
\end{remark}}

With this regularity condition in place, we approximate $F$ from
finitely many point evaluations using the standard fully connected
neural networks introduced in Definition~\ref{def:classic}. We consider
either the tanh activation
   $ \sigma(x)=\frac{e^x-e^{-x}}{e^x+e^{-x}}$
or the ReLU activation $\sigma(x)=\max\{x,0\}$.
Our aim is to construct a neural network $\widehat{G}: \RR^N \to \RR$ and obtain an upper bound on the following uniform approximation error
\begin{equation}\label{eq:uniapprox}
    \sup_{f\in \mathcal{K}}\left|F(f) - \widehat{G}(f(\Bar{t}))\right|,
\end{equation}
where the network inputs $f(\Bar{t}) = (f(t_1), \ldots, f(t_{N}))\in \RR^{N}$ are function values at sufficient but finitely many discrete locations $\Bar{t} = \{t_1, \ldots, t_N\}\subset \mathcal{X}$. 
Uniform spacing of locations is not necessary.

As mentioned earlier, our proposed method does not require prior knowledge of the underlying kernel. Furthermore, our point evaluation approach is effective in both Euclidean and non-Euclidean domains, whereas integration-based methods are often ill-suited for non-Euclidean domains, such as set unions involving the unit circle or high-dimensional tori (i.e., products of unit circles). In many practical scenarios, the exact domain 
$\mathcal{X}$ of the input functions may be uncertain (even when it is known to lie within a subset of a Euclidean space). For example, $\mathcal{X}$
could be a unit circle (or more generally, a torus) that is a subset of the 2D cube $[-1,1]^2$ possessing a distinct 1D manifold structure.

In such cases, integrating input functions against \second{kernel sections $K(\cdot, x)$} becomes infeasible, not only because the domain’s geometric structure is unknown, but also because the explicit form of the \second{kernel sections} is typically unavailable. Our proposed method circumvents these limitations by relying solely on point evaluations

\subsection{Main Contributions}
The main novelty of our proposed method lies in its simple representation of functional inputs.
For an input \(f\in\mathcal{H}_K\), we use only its point evaluations at a finite set
\(\Bar{t}={t_1,\ldots,t_N}\subset\mathcal{X}\),
\[
f(\Bar{t})
:=
\big(f(t_1),\ldots,f(t_N)\big)^T
\in\mathbb{R}^N,\]
which can be directly fed into a standard neural network.
The sampling points need not form a uniform partition of \(\mathcal X\).
This design is motivated by the common practice in science and engineering of representing continuous signals, images, and other functional objects through measurements taken at finitely many locations.
Conversely, the integration-type function expansion with a set of pre-specified basis functions employed in previous works is not a typical approach in practice.
Our construction avoids such integration-type preprocessing by exploiting interpolating orthogonal projections in RKHSs: point evaluations alone are sufficient to recover a finite-dimensional approximation of the input function.

Our first set of contributions concerns the approximation of nonlinear functionals on RKHSs.
We establish novel error bounds for the approximation of functionals on RKHS induced by a general Mercer kernel (Theorem \ref{thm:general}).
A central finding of this result is the trade-off associated with the number \(N\) of point evaluations.
Increasing \(N\) improves the accuracy of the interpolating projection, but at the same time increases the input dimension and hence the size of the neural network required to approximate the resulting finite-dimensional map.
By determining the optimal choices of $N$, 
we derive explicit rates of approximating functionals on Sobolev spaces (Theorem \ref{thm:Sobolev}), and on RKHS's induced by the inverse multiquadric kernel (Theorem \ref{thm:multi}), the Gaussian kernel (Theorem \ref{thm:Gaussian}), and kernels on $p$-torus (Theorem \ref{thm:aj} and \ref{thm:1+j}), respectively.
These examples reveal how the geometry and smoothness of the underlying kernel affect the resulting approximation rates.
\second{In Section \ref{subsec:parameter-complexity-comparison}, we further compare the number of network parameters required to achieve a prescribed accuracy across the different kernel classes and between the tanh and ReLU architectures.
This comparison illustrates how both the regularity of the RKHS and the choice of activation function influence the parameter complexity of functional approximation.}

\second{Our second set of contributions addresses statistical learning from finitely many functional observations.
We study empirical risk minimization over the proposed neural-network classes; and establish an oracle inequality showing that the generalization error is controlled by the combination of the neural-network approximation error and the estimation error arising from finite samples (Proposition \ref{thm:oracle-point-evaluation}).
In addition to showing that the proposed networks can approximate RKHS functionals to arbitrary accuracy, our analysis quantifies how the approximation and statistical complexities jointly determine the generalization error.}

 \subsection{Organization of this work}
 The remainder of the paper is organized as follows. Section~\ref{sec:motivation} presents applications that motivate the study of learning nonlinear functionals.
Our approximation results are organized into three parts. Section~\ref{sec:mainresultI} derives explicit approximation rates for RKHSs induced by Sobolev, inverse multiquadric, and Gaussian kernels in Theorems~\ref{thm:Sobolev}, \ref{thm:multi}, and~\ref{thm:Gaussian}, respectively. Section~4 considers RKHSs induced by kernels on the \(p\)-torus and presents Theorems~\ref{thm:aj} and~\ref{thm:1+j}. Finally, Section~\ref{sec:mainresultII} establishes approximation results for functionals on general RKHSs in Theorem~\ref{thm:general}.
\second{Section~\ref{sec:generalization} develops the generalization analysis for learning RKHS functionals}, while Section~\ref{sec:method} outlines the main ideas underlying our neural operator framework. Concluding remarks are given in Section~\ref{sec:conclusion}. Appendix~\ref{sec:append_note} provides additional remarks on reproducing kernels over the $p$-torus, and Appendix~\ref{sec: regression} applies our results to regression maps in generalized functional regression models. Finally, Appendices~\ref{sec:appendA}, \ref{sec:appendcor}, and~\ref{sec:appendB} contain the proofs of the supporting lemmas, supporting corollaries, and main results, respectively.
 
\section{Motivations}
\label{sec:motivation}
Many problems in statistics and machine learning involve learning nonlinear functionals, including functional regression, ODE/PDE learning, and distribution regression. Motivated by the growing availability of functional data and its broad range of applications, we next present several representative settings in which nonlinear functional learning arises.

\vspace{4pt}
\noindent \underline{\bf Example 1: Functional Regression}
\vspace{4pt}

Functional data --- including curves, time series, and images --- are increasingly common. Functional regression studies relationships between functional covariates and response variables.  A classical example is the \textit{generalized functional linear model} (FLM) of \citep{muller2005generalized}.  
In this model, we aim to learn a regression map from a function covariate to a scalar response.
 
Let $X$ be a square-integrable random function defined over a compact interval $\mathcal{X}$, and $Y$ be a scalar response. 
In the model, the conditional mean of $Y$ given $X$ is modeled through a regression functional $F:L^2(\mathcal{X})\rightarrow \RR$, where
\second{
\begin{equation}
Y = F(X)+\varepsilon,
\qquad
F(f) = g\left(\int_{\mathcal X}\beta(t)f(t) dt\right),
\quad f\in L^2(\mathcal X).
\end{equation}}
Here, $\beta \in L^2(\mathcal{X})$ is an unknown functional parameter,
$g$ is a link function that may possess high-order smoothness (e.g., a logit link), and $\varepsilon$ is a centered noise random variable. 
A fundamental problem in statistics is to learn the regression map $F$, and hence the conditional mean $\mathbb E[Y|X]
=
F(X)
=
g\left(\int_{\mathcal X}\beta(t)X(t) dt\right)$.

To solve the functional regression problem, it is crucial to learn the regression map $F:L^2(\mathcal{X})\rightarrow \RR$. Although neural networks have been widely applied to vector, text, and graph data, their use with functional inputs seems less established. 
In this paper, we show that for any given function covariate $X$ residing in a RKHS inside the $L^2$ space, we can approximate the regression map of generalized FLM using fully-connected neural networks. 
Detailed discussions can be found in Appendix \ref{sec: regression}. 

\vspace{10pt}

\noindent \underline{\bf Example 2: ODE/PDE Learning}

\vspace{4pt}

Systems of coupled ODEs and PDEs are undoubtedly the most widely used models in science and engineering, as they often capture the dynamical behavior of physical systems.
Even simple ODE functions can describe complex dynamical behaviors. Recently, an active line of research \citep{raissi2019physics, wang2021learning, li2020fourier} has been developing deep learning tools to infer solutions of ODEs/PDEs with lower cost compared to traditional tools (e.g., finite element methods). 

In solving ODEs and PDEs, we aim to learn a map from the parametric function space to the ODE/PDE solution space. Consider a very simple initial value problem of an ODE: 
\begin{equation}\label{ODE}
    \begin{cases}
        \frac{dh(x)}{dx} = g(x,f(x),h(x)), \qquad x \in [a,b],\\
        h(a) = h_0.
    \end{cases}
\end{equation}
We aim to learn $h(b)$ for any function $f$, that is,  a map $F$ from an input $f$ to an output $h(b)\in \RR$. If we denote the solution to (\ref{ODE}) as $G(f)$, then 
\begin{equation}
   F(f) = G(f)(b),
\end{equation}
where $G$ is a nonlinear operator  being the unique solution to the integral equation
\begin{equation}
    G(f)(x) = h_0 + \int_{a}^x g(t, f(t), G(f)(t))dt.  
\end{equation}

Our work contributes to the area of ODE/PDE learning by proving that a standard fully-connected  neural network is sufficient to approximate (to any accuracy) a functional that is a solution map to a problem like (\ref{ODE}). 

\vspace{10pt}

\noindent \underline{\bf Example 3: Distribution Regression}

\vspace{5pt}

Distribution regression provides a natural setting for learning nonlinear functionals on RKHSs. It concerns regression problems in which the inputs are probability distributions and the responses are real-valued. Let $\mathcal P(\mathcal X)$ denote the set of Borel probability measures on $\mathcal X$. Suppose the data pairs 
$$
\{(\mu_i,Y_i)\}_{i=1}^n,
\qquad
\mu_i\in\mathcal P(\mathcal X),\quad Y_i\in\mathbb R,
$$
are independently drawn from a joint distribution. In practice, each $\mu_i$ is observed only through samples
$$
X_{i,1},\ldots,X_{i,N_i}\overset{\mathrm{i.i.d.}}{\sim}\mu_i.
$$
Thus the observed data we actually have in hand are
\[
\left\{
\left(
\{X_{i,j}\}_{j=1}^{N_i}, Y_i
\right)
\right\}_{i=1}^{n}.
\]
The goal of distribution regression is to learn the map from an underlying probability distribution $\mu$ to its associated response $Y$ using only such two-stage observations.

A classical approach to distribution regression uses kernel mean embeddings \citep{szabo2016learning}. Let $K:\mathcal X\times\mathcal X\to\mathbb R$ be a Mercer kernel with associated RKHS $\mathcal H_K$. A distribution $\mu\in\mathcal P(\mathcal X)$ is represented by its kernel mean embedding

$$
m_\mu
=
\int_{\mathcal X}K(\cdot,x)\,d\mu(x)
\in\mathcal H_K.
$$
Denote the collection of such embeddings by
$$
\mathcal K
=
\{m_\mu:\mu\in\mathcal P(\mathcal X)\}
\subseteq\mathcal H_K,
$$
distribution regression can therefore be viewed as learning a functional
$$
F:\mathcal K\longrightarrow\mathbb R,
$$
through the composition
$$
\mu
\longmapsto
m_\mu
\longmapsto
F(m_\mu).
$$
Thus, learning nonlinear functionals on subsets of RKHSs arises naturally as a component of solving the distribution regression problem.

\second{More recently, \citet{shi2025learning} proposed a neural-network approach in which the fixed kernel feature map is replaced by learned features.  Since $\mu_i$ is not directly observed, the network is evaluated on its empirical measure $\widehat\mu_i=\frac{1}{N_i}\sum_{j=1}^{N_i}\delta_{X_{i,j}}$. Their architecture first applies a neural network $\phi$ pointwise to samples from the input distribution and aggregates the resulting features by averaging,
$$
\widehat\mu_i
\longmapsto
\frac{1}{N_i}\sum_{j=1}^{N_i}\phi(X_{i,j})
=
\int_{\mathcal X}\phi(x)\,d\widehat\mu_i(x),
$$
where the equality follows from the definition of $\mu_i$ and the Dirac measure. Afterwards, another neural network $\psi$  is applied to map this finite-dimensional representation to the response. At the population level, the resulting architecture has the form
$$
\mu
\longmapsto
\int_{\mathcal X}\phi(x)\,d\mu(x)
\longmapsto
\psi\left(
\int_{\mathcal X}\phi(x)\,d\mu(x)
\right).
$$

They further establish approximation guarantees for structured nonlinear functionals of distributions of the form
$$
F(\mu)
=
f\left(
\int_{\mathcal X}G(x)\,d\mu(x)
\right),
$$
including cases in which $G(x)=g(\xi^\top x)$ is a ridge function or $G(x)=g(Q(x))$ with $Q$ a polynomial.}

These approaches illustrate how nonlinear functional learning naturally arises in distribution regression. Our results address the complementary problem of approximating and learning nonlinear functionals defined directly on RKHSs.

\section{Main Results I: Approximation of RKHS Functionals induced by specific kernels}\label{sec:mainresultI}
We present the first part of our main results here.
In this section, we first consider the domain  $\mathcal{X} = [0,1]^d$ and give explicit error rates of approximating functionals on RKHS's--- induced by the following three specific reproducing kernels --- to demonstrate our methods and approaches: 
\begin{itemize}
\item Sobolev space $W^r_2(\mathcal{X})$. It is known that the Sobolev space $W^r_2(\RR^d)$ of order $r$ is an RKHS if $r>d/2$ (see, e.g., \citep{brezis2011functional, wainwright2019high}). 

\item Inverse multiquadric kernel: $K(u,v) = (\sigma^2 + |u-v|^2)^{-\beta}$, where $u,v \in \mathcal{X}$ and $\sigma, \beta >0$. The parameter $\sigma$ is often called the {\it shape parameter}.

\item Gaussian kernel: $K(u,v)= e^{-\frac{|u-v|^2}{2\sigma^2}}$, where $u,v \in \mathcal{X}$ and $\sigma>0$.

\end{itemize} 

\second{The above kernels are important examples of {\it translation-invariant kernels} (also known as {\it convolution-type kernels}), which take the form
\begin{equation}\label{eq:translation}
        K(u,v) = \phi(u-v), \qquad  u,v \in \RR^d
    \end{equation}
    for some function $\phi:\RR^d \rightarrow \RR$. 
    Throughout this paper, we use the Fourier transform for function $\phi$ in $\RR^d$:
    \begin{equation}
    \widehat{\phi}(\xi) = \int_{\RR^d} \phi(x) e^{-2\pi
    i\xi \cdot x} dx, \qquad \xi\in \RR^d.
\end{equation}}
From the  inverse Fourier transform, we know if $\widehat{\phi} \in \second{L^1(\RR^d)}$,
    $\phi(x) = \int_{\RR^d} \widehat{\phi}(\xi) e^{2\pi i\xi \cdot x} d\xi.$
\begin{lemma} \label{lemma:trans}
    A translation-invariant kernel $K$ induced by $\phi: \RR^d \to \RR$ is  positive semi-definite if its Fourier transform $\widehat{\phi}$ is real-valued and non-negative. 
\end{lemma}
The positivity of the Fourier transform
almost everywhere is sufficient for the positive definiteness of a kernel \citep{schaback2006kernel}. 
Lemma \ref{lemma:trans} is a well-established result in the literature, we omit its proof for brevity.

In the following subsections, we present explicit error bounds for approximating functionals on Sobolev space and RKHS's induced by inverse multiquadric and Gaussian Kernels. 
The network input is a function $f$ evaluated at some sampling points $\{f(t_1), \ldots, f(t_N)\}$,
where these points $\{t_1,\ldots, t_N\}$ need not be uniformly partitioned in $\mathcal{X}$. 
Since we consider RKHSs induced by translation-invariant kernels here, we  choose the regular grid $\{t_1,\ldots, t_N\} = \{0, \frac{1}{m},\ldots, \frac{m}{m}\}^d$ for $m\in \NN$ and $N = (m+1)^d$.

\second{In Sections~\ref{subsec:parameter-complexity-comparison} and~\ref{subsec:existing_result}, we further compare the approximation rates across these kernel classes in a unified table, discuss how the geometry of the underlying RKHS affects the results, and relate our results to existing approximation theory in the literature.}

\subsection{Approximation of Functionals on Sobolev Space $W^r_2(\mathcal{X})$}\label{Sobolev}
We consider the Sobolev Space $W^r_2(\mathcal{X})$ of arbitrary real order $r>0$  and $d\in \NN$. 
The following well-known result suggests that when the embedding condition $r>d/2$ holds, $W^r_2(\mathcal{X})$ is a RKHS. 
\begin{proposition}
    When $r>d/2$, the standard Sobolev space $W^r_2(\RR^d)$ is the RKHS induced by a translation-invariant  kernel $K(u,v) = \phi (u-v)$, where 
\begin{equation}\label{fourier_Sobolev}
    \widehat \phi (\xi) = (1+|\xi|^2)^{-r}, \qquad  \xi \in \RR^d.
\end{equation}
\end{proposition}

From Lemma \ref{lemma:trans}, since $\widehat \phi(\xi) = (1+|\xi|^2)^{-r} >0$ for all $\xi \in \RR^d$, $K$  is a reproducing kernel.
The classic Sobolev Embedding Theorem tells us that, for $r>d/2$ and $r-d/2 \notin \NN$, we have $W^r_2(\mathcal{X}) \subset W_\infty^{r-d/2}(\mathcal{X})$, where $W_\infty^{r-d/2}(\mathcal{X})$ represents the H\"older space of order $r-d/2$. 
Moreover, we have
\begin{equation*}
    \|f\|_{W_\infty^{r-d/2}} \leq C_{r,d}  \|f\|_{W^r_2(\RR^d)},
\end{equation*}
where $C_{r,d}$ is a constant depending on only $r$ and $d$.
When $r>d/2$ and $r-d/2 \in \NN$, we can easily extend the above results to get $W^r_2(\mathcal{X}) \subset W_\infty^{r-d/2-\varepsilon}(\mathcal{X})$ and $\|f\|_{W_\infty^{r-d/2 - \varepsilon}} \leq C_{r,d}  \|f\|_{W^r_2(\RR^d)}$, with any $0< \varepsilon < r-d/2$. Hereafter we assume $r-d/2 \notin \NN$.

The following result establishes an explicit error bound in approximating functionals on the Sobolev Space $W^r_2(\mathcal{X})$ using \second{either tanh or ReLU neural networks.} Its proof is relegated to the Appendix \ref{sec:appendB}.

\begin{theorem}[Approximation of Functionals on Sobolev Space]\label{thm:Sobolev}
    Let $r\in \RR_+, d\in \NN$, $\mathcal{X} = [0,1]^d$. Assume that $r-d/2>1$ and $r-d/2 \notin \NN$. \second{Suppose Assumption \ref{assump:holder-F} holds with $0 < s \leq 1$.} 
    Consider $ 
    \mathcal{K} := \{f\in W^r_2(\mathcal{X}): \|f\|_{W^r_2(\mathcal{X})} \leq 1\}
$. 

\begin{enumerate}
    \item[\textnormal{(i)}] \textbf{Tanh approximation.}
    For every $M\in \NN$ with $M>(5\cdot 4^d)^3$, let
\begin{eqnarray*}
        m
        =\left\lceil M^{\frac{s}{s(4r -\frac{3d}{2}-1)+\frac{5d}{2}}}\right\rceil, \qquad \Bar{t}=\left\{0, \frac{1}{m},\ldots, \frac{m}{m}\right\}^d, \qquad \hbox{and } \quad  N= (m+1)^d.
    \end{eqnarray*}
    Then there exists a tanh neural network
    $\widehat G:\mathbb{R}^{N}\to\mathbb{R}$ with \underline{two} hidden layers of widths at most $N(M-1)$ and $3\left(\frac{N+1}{2}\right) (5M)^N$ 
    such that
\begin{equation*}
    \sup_{f\in \mathcal{K}}\left|F(f) - \widehat{G}(f(\Bar{t}))\right|\leq C_{r, s, d, F} \left(\frac{1}{M}\right)^{\frac{s^2 (2
    r-d)}{s(4r-\frac{3d}{2}-1)+\frac{5d}{2}}},
\end{equation*}
where $C_{r, s, d, F}$ is a constant depending on $r, s, d$ and the Lipschitz constant of $F$.

\second{
    \item[\textnormal{(ii)}] \textbf{ReLU approximation.} Let $\gamma = s\left(4r-\frac{3d}{2}-1\right)+\frac d2$. For every $M, L \in \NN$ with $ML \geq e^{\frac{2\gamma}{s}}$, let \begin{eqnarray*}
        m
=
\left\lfloor \left(\frac{B}{\log B}\right)^{\frac{1}{d}}\right\rfloor-1 \quad \hbox{with} \quad  B=\frac{2sd}{\gamma} \log(ML),\\
       \Bar{t}=\left\{0, \frac{1}{m},\ldots, \frac{m}{m}\right\}^d, \qquad \hbox{and } \quad  N= (m+1)^d.
    \end{eqnarray*}
    There there exists a ReLU neural network $\widehat{G}: [-1, 1]^N \to {\mathbb R}$ with depths $11L + 2N + 18$ and widths $3^{N+3}\max\{N \lfloor M^{1/N} \rfloor, M+2\}$ such that 
\begin{eqnarray*}
\sup_{f\in\mathcal K}
\left|F(f)-\widehat G(f(\bar t))\right|
\leq
C_{r,s,d,F}
\left(
\frac{\log(ML)
}{
 \log\left(\log(ML)\right)
}
\right)^{-\frac{s(2r-d)}{d}}
\end{eqnarray*}
where $C_{r, s, d, F}$ is a constant depending on $r, s, d$ and the Lipschitz constant of $F$.
}
\end{enumerate}

\end{theorem}

\subsection{Approximation of RKHS Functionals Induced by Inverse Multiquadric Kernel}\label{inverse}
Here, we present the result on the approximation of functionals defined over $\mathcal{K}$, an RKHS induced by the inverse multiquadric kernel 
\begin{equation}\label{multi}
    K(u,v) = (\sigma^2 + |u-v|^2)^{-\beta}, \qquad  u,v \in \mathcal{X}, \ \sigma, \beta >0.
\end{equation}
The inverse multiquadric kernel is a reproducing kernel as long as $\sigma,\beta >0$ (see \citep{bozzini2015radial, tolstikhin2017minimax}). 
The proof of the following theorem is given in Appendix \ref{sec:appendB}. 

\begin{theorem}[Approximation of RKHS Functionals induced by inverse multiquadric Kernel]\label{thm:multi}
Let $d\in \NN$, $\sigma, \beta >0$, $\mathcal{X} = [0,1]^d$, $M_d = 12 \left(\frac{\pi \Gamma^2\left(\frac{d+2}{2}\right)}{9}\right) \leq 6.38d$, and $c>0$ is a positive constant given for the multiquadric kernel by (\ref{f-Pf_multi}).
\second{Suppose Assumption \ref{assump:holder-F} holds with $0 < s \leq 1$.} 
Consider $
    \mathcal{K} := \{f\in \mathcal{H}_K: \|f\|_{\mathcal{H}_K} \leq 1\}
$ with $\mathcal{H}_K$ induced by an inverse multiquadric kernel given in (\ref{multi}). \begin{enumerate}
    \item[\textnormal{(i)}] \textbf{Tanh approximation.}
    There exists some $M_0>0$ such that for every $M\in \NN$ with $M>M_0$, by taking 
\begin{eqnarray*}
        m=\left\lceil\frac{\log (M)}{4M_d\sigma+ \frac{c}{\sqrt{d}}}\right\rceil, \qquad \Bar{t}=\left\{0, \frac{1}{m},\ldots, \frac{m}{m}\right\}^d, \qquad \hbox{and } \quad  N= (m+1)^d,
    \end{eqnarray*}
  there exists a tanh neural network
    $\widehat G:\mathbb{R}^{N}\to\mathbb{R}$ with \underline{two} hidden layers of widths at most $N(M-1)$ and $3\left(\frac{N+1}{2}\right) (5M)^N$ 
    such that
\begin{equation*}
    \sup_{f\in \mathcal{K}}\left|F(f) - \widehat{G}(f(\Bar{t}))\right|\leq C_{\sigma, \beta, s, d, F} (\log (M))^{\frac{5}{2}d - \frac{s}{2}}\left(\frac{1}{M}\right)^{\frac{cs}{4M_d \sqrt{d}\sigma+c}},
\end{equation*}
where $C_{\sigma, \beta, s, d, F}$ is a constant depending on $\sigma, \beta, s, d$ and the Lipschitz constant of $F$.

\second{
    \item[\textnormal{(ii)}] \textbf{ReLU approximation.}  For every $M, L \in \NN$ with $ML \geq e^{2^d (4\sigma M_d + c/\sqrt{d})}$, let
\begin{equation*}
    m
= \left\lfloor \left(\frac{2\log(ML)}{4\sigma M_d +c/\sqrt{d}}\right)^{\frac{1}{d+1}}\right\rfloor -1,
\quad \Bar{t}=\left\{0, \frac{1}{m},\ldots, \frac{m}{m}\right\}^d, \quad \hbox{and } \quad  N= (m+1)^d.
\end{equation*}
    There there exists a ReLU neural network $\widehat{G}: [-1, 1]^N \to {\mathbb R}$ with depths $11L + 2N + 18$ and widths $3^{N+3}\max\{N \lfloor M^{1/N} \rfloor, M+2\}$ such that 
\begin{eqnarray*}
\sup_{f\in\mathcal K}
\left|F(f)-\widehat G(f(\bar t))\right|
\leq
C_{\sigma,\beta,d,s,F}
\exp\left\{-\frac{cs}{4\sqrt{d}}\left(\frac{2\log(ML)}{4\sigma M_d +c/\sqrt{d}}\right)^{\frac{1}{d+1}}
\right\},
\end{eqnarray*}
where $C_{\sigma, \beta, s, d, F}$ is a constant depending on $\sigma, \beta, s, d$ and the H\"{o}lder constant of $F$.
}
\end{enumerate}
\end{theorem}

\subsection{Approximation of RKHS Functionals Induced by Gaussian Kernel}\label{subsec:Gaussian}
The Gaussian kernel is an important translation-invariant kernel given by
\begin{equation}\label{Gaussian}
    K(u,v)= e^{-\frac{|u-v|^2}{2\sigma^2}}, \qquad u,v \in \mathcal{X}, \sigma >0,
\end{equation}
or $\phi(x) = e^{-\frac{x^2}{2\sigma^2}}$ for $x\in\mathcal{X}$. 
From Lemma \ref{lemma:trans}, the Gaussian kernel is a reproducing kernel because its Fourier transform is known to be $\widehat{\phi}(\xi) = (2\sigma^2\pi)^{\frac{d}{2}}e^{-2\sigma^2\pi^2 |\xi|^2}> 0$ for all $\xi \in \RR^d$. 
The following results give explicit error bounds of the approximation of functional $F$ over $\mathcal{K} = \{f\in \mathcal{H}_K: \|f\|_{\mathcal{H}_K} \leq 1\}
$ with $\mathcal{H}_K$ induced by a Gaussian kernel. Its proof is given in Appendix \ref{sec:appendB}. 

\begin{theorem}[Approximation of RKHS Functionals induced by Gaussian Kernel]\label{thm:Gaussian}
Let $d\in \NN$, $\sigma >0$, $\mathcal{X} = [0,1]^d$. We also let $c>0$ be some positive constant. 
\second{Suppose Assumption \ref{assump:holder-F} holds with $0 < s \leq 1$.} 
\begin{enumerate}
    \item[\textnormal{(i)}] \textbf{Tanh approximation.} There exists some $M_0\in \NN$ such that for every $M\in \NN$ with $M>M_0$, by taking
\begin{equation}\label{m}
        m = \left\lfloor\frac{\sqrt{2}}{\sigma \pi \sqrt{d}}\left(\sqrt{\log M} - c (\log M)^{\frac{1}{4}}\right) \right\rfloor, \quad   \Bar{t}=\left\{0, \frac{1}{m},\ldots, \frac{m}{m}\right\}^d, \quad  N= (m+1)^d,
    \end{equation} 
 we have a tanh neural network $\widehat{G}: \RR^N \to \RR$ with two hidden layers of widths at most $N(M-1)$ and $3\left\lceil\frac{N+1}{2}\right\rceil (5M)^N$ satisfying 
\begin{equation}\label{thm:Gaussian_bound}
    \sup_{f\in \mathcal{K}}\left|F(f) - \widehat{G}(f(\Bar{t}))\right|\leq C_{\sigma, c, s, d, F} \ \exp\left(-\frac{cs}{8\sigma \pi d} \sqrt{\log M}  \log (\log (M))\right),
\end{equation}
where $C_{\sigma, c,  s, d, F}$ is a constant depending on $\sigma, c, s, d$ and the Lipschitz constant of $F$.

\second{
    \item[\textnormal{(ii)}] \textbf{ReLU approximation.} For every $M, L \in \NN$, let
$C_0=\left(\frac{1}{2^{d-1}\sigma^2\pi^2 d}\right)^{\frac{1}{d+2}}$.
Take
\begin{equation*}
m=\left\lfloor C_0 (\log(ML))^{\frac{1}{d+2}}\right\rfloor, \qquad \Bar{t}=\left\{0, \frac{1}{m},\ldots, \frac{m}{m}\right\}^d, \qquad \hbox{and } \quad  N= (m+1)^d.
\end{equation*}
For $ML$ sufficiently large, specifically for
   $ \log(ML)>
\left(\frac{2\sqrt d}{C_0}\right)^{d+2}$,
we have
\[
m=\left\lfloor C_0 (\log(ML))^{\frac{1}{d+2}}\right\rfloor
\geq
\frac{C_0}{2}(\log(ML))^{\frac{1}{d+2}}
>
\sqrt d.
\]
    Then there exists a ReLU neural network $\widehat{G}: [-1, 1]^N \to {\mathbb R}$ with depths $11L + 2N + 18$ and widths $3^{N+3}\max\{N \lfloor M^{1/N} \rfloor, M+2\}$ such that 
\begin{eqnarray*}
&&\sup_{f\in\mathcal K}
\left|F(f)-\widehat G(f(\bar t))\right|\\
&\leq&
C_{\sigma,c,d,s,F}
\exp\left(
-\frac{csC_0}{4(d+2)\sqrt d}
\left(\log\!\left(ML\right)\right)^{\frac{1}{d+2}}
\log\log\!\left(ML\right)
\right),
\end{eqnarray*}
where $C_{\sigma, c,  s, d, F}$ is a constant depending on $\sigma, c, s, d$ and the Lipschitz constant of $F$.
}
\end{enumerate}
\end{theorem}

\second{
\subsection{Comparison of Parameter Complexity Across Kernel Classes}
\label{subsec:parameter-complexity-comparison}
The preceding theorems provide explicit approximation error bounds for RKHS functionals associated with the Sobolev, inverse multiquadric, and Gaussian kernels. To facilitate comparison across these kernel classes, we reformulate these bounds in terms of the network size required to achieve a prescribed accuracy  $\epsilon >0$. Specifically, for each kernel class and network architecture, we find the number of network parameters so that the approximation error is at most $\epsilon$, and then express the resulting total number of network parameters as a function of $\epsilon$. The resulting parameter-complexity are summarized in Table~\ref{tab:compare}. }

\begin{table}[h]
\centering
\normalsize
\renewcommand{\arraystretch}{1.8}
\setlength{\tabcolsep}{8pt}
\caption{\second{Required number of network parameters for approximating functionals defined on RKHSs induced by different kernels up to accuracy $\varepsilon>0$.}}
\label{tab:compare}
\second{
\begin{tabularx}{\textwidth}{|>{\centering\arraybackslash}p{0.23\textwidth}|
                                >{\centering\arraybackslash}X|
                                >{\centering\arraybackslash}X|}
\hline
\multirow{2}{*}{\textbf{Kernel class}}
& \multicolumn{2}{c|}{\textbf{Number of network parameters}} \\
\cline{2-3}
& \textbf{tanh network}
& \textbf{ReLU network} \\
\hline

Sobolev kernel 
&
\makecell[c]{
$\displaystyle
\exp\left\{
\mathcal O\left(
\varepsilon^{-\frac{d}{s(2r-d)}}\log\frac{1}{\varepsilon}
\right)
\right\}$
}
&
\makecell[c]{
$\displaystyle
\exp\left\{
\mathcal O\left(
\varepsilon^{-\frac{d}{s(2r-d)}}\log\frac{1}{\varepsilon}
\right)
\right\}$
}
\\
\hline

Inverse multiquadric kernel 
&
\makecell[c]{
$\displaystyle
\exp\left\{
\mathcal O\left(
\left(\log\frac{1}{\varepsilon}\right)^{d+1}
\right)
\right\}$
}
&
\makecell[c]{
$\displaystyle
\exp\left\{
\mathcal O\left(
\left(\log\frac{1}{\varepsilon}\right)^{d+1}
\right)
\right\}$
}
\\
\hline

Gaussian kernel
&
\makecell[c]{
$\displaystyle
\exp\left\{
\mathcal O\left(
\left(
\frac{\log(1/\varepsilon)}
{\log\log(1/\varepsilon)}
\right)^{d+2}
\right)
\right\}$
}
&
\makecell[c]{
$\displaystyle
\exp\left\{
\mathcal O\left(
\left(
\frac{\log(1/\varepsilon)}
{\log\log(1/\varepsilon)}
\right)^{d+2}
\right)
\right\}$
}
\\
\hline
\end{tabularx}}
\end{table}

\second{
The parameter-complexity bounds in Table~\ref{tab:compare} are obtained by inverting the corresponding approximation error bounds and substituting the resulting choices into the network parameter count. Because this conversion follows the same standard calculation in each case, repeating the calculation for every kernel class would be repetitive. Thus, we only provide a detailed derivation for the Gaussian kernel as a representative example in Appendix~\ref{app:parameter-complexity}; the remaining bounds follow analogously from their corresponding approximation theorems.

\paragraph{Effect of the kernel class.}
Table~\ref{tab:compare} reveals a clear distinction between kernels of finite and infinite smoothness. The Sobolev kernel of order $r$ is an RKHS and a function class with finite smoothness. Its parameter complexity is exponential in a negative power of $\varepsilon$. In particular, the exponent
\[
\frac{d}{s(2r-d)}
\]
decreases as the smoothness order $r$ increases, showing that the unit ball of a higher-order Sobolev space can be approximated more efficiently. Nevertheless, complexity deteriorates with the ambient dimension $d$ and improves with the H\"older regularity $s$ of the functional. By contrast, the inverse multiquadric and Gaussian kernels are infinitely smooth, and their parameter complexities depend exponentially only on the powers of $\log(1/\varepsilon)$. These bounds are therefore substantially smaller than the Sobolev case. Because their spectral densities decay rapidly, the inverse multiquadric and Gaussian kernels strongly restrict how rapidly functions in their RKHS unit balls can oscillate.
Consequently, these functions are smoother and belong to a more restricted input class, which leads to more favorable approximation bounds.

Among the two infinitely smooth kernels, the Gaussian kernel has a faster spectral decay and induces a more restrictive RKHS: its Fourier transform decays like $\exp(-c\|\xi\|^2)$, whereas that of the inverse multiquadric kernel decays essentially like $\exp(-c\|\xi\|)$ up to polynomial factors. Nevertheless, the upper bound obtained for the inverse multiquadric kernel is asymptotically smaller than the Gaussian-kernel bound. Indeed, we know
\[
\log(1/\varepsilon)^{d+1}
=
o\left(
\left(\frac{\log(1/\varepsilon)}{\log \log(1/\varepsilon)}\right)^{d+2}
\right).
\]
This does not contradict the greater smoothness or smaller capacity of the Gaussian RKHS. The derived complexity depends not only on the size of the RKHS unit ball, but also on the stability of the sampling-based representation and the regularity of the resulting finite-dimensional map. In our construction, the relevant regularity bound grows exponentially in $m^2$ for the Gaussian kernel, compared with exponential growth in $m$ (up to polynomial factors) for the inverse multiquadric kernel. Balancing this growth against the kernel interpolation error produces the different rates in Table~\ref{tab:compare}.

\paragraph{Effect of the activation function.}
Finally, tanh and ReLU networks attain the same leading-order parameter complexity for every kernel class  in Table~\ref{tab:compare}. This indicates that, within our framework, the dominant dependence on $\varepsilon$ is governed primarily by the kernel-induced discretization error and the regularity of the functional, rather than by the activation function. The two activations nevertheless lead to different network constructions: the tanh approximation uses a shallow architecture, whereas the ReLU approximation generally requires greater depth. The matching asymptotic orders do not imply that the two architectures are identical in practice, but they show that the smoothness of the tanh activation does not improve the leading-order parameter complexity over ReLU for the kernel classes studied here.}

\second{
\subsection{Comparison with Existing Approximation Results}\label{subsec:existing_result}
To the best of our knowledge, the only existing approximation results for nonlinear functionals defined on RKHSs are those of \citep{shi2025nonlinear}.

As discussed in Section~\ref{sec:relatedwork}, their method is a kernel-embedded variant of  functional MLP, with the following architecture:
$$
f
\;\xmapsto{\;K\;}\;
L_Kf
\;\xmapsto{\;K_0\;}\;
T(L_Kf)
=
\left(
\langle L_Kf,\phi_1\rangle_{L^2(\mu)},
\ldots,
\langle L_Kf,\phi_{d_1}\rangle_{L^2(\mu)}
\right)^T
\;\longmapsto\;
\text{ReLU network},
$$
where \(K\) is an embedding kernel and \(\{\phi_j\}_{j=1}^{d_1}\) are the leading eigenfunctions associated with a projection Mercer kernel \(K_0\). Their general framework considers input function classes \(\mathcal F\subset L_\infty(\mathcal X)\cap L_2(\mathcal X)\), and they further study Gaussian RKHS and Sobolev space as two special examples.
They assume that the target functional \(F:\mathcal F\to\mathbb R\) has modulus of continuity satisfying
$$
\omega_F(q)\le C_F q^\lambda,\qquad 0<\lambda\le1,
$$
where
$$
\omega_F(q)
=
\sup_{\substack{f,g\in\mathcal F\\
\|f-g\|_{L^2(\mu)}\le q}}
|F(f)-F(g)|.
$$
Equivalently, \(F\) is H\"older continuous with respect to the \(L^2(\mu)\) norm:
$$
|F(f)-F(g)|
\le C_F\|f-g\|_{L^2(\mu)}^\lambda,
$$
where \(\mu\) is a prescribed positive Borel measure on \(\mathcal X\).

For the approximation of nonlinear functionals defined on a Gaussian RKHS, \citep{shi2025nonlinear} assume that the input functions belong to some Gaussian RKHS and exploit this prior knowledge in their construction by choosing both the embedding and projection kernels based on the underlying Gaussian kernel. They also assume that a reference measure $\mu$ on the input domain is specified. They obtain an approximation rate of 
\[
\mathcal O\left(
\exp\left(-c(\log \mathcal{N})^{\frac{1}{d+1}}\right)\right),
\]
where $\mathcal{N}$ is the number of ReLU network parameters. This corresponds to a parameter complexity
$$\exp\left\{
\mathcal O\left(
\left(\log\frac{1}{\varepsilon}\right)^{d+1}
\right)
\right\}$$
which is asymptotically smaller than our Gaussian-kernel bound in Table~\ref{tab:compare}. Their result, however, is obtained through a kernel-adapted construction that explicitly uses knowledge of the underlying Gaussian kernel and its associated eigenfunctions. In contrast, our architecture uses only the point evaluations \(f(t_1),\ldots,f(t_N)\) and does not require the underlying kernel or its eigensystem to be known, which can be particularly useful when the function space generating the inputs is unknown in practice.

For input functions in a compact subset of the zero-boundary Sobolev space \(H_0^\alpha([0,1]^d)\), \citep{shi2025nonlinear} obtain an approximation rate of
$$
\mathcal O\left(
(\log \mathcal N)^{-\frac{\alpha\lambda}{d}}
\right),
$$
up to \(\log\log \mathcal N\) factors.
This has the same broad logarithmic form as our approximation result for functionals on the Sobolev space \(W_2^r([0,1]^d)\) , and both results correspond to parameter complexities exponential in a negative power of \(\varepsilon\).
The two settings are nevertheless different: \citep{shi2025nonlinear} consider \(H_0^\alpha([0,1]^d)\), whereas our result applies to the standard Sobolev space \(W_2^r([0,1]^d)\) of arbitrary real order $(r>d/2)$. Their construction again uses specially chosen embedding and projection kernels adapted to the assumed smoothness of the input class, while our representation relies only on point evaluations.

We emphasize that the rates in \citep{shi2025nonlinear} and ours are not directly comparable, since the regularity assumptions imposed on the target functional are different.}

\begin{remark}[Comparison with a lower bound for $L^p$ functionals]
\label{rem:lp-lower-bound}
A useful benchmark is provided by
\citep[Theorem~2.2]{mhaskar1997neural}, who studied the approximation
of continuous functionals
\[
    F:L^p(\mathcal X)\to\RR,
    \qquad 1\leq p\leq\infty,
\]
on certain compact subsets of
$W^\beta_\infty(\mathcal X)$, where $\beta>0$ describes the
smoothness of the input functions. For functionals satisfying a
H\"older condition of order $0<\lambda\leq1$, they obtained a lower bound
\[
    \sup_{f}
    \left|F(f)-\widehat G(f)\right|
    \geq
    C(\log\mathcal N)^{-\frac{\beta\lambda}{d}},
\]
where $\widehat G$ is a neural network with a sigmoid-type infinitely
differentiable activation function (e.g., tanh), and
$\mathcal N$ denotes the number of network parameters.

Equivalently, to attain an approximation error at most
$\varepsilon$, the worst-case parameter complexity must satisfy
\[
    \mathcal N
    \geq
    \exp\left\{
        \Omega\left(
            \varepsilon^{-\frac{d}{\beta\lambda}}
        \right)
    \right\}.
\]
Thus, a logarithmic approximation rate corresponds to a parameter
requirement that is exponential in a negative power of $\varepsilon$.

This lower bound has the same broad stretched-exponential form as the
upper bound obtained in Table~\ref{tab:compare} for the Sobolev kernel,
namely
\[
    \exp\left\{
        \mathcal O\left(
            \varepsilon^{-\frac{d}{s(2r-d)}}
            \log\frac{1}{\varepsilon}
        \right)
    \right\}.
\]
By contrast, our upper bounds for the inverse multiquadric and Gaussian
kernels depend exponentially only on powers of
$\log(1/\varepsilon)$:
\[
    \exp\left\{
        \mathcal O\left(
            \left(\log\frac{1}{\varepsilon}\right)^{d+1}
        \right)
    \right\}
    \qquad \hbox{and} \qquad \exp\left\{
        \mathcal O\left(
            \left(
                \frac{\log(1/\varepsilon)}
                     {\log\log(1/\varepsilon)}
            \right)^{d+2}
        \right)
    \right\},
\]
respectively. These quantities grow asymptotically more slowly than
$\exp\{c\varepsilon^{-q}\}$ for every $c,q>0$. 

We emphasize, however, that these results are not directly comparable.
The lower bound of \citet{mhaskar1997neural} concerns functionals defined on compact subsets of $L^p$-type function
spaces, whereas our results concern H\"older functionals defined on
RKHS unit balls. 
Nevertheless, their result suggests
that the approximation of continuous functionals may be in general n slow.
\end{remark}

\section{Main Result II: Approximation of Functionals on Torus}\label{subsec:ptorus}

Periodic and angular variables arise naturally in a variety of applications, including directional statistics, geoscience, and molecular structural analysis \citep{mardia2007protein,lund1999cluster,agostinelli2007robust,eltzner2018torus}. A univariate angular observation can be represented as a point on the unit circle $\mathbb S^1$, or equivalently by an angle $\theta\in[0,2\pi)$. Examples include bird flight directions and the orientations of wind and ocean currents \citep{di2011kernel}. When several angular variables are observed jointly, their natural domain is a product of circles, giving rise to toroidal data. For example, RNA backbone conformations are characterized by multiple torsion angles and can therefore be naturally represented on a toroidal domain \citep{murray2003conformations,eltzner2018torus}.

Motivated by such periodic data, in this section we study the approximation of functionals defined on RKHSs over the \(p\)-torus. \second{The analysis may be viewed as the periodic counterpart of the translation-invariant kernel setting on \(\mathbb R^d\) considered earlier. In the Euclidean setting, translation-invariant kernels are characterized through their Fourier transforms; on the torus, the analogous spectral representation is given by Fourier series, with Fourier coefficients indexed by \(j\in\mathbb Z^p\). We exploit this representation to relate the spectral decay of the kernel to the approximation error of our point-evaluation discretization and thereby derive explicit neural network approximation rates.}

The $p$-torus is the $p$-fold product of a unit circle denoted by $\mathbb{T}^p$. Recall that the unit circle is given by
\begin{equation}
    \mathbb{S}^1 : = \{x\in\RR^2 : \|x\|_{2} = 1\} \subset \RR^2, \end{equation}
    which, upon identifying \(\mathbb R^2\) with \(\mathbb C\), can equivalently be written as
    \begin{equation}
\mathbb{S}^1
=
\{e^{it}:t\in\mathbb{R}\}.
\end{equation}
The $p$-torus is then defined by 
\[
\mathbb T^p:=(\mathbb S^1)^p, \qquad p \geq 1,
\]
and admits the standard embedding
\[
\mathbb{T}^p=
\left\{
\big((\cos \xi_1,\sin \xi_1),\ldots,(\cos \xi_p,\sin \xi_p)\big):
\xi\in\mathbb{R}^p
\right\}
\subset \mathbb{R}^{2p}.
\]
\second{Equivalently, it can be viewed intrinsically as the quotient space
$\mathbb{T}^p \simeq \mathbb{R}^p/(2\pi\mathbb{Z})^p$,
so that functions on $\mathbb{T}^p$ can be identified with functions on $\mathbb{R}^p$ that are $2\pi$-periodic in each coordinate. 
Under this representation, the characters
$$
e^{ij\cdot \xi},
\qquad j\in\mathbb Z^p,
$$
form the natural Fourier basis on \(\mathbb T^p\).}

We first consider the kernel $K$ on $\mathbb{S}^1$
\begin{equation}\label{kernel_S1}
K(e^{i\xi},e^{i\eta})=
\sum_{j=-\infty}^{\infty}
\theta_j e^{ij(\xi-\eta)},
\qquad \xi,\eta\in\mathbb{R},
\end{equation}
where the coefficients $\theta_j$ satisfy $\theta_j=\theta_{-j}$ and $\theta_j\geq 0$ for all $j\in\mathbb{Z}$. This kernel is symmetric, real-valued, and positive semidefinite; see Remark \ref{remark:PSD} in Appendix \ref{sec:append_note}.
More generally, on \(\mathcal X=\mathbb T^p\), we consider kernels of the form
\begin{equation}\label{kernel_torus}
K(e^{i\xi},e^{i\eta})
=
\sum_{j\in\mathbb{Z}^p}
\theta_j e^{ij\cdot(\xi-\eta)},
\qquad
\xi,\eta\in\mathbb{R}^p,
\end{equation}
where $\theta_j \geq 0$ for all $j\in\ZZ^p$, $\theta_{j} = \theta_{\varepsilon \circ j}$ with $\varepsilon \circ j := (\varepsilon_1 j_1, \ldots, \varepsilon_p j_p)$ for all $\varepsilon \in \{-1,1\}^p$, and $j\cdot x =  j_1x_1 + \cdots + j_px_p$.  
As shown in Remark \ref{remark:PSD_torus} in Appendix \ref{sec:append_note}, the kernel defined in \eqref{kernel_torus} is positive semidefinite. In what follows, we consider RKHSs induced by kernels of this form \eqref{kernel_torus}.

\second{The Fourier coefficients \(\{\theta_j\}_{j\in\mathbb Z^p}\) play a role analogous to the Fourier transform of a translation-invariant kernel on \(\mathbb R^d\). Different rates of decay of \(\theta_j\) therefore give rise to RKHSs with different spectral regularity and, as shown below, different approximation rates.}

To approximate functionals defined on RKHS over \(\mathbb T^p\), we again discretize the input functions through finitely many point evaluations. As discussed in Section \ref{sec:relatedwork}, many existing approaches represent an input function through its inner products with a pre-specified set of basis functions. 
Such approaches can be particularly useful when a suitable basis is explicitly known. 
\second{On \(\mathbb{T}^p\), the natural choice is the Fourier basis arising from its periodic structure.}

Our construction takes a different approach. Our neural network receives only the point evaluations
$$
\bigl(f(t_1),\ldots,f(t_N)\bigr),
\qquad
t_1,\ldots,t_N\in\mathbb T^p,
$$
and therefore does not require the network architecture itself to be built from a Fourier basis. 
\second{More generally, this feature is useful in settings where a suitable domain-adapted basis is not explicitly known.} For RKHS inputs, point evaluations are also naturally compatible with the reproducing property.

\second{The Fourier structure of \(\mathbb T^p\) is nevertheless important for our theoretical analysis. In particular, the Fourier coefficients \(\{\theta_j\}\) of the kernel allow us to quantify the error incurred by discretizing the input function through finitely many point evaluations. Thus, the Fourier basis is used in deriving our approximation theory, but it is not built into the architecture.}

Different choices of Fourier coefficients $\theta_j$ in the reproducing kernel \eqref{kernel_torus} induce different RKHSs. The decay of \(\theta_j\) determines the spectral regularity of the corresponding RKHS. We consider two representative spectral-decay regimes.
Let $\|j\|^2= \sum_{k=1}^p |j_k|^2$. 
We consider:
\begin{enumerate}
 \item $\theta_j = a^{\|j\|}, \qquad \forall j\in\ZZ^p \text{ with some } a\in (0,1)$; \label{coefficient1}
    \item $\theta_j = \left(1+\|j\|^2\right)^{-\gamma}, \qquad \forall j\in\ZZ^p, \gamma>(p+1)/2. $ \label{coefficient2}
\end{enumerate}

To discretize the input functions through point evaluations, we consider an equispaced product grid on \(\mathcal{X}=\mathbb{T}^p\). For an integer \(m\geq 1\), let \(N=m^p\) and define
\begin{eqnarray}  \bar{t}&=&\left(t_k:=e^{i\frac{2\pi}{m}k}\right)_{k\in \{0,1,\ldots, m-1\}^p}\label{torus_points}\\
     &=&\left( e^{i\frac{2\pi}{m}k_1}, \cdots, e^{i\frac{2\pi}{m}k_p}\right), \qquad \text{for } k_1,\ldots, k_p \in \{0,1,\ldots, m-1\}. \nonumber
\end{eqnarray}
Thus, each coordinate of the torus is sampled at $m$ equally spaced points, resulting in a total of $N=m^p$ evaluation points.

We adopt the grid $\bar t$ in \eqref{torus_points} to derive explicit approximation error rates for the two classes of RKHSs considered above. The use of an equispaced grid, however, is not essential to our general approximation framework. As shown in Theorem \ref{thm:general} in the next section, nonlinear functionals can be approximated using any sufficiently dense collection of evaluation points.

We next present two approximation results corresponding to the two spectral-decay regimes introduced above. Theorem \ref{thm:aj} considers the RKHS induced by the kernel \eqref{kernel_torus} with exponentially decaying coefficients
$\theta_j = a^{\|j\|}$
whereas Theorem \ref{thm:1+j} considers the polynomially decaying coefficients $\theta_j = \left(1+\|j\|^2\right)^{-\gamma}$.
The proofs are deferred to Appendix \ref{sec:appendB}.

\begin{theorem}[Approximation of functionals on toroidal RKHSs with exponential spectral decay]
\label{thm:aj}
Let $p\in\mathbb{N}, \mathcal{X}=\mathbb{T}^p$, and \second{suppose Assumption \ref{assump:holder-F} holds for some \(0<s\leq 1\).} Let \(\mathcal{H}_K\) be the RKHS induced by the kernel $K$ in \eqref{kernel_torus} with
\[
\theta_j=a^{\|j\|},\qquad j\in\mathbb{Z}^p,
\]
for some $a\in(0,1)$, and define its unit ball by $
    \mathcal{K} := \{f\in \mathcal{H}_K: \|f\|_{\mathcal{H}_K} \leq 1\}.
$

\begin{enumerate}
\item[\textnormal{(i)}] \textbf{Tanh networks.}
There exists $M_0\in\mathbb{N}$ such that, for every $M>M_0$, let
\[
m=  \left \lfloor\frac{2}{\sqrt{p}s\log(1/a)}\log\left(\frac{M^s}{(\log(M))
       ^{\frac{p}{2}(5-s)
       }}\right)\right\rfloor,
\qquad
N=m^p,
\]
and let $\bar t\subset\mathbb{T}^p$ be the equispaced product grid in \eqref{torus_points} corresponding to this choice of $m$. Then there exists a two-hidden-layer tanh neural network
$\widehat G:\mathbb{R}^N\to\mathbb{R}$
with widths at most
$N(M-1)$ and
$3\left\lceil\frac{N+1}{2}\right\rceil(5M)^N$,
respectively, such that
\[
\sup_{f\in\mathcal K}
\left|
F(f)-\widehat G\big(f(\bar t)\big)
\right|
\leq
C_{a,p,s,F}(\log M)^{-\frac{s}{2}},
\]
where $ C_{a,p,s,F}$ is a constant depending on $a,p,s$ and the H\"{o}lder constant $C_F$.

\second{
\item[\textnormal{(ii)}] \textbf{ReLU networks.}
For $M,L\in\mathbb{N}$, let
\[
C_0=
\left(
\frac{1}{\sqrt p\log(1/a)}
\right)^{\frac{1}{p+1}},
\qquad
m
=
\left\lfloor
C_0(2\log(ML))^{\frac{1}{p+1}}
\right\rfloor,
\qquad
N=m^p,
\]
and let $\bar t\subset\mathbb{T}^p$) be the corresponding equispaced product grid in \eqref{torus_points}. For $ML$ sufficiently large, there exists a ReLU neural network
$\widehat G:[-1,1]^N\to\mathbb{R}$
with depth at most
\[
11L+2N+18
\]
and width at most
\[
3^{N+3}
\max\left\{
N\left\lfloor M^{1/N}\right\rfloor,
M+2
\right\},
\]
such that
\[
\sup_{f\in\mathcal K}
\left|
F(f)-\widehat G\big(f(\bar t)\big)
\right|
\leq
C_{a,p,s,F}
\big(\log(ML)\big)^{-\frac{s}{2(p+1)}}.
\]
Here,  $ C_{a,p,s,F}$ is a constant depending on $a,p,s$ and the H\"{o}lder constant $C_F$.}
\end{enumerate}
\end{theorem}

\begin{theorem}[Approximation of functionals on toroidal RKHSs with polynomial spectral decay]
\label{thm:1+j}
Let $p\in\mathbb{N}, \mathcal{X}=\mathbb{T}^p$, and \second{suppose Assumption \ref{assump:holder-F} holds for some $0<s\leq 1$}. Let $\mathcal{H}_K$ be the RKHS induced by the kernel $K$ in \eqref{kernel_torus} with
\[
\theta_j=(1+\|j\|^2)^{-\gamma},
\qquad j\in\mathbb{Z}^p,
\]
for some $\gamma>(p+1)/2$, and define
 its unit ball by $
    \mathcal{K} := \{f\in \mathcal{H}_K: \|f\|_{\mathcal{H}_K} \leq 1\}.
$

\begin{enumerate}
\item[\textnormal{(i)}] \textbf{Tanh networks.}
For every $M\in \NN$ with $M>6^{\frac{5p+s(4\gamma -p-1
)}{5p(1-s)+s(4\gamma -1)}}$, let
\[
m = \left\lceil M^{\frac{s}{\frac{5p}{2}+s(2\gamma-\frac{p}{2}-\frac{1}{2})}}\right\rceil,
\qquad
N=m^p
\]
and let $\bar t\subset\mathbb{T}^p$ be the equispaced product grid in \eqref{torus_points} corresponding to this choice of $m$. Then there exists a two-hidden-layer tanh neural network
$\widehat G:\mathbb{R}^N\to\mathbb{R}$
with widths at most
$N(M-1)$ and 
$3\left\lceil\frac{N+1}{2}\right\rceil(5M)^N$,
respectively, such that
\[
\sup_{f\in\mathcal K}
\left|
F(f)-\widehat G\big(f(\bar t)\big)
\right|
\leq
C_{\gamma,p,s,F}
M^{-\frac{s^2}{5p+s(4\gamma-p-1)}},
\]
where $ C_{\gamma,p,s,F}$ is a constant depending on $\gamma,p,s$ and the H\"{o}lder constant $C_F$.
\second{
\item[\textnormal{(ii)}] \textbf{ReLU networks.}
For $M,L\in\mathbb{N}$, define
\[
\beta
:=
\frac p2+s\left(2\gamma-\frac p2-1\right),
\qquad
C_0
:=
\left(
\frac{sp}{\beta+s/2}
\right)^{1/p}.
\]
Let
\[
m
=
\left\lfloor
C_0
\left(
\frac{\log(ML)}{\log\log(ML)}
\right)^{1/p}
\right\rfloor,
\qquad
N=m^p,
\]
and let $\bar t\subset\mathbb{T}^p$ be the corresponding equispaced product grid in \eqref{torus_points}.
Choose $A_0>e$ sufficiently large so that
$\frac{A_0}{\log A_0}
\geq
\left(\frac{2}{C_0}\right)^p$.
Then, for $ML>e^{A_0}$, there exists a ReLU neural network
$\widehat G:[-1,1]^N\to\mathbb{R}$
with depth at most
$$11L+2N+18$$
and width at most
\[
3^{N+3}
\max\left\{
N\left\lfloor M^{1/N}\right\rfloor,
M+2
\right\},
\]
such that
\[
\sup_{f\in\mathcal K}
\left|
F(f)-\widehat G\big(f(\bar t)\big)
\right|
\leq
C_{\gamma,p,s,F}
\left(
\frac{\log\log(ML)}
{\log(ML)}
\right)^{\frac{s}{2p}}.
\]}
\end{enumerate}
\end{theorem}

\second{
\subsection{Comparison of Parameter Complexity Across Spectral-Decay Regimes}
To illustrate how the spectral decay of the kernel affects the approximation result,
Table~\ref{tab:torus-functional-approximation} summarizes the number of network parameters required to achieve a prescribed accuracy $\varepsilon>0$. These bounds are obtained from Theorems~\ref{thm:aj} and \ref{thm:1+j} by choosing the network parameters $M$ and $L$ so that the corresponding approximation error is at most $\varepsilon$, and then substituting these choices into the network-size bounds. We provide a detailed derivation for polynomially decaying Fourier coefficients in Appendix~\ref{app:parameter-complexity}; the exponentially decaying case follows analogously.}

 \begin{table}[h]
\centering
\normalsize
\renewcommand{\arraystretch}{1.8}
\setlength{\tabcolsep}{8pt}
\caption{\second{Required numbers of network parameters for approximating functionals on RKHSs over $\mathbb T^p$
with exponential and polynomial Fourier coefficient decay to accuracy $\varepsilon$.}}
\label{tab:torus-functional-approximation}
\second{
\begin{tabularx}{\textwidth}{|>{\centering\arraybackslash}p{0.34\textwidth}|
                                >{\centering\arraybackslash}X|
                                >{\centering\arraybackslash}X|}
\hline
\multirow{2}{*}{\textbf{Fourier coefficients}}
& \multicolumn{2}{c|}{\textbf{Number of  network parameters}} \\
\cline{2-3}
& \textbf{tanh network}
& \textbf{ReLU network} \\
\hline

\makecell[c]{
$\displaystyle \theta_j = a^{\|j\|}$,
$\displaystyle a\in(0,1)$
}
&
\makecell[c]{
$\displaystyle
\exp\left\{
\mathcal O\left(
\varepsilon^{-\frac{2(p+1)}{s}}
\right)
\right\}$
}
&
\makecell[c]{
$\displaystyle
\exp\left\{
\mathcal O\left(
\varepsilon^{-\frac{2(p+1)}{s}}
\right)
\right\}$
}
\\
\hline

\makecell[c]{
$\displaystyle \theta_j = \left(1+\|j\|^2\right)^{-\gamma}$,
$\displaystyle \gamma>\frac{p+1}{2}$
}
&
\makecell[c]{
$\displaystyle
\exp\left\{
\mathcal O\left(
\varepsilon^{-\frac{2p}{s}}
\log\frac{1}{\varepsilon}
\right)
\right\}$
}
&
\makecell[c]{
$\displaystyle
\exp\left\{
\mathcal O\left(
\varepsilon^{-\frac{2p}{s}}
\log\frac{1}{\varepsilon}
\right)
\right\}$
}
\\
\hline
\end{tabularx}}
\end{table}

\second{
\paragraph{Effect of spectral decay.}
Table~\ref{tab:torus-functional-approximation} shows that exponentially
decaying Fourier coefficients lead to the parameter complexity
\[
\exp\left\{
\mathcal O\left(
\varepsilon^{-\frac{2(p+1)}{s}}
\right)
\right\},
\]
whereas polynomially decaying Fourier coefficients lead to
\[
\exp\left\{
\mathcal O\left(
\varepsilon^{-\frac{2p}{s}}
\log\frac{1}{\varepsilon}
\right)
\right\}.
\]
Since
\[
\varepsilon^{-\frac{2p}{s}}
\log\frac{1}{\varepsilon}
=
o\left(
\varepsilon^{-\frac{2(p+1)}{s}}
\right)
\qquad\text{as }\varepsilon\to0,
\]
the required parameter-complexity bound is asymptotically smaller for
the polynomial-decay case. In both cases, the complexity increases with
the torus dimension $p$ and decreases as the H\"older regularity $s$ of
the functional increases.

Without entering into the technical details here, we briefly explain the
mechanism behind this difference. The parameter complexity is
governed by two kernel-dependent quantities: the kernel interpolation
error, which determines how many sampling points are needed, and the regularity of the finite-dimensional function induced by the sampling points, which
determines how large the neural network must be. For both spectral-decay
regimes, attaining accuracy $\varepsilon$ requires a grid resolution
$m=\mathcal O(\varepsilon^{-2/s})$ and hence
$N=m^p=\mathcal O(\varepsilon^{-2p/s})$ sampling points. The difference
arises from the second factor.
When $\theta_j=a^{\|j\|}$, the H\"older constant of the
finite-dimensional function grows exponentially with $m$, up to
polynomial factors. This results in a sufficient total parameter count
of
$\exp\left\{\mathcal O\left(m^{p+1}\right)\right\}$.
By contrast, when
$\theta_j=(1+\|j\|^2)^{-\gamma}$, the H\"older constant grows only
polynomially with $m$, resulting in a sufficient total parameter count
of
$\exp\left\{\mathcal O\left(m^p\log m\right)\right\}$.

Thus, although exponential Fourier decay corresponds to a smoother and
more restrictive RKHS unit ball, our construction yields a larger
sufficient parameter bound for this case. In Section~\ref{sec:mainresultII}, we introduce a general approximation
framework and decompose the error into a finite-dimensional
representation error and a neural-network approximation error. This
decomposition makes precise how the kernel interpolation accuracy and
the regularity of the induced finite-dimensional function jointly
determine the rates reported here.

\paragraph{Effect of the activation function.}
For each spectral-decay regime, tanh and ReLU networks attain the same leading-order parameter complexity. This indicates that the dominant dependence on $\varepsilon$ is determined by the sampling resolution and the regularity of the induced finite-dimensional function rather than by the activation function itself. }

\section{Main Results III: Approximation of Functionals on General RKHSs}
\label{sec:mainresultII}

In this section, we give our results on approximating functionals on an RKHS induced by a general Mercer kernel.

Let $(\mathcal{X},d_{\mathcal{X}})$ be a compact metric space,
and let $\mathcal{H}_K$ be the RKHS induced by a Mercer kernel
$K:\mathcal{X}\times\mathcal{X}\to\mathbb{R}$. We assume that $K$ is
$\alpha$-Hölder continuous in its second argument: for some
$\alpha\in(0,1]$ and $C_K\geq 0$,
\begin{equation}
\label{reg_K}
    |K(u,v)-K(u,\widetilde v)|
    \leq C_K d_{\mathcal{X}}(v,\widetilde v)^\alpha,
    \qquad u,v,\widetilde v\in\mathcal{X}.
\end{equation}

In $\mathcal{X}$, we choose a set of sampling locations $\Bar{t}=\{t_i\}_{i=1}^N \subset \mathcal{X}$ for $N\in\NN$ such that its Gram matrix \begin{equation}\label{gram}
    K[\Bar{t} ] = (K(t_i, t_j))_{i,j=1}^N 
    =  \begin{bmatrix}
    K(t_1, t_1) & K(t_1, t_2) & \cdots & K(t_1, t_N)\\
    K(t_2, t_1) & K(t_2, t_2) & \cdots & K(t_2, t_N)\\
    \vdots & \vdots & \ddots & \vdots\\
    K(t_N, t_1) & K(t_N, t_2) & \cdots & K(t_N, t_N)
      \end{bmatrix}
\end{equation} is invertible. 
 The sampling locations need not be
uniformly distributed; their geometric quality is described by the fill
distance
\begin{equation}
\label{fill}
    h_{\bar t}
    :=h_{\bar t,\mathcal{X}}
    :=\sup_{x\in\mathcal{X}}
      \min_{1\leq i\leq N}d_{\mathcal{X}}(x,t_i),
\end{equation}
which measures the radius of the largest unsampled region of
$\mathcal{X}$ (thus indicates how well $\Bar{t}$ fill out the domain $\mathcal{X}$). A sequence of point sets becomes dense in $\mathcal{X}$ when
$h_{\bar t}\to 0$ as $N\to\infty$.

We also define the separation radius
\begin{equation}
\label{separation}
    q_{\bar t}
    :=\frac{1}{2}
      \min_{\substack{1\leq i,j\leq N\\i\neq j}}
      d_{\mathcal{X}}(t_i,t_j),
\end{equation}
which quantifies the minimum spacing between distinct sampling locations.
Whereas a small fill distance ensures good coverage of the domain, a
separation radius of comparable order prevents the points from becoming
excessively clustered.

A central quantity in kernel interpolation is the power function. For the
point set $\bar t$, the power function is defined as \citep{fasshauer2005meshfree, pazouki2011bases, kanagawa2018gaussian}
\begin{align}
\label{power}
    \epsilon_K(\bar t)
    &:=
    \sup_{x\in\mathcal{X}}
    \min_{c\in\mathbb{R}^N}
    \left\|K_x-\sum_{i=1}^N c_iK_{t_i}\right\|_{\mathcal{H}_K}
    \nonumber\\
    &=
    \sup_{x\in\mathcal{X}}
    \min_{c\in\mathbb{R}^N}
    \left(
        K(x,x)
        -2\sum_{i=1}^N c_iK(x,t_i)
        +\sum_{i,j=1}^N c_ic_jK(t_i,t_j)
    \right)^{1/2},
\end{align}
where $K_x:=K(\cdot,x)$. This quantity measures how well the kernel section
$K_x$ can be approximated by
$\operatorname{span}\{K_{t_1},\ldots,K_{t_N}\}$. Equivalently, it controls
the worst-case uniform error of kernel interpolation over the unit ball of
$\mathcal{H}_K$.

 The following theorem presents the result of approximating functional on RKHS induced by a general Mercer kernel. For
$f\in\mathcal{H}_K$, we write
\[
    f(\bar t):=(f(t_1),\ldots,f(t_N))^\top.
\]

\begin{theorem}[Approximation of functionals on general RKHSs]
\label{thm:general}
Let $d\in\mathbb{N}$, $\mathcal{X}=[0,1]^d$, and
\[
    \mathcal{K}
    :=\left\{
        f\in\mathcal{H}_K:
        \|f\|_{\mathcal{H}_K}\leq 1
    \right\},
\]
where $\mathcal{H}_K$ is induced by a Mercer kernel $K$ satisfying
\eqref{reg_K} for some $\alpha\in(0,1]$ and $C_K\geq 0$. Suppose
Assumption~\ref{assump:holder-F} holds for some $s\in(0,1]$ and
$C_F\geq 0$.

Take $\bar t=\{t_1,\ldots,t_N\}\subset\mathcal{X}$ such that
$K[\bar t]$ is invertible, and define
\begin{equation}
\label{eq:general_CG}
    C_G
    :=
    C_F
    \left(
        1+
        \|K[\bar t]^{-1}\|_{\mathrm{op}}
        \sqrt{N}\,C_K h_{\bar t}^{\alpha}
    \right)^s.
\end{equation}
Then the following statements hold.

\begin{enumerate}
    \item[\textnormal{(i)}]
    \textbf{Tanh networks.}
    For every $M\in\mathbb{N}$ satisfying $M>5N^2$, there exists a
    tanh network $\widehat G_{\mathrm{tanh}}:\mathbb{R}^N\to\mathbb{R}$
    with two hidden layers of widths at most
    \[
        N(M-1)
        \qquad\text{and}\qquad
        3\left\lceil\frac{N+1}{2}\right\rceil(5M)^N,
    \]
    respectively, such that
    \begin{equation}
    \label{thm:general_bound_tanh}
        \sup_{f\in\mathcal{K}}
        \left|
            F(f)-\widehat G_{\mathrm{tanh}}\bigl(f(\bar t)\bigr)
        \right|
        \leq
        C_F\epsilon_K(\bar t)^s
        +
        \frac{(14N^2+1)\sqrt{N}\,C_G}{M^s}.
    \end{equation}

    \item[\textnormal{(ii)}]
   \second{ \textbf{ReLU networks.}
    For every $M,L\in\mathbb{N}$, there exists a ReLU network
    $\widehat G_{\mathrm{ReLU}}:\mathbb{R}^N\to\mathbb{R}$ with depth
    \[
        11L+2N+18
    \]
    and width
    \[
        3^{N+3}
        \max\left\{
            N\left\lfloor M^{1/N}\right\rfloor,
            M+2
        \right\},
    \]
    such that
    \begin{equation}
    \label{thm:general_bound_relu}
        \sup_{f\in\mathcal{K}}
        \left|
            F(f)-\widehat G_{\mathrm{ReLU}}\bigl(f(\bar t)\bigr)
        \right|
        \leq
        C_F\epsilon_K(\bar t)^s
        +
        131\cdot 2^s\sqrt{N}\,C_G
        \left(
            M^2L^2\log_3(M+2)
        \right)^{-s/N}.
    \end{equation}}
\end{enumerate}
\end{theorem}

The bounds in \eqref{thm:general_bound_tanh} and
\eqref{thm:general_bound_relu} share the same error decomposition. The first
term,
    $C_F\epsilon_K(\bar t)^s$,
is the  error incurred by replacing an input function with its
kernel interpolant based on the point evaluations $f(\bar t)$. This term is
independent of the neural network architecture and generally decreases as
the sampling set becomes denser and the power function
$\epsilon_K(\bar t)$ tends to zero. 

The second term is the error of
approximating the resulting finite-dimensional map by a neural network. It
decreases as the network-size parameters increase, although its precise
form differs between the tanh and ReLU architectures.

Both bounds therefore exhibit the same three-way trade-off among sampling
resolution, numerical conditioning, and network size. Increasing $N$ and
choosing a denser sampling set generally reduce the interpolation error.
At the same time, closely spaced sampling locations may make the Gram matrix
$K[\bar t]$ ill-conditioned. This effect enters 
through
\[
    C_G
    =
    C_F
    \left(
        1+
        \|K[\bar t]^{-1}\|_{\mathrm{op}}
        \sqrt{N}\,C_Kh_{\bar t}^{\alpha}
    \right)^s
\]
and may partially offset the benefit of a smaller fill distance. Moreover,
increasing $N$ raises the dimension of the finite-dimensional map and consequently increases the required network complexity.

Given $\bar t$, the neural network approximation error can be
reduced by increasing $M$ in the tanh case or $M$ and $L$ in the ReLU case.
This improvement, however, comes at the cost of increasing the width or depth
of the network. Thus, neither the sampling resolution nor the network size
should be chosen independently: an effective approximation requires
balancing the decay of $\epsilon_K(\bar t)$, the conditioning of
$K[\bar t]$, and the size of the neural network. For the specific
kernel classes considered in the preceding section, explicit estimates of
the power function and the inverse Gram matrix allow us to quantify this
balance and select the sampling and network size parameters accordingly.

\second{
\section{Statistical Learning of RKHS Functionals}\label{sec:generalization}
The preceding sections established explicit error bounds for approximating RKHS functionals. These results guaranty the existence of neural networks that achieve a target approximation accuracy, but they do not by themselves address whether such functionals can be learned efficiently from finitely many observations. 

In this section, we place the approximation problem in a statistical regression framework, where independent training pairs $\{(f_i,Y_i)\}_{i=1}^n$ are observed and the target is to learn the regression functional $F_\rho(f):=\mathbb{E}[Y\mid f]$.
We study empirical risk minimization over the neural network classes constructed in the preceding approximation results. We establish generalization bounds that separate the approximation error from the estimation error, with the latter governed by the statistical complexity of the corresponding network class.

\subsection{Regression Setting and Empirical Risk Minimization}

Assume $|Y|\leq B$ almost surely
for some $B>0$.
Let $\rho$ be a Borel probability measure on $\mathcal{K}\times [-B,B]$,
and let $(f,Y)\sim\rho$, where $f$ is a function taking values in
$\mathcal{K}$ and $Y$ is a scalar response. We consider the
regression model
\[
Y=F_\rho(f)+\epsilon,
\qquad
\mathbb{E}[\epsilon\mid f]=0,
\]
where  the regression functional $F_\rho:\mathcal{K}\to \RR$ is the conditional expectation given by
\[
F_\rho(f)
:=
\mathbb{E}[Y\mid f]
=
\int_{\mathbb{R}}y \, d\rho(y\mid f),
\qquad f\in\mathcal{K}.
\]
We assume that $F_\rho$ satisfies
Assumption~\ref{assump:holder-F} with some $0<s\leq1$. Since $|Y|\leq B$ almost surely, we know $|F_\rho(f)|\leq B$.

Let
$\{(f_i,Y_i)\}_{i=1}^n$
be independent copies of $(f,Y)\sim\rho$. The functions $f_i$ are not observed over the entire domain. Instead, we observe their values at fixed sampling locations
$\bar t=(t_1,\ldots,t_N)\in\mathcal X^N$,
that is
\[
f_i(\bar t)
=
\bigl(f_i(t_1),\ldots,f_i(t_N)\bigr)^\top
\in\mathbb{R}^N.
\]
Thus, the training data we have on hand are
$\mathcal D_n
:=
\{(f_i(\bar t),Y_i)\}_{i=1}^n$.
Define the deterministic sampling operator
\[
S_N:\mathcal K\to\mathbb R^N,
\qquad
S_Nf=f(\bar t).
\]

We next introduce a general neural network function class on $\mathbb{R}^N$.
Let $\sigma:\mathbb{R}\to\mathbb{R}$ be an activation function applied
componentwise. For a network with $D$ hidden layers, let $\boldsymbol p=(p_0,\ldots,p_D,p_{D+1})$ denote its layer-width vector, where $p_0=N$ is the input dimension, $p_\ell$ is the width of the $\ell$-th hidden layer for $\ell=1,\ldots,D$, and $p_{D+1}=1$ is the output dimension. We place no restrictions on the network parameters and define
\begin{equation}\label{neural_function} \mathcal G_{\sigma,\boldsymbol p} := \left\{ A_{D+1}\circ\sigma\circ A_D\circ\cdots\circ\sigma\circ A_1: \begin{array}{l} A_\ell(x)=W_\ell x+b_\ell,\\ W_\ell\in\mathbb{R}^{p_\ell\times p_{\ell-1}}, \quad b_\ell\in\mathbb{R}^{p_\ell},\\ \ell=1,\ldots,D+1 \end{array} \right\}. \end{equation}
Although $G : \RR^N \to \RR \in \mathcal G_{\sigma,\boldsymbol p}$ is a vector-input neural network, its composition with $S_N$ defines a functional on $\mathcal{K}$ as
\begin{eqnarray*}
    F(f) := G(S_Nf) = G(f(\Bar{t})).
\end{eqnarray*}
We then define our hypothesis space for learning RKHS functionals as
\[
\mathcal H: = \mathcal H_{N,\sigma,\boldsymbol p}
:=
\left\{
F:\mathcal K\to\RR:
F=G\circ S_N,\quad
G\in\mathcal G_{\sigma,\boldsymbol p}
\right\},
\]
which consists of functionals on $\mathcal{K}$ that depend on $f$ only through its $N$ sampled values $f(\bar t)$.

For a measurable functional $F:\mathcal K\to\mathbb R$, define its
population risk  by
\[
\mathcal E(F)
:=
\mathbb E\bigl[(Y-F(f))^2\bigr].
\]
By definition, the regression functional minimizes the population risk. For any $F: \mathcal K \to 
    \RR$, its excess risk relative to $F_\rho$ is 
\[
\mathcal E(F)-\mathcal E(F_\rho)
=
\|F-F_\rho\|_{L^2(\rho_{\mathcal K})}^2,
\]
where $\rho_{\mathcal K}$ denotes the marginal distribution of $f$ on
$\mathcal K$.
The empirical risk associated with $\mathcal D_n$ is
\[
\mathcal E_n(F)
:=
\frac1n\sum_{i=1}^n
\bigl(Y_i-F(f_i)\bigr)^2.
\]
In particular, for $F=G\circ S_N$, we have
$\mathcal E_n(G\circ S_N)
=
\frac1n\sum_{i=1}^n
\bigl(Y_i-G(f_i(\bar t))\bigr)^2$,
which depends only on the observed data.

Let
$\pi_B(u):=\max\{-B,\min\{u,B\}\}$
and define the clipped hypothesis class
\begin{equation}\label{eqn:clip}
 \pi_B \mathcal H:= 
\pi_B\mathcal H_{N,\sigma,\boldsymbol p}
:=
\left\{
\pi_B\circ G\circ S_N:
G\in\mathcal G_{\sigma,\boldsymbol p}
\right\}.
\end{equation}
Since $|F_\rho(f)|\leq B$, we know $|\pi_B (F(f)) - F_\rho(f)| \leq |F(f) - F_\rho(f)|$.
We define the empirical risk minimizer by
\[
\widehat F_{\mathcal D_n}
\in
\operatorname*{arg\,min}_{F\in
\pi_B\mathcal H}
\mathcal E_n(F).
\]
Equivalently, if
\[
\widehat G_{\mathcal D_n}
\in
\operatorname*{arg\,min}_{G\in\mathcal G_{\sigma,\boldsymbol p}}
\frac1n\sum_{i=1}^n
\bigl(Y_i-\pi_B(G(f_i(\bar t)))\bigr)^2,
\]
then
\begin{equation}\label{clipERM}
\widehat F_{\mathcal D_n}
=
\pi_B\circ\widehat G_{\mathcal D_n}\circ S_N.
\end{equation}
Our objective is to bound
\[
\mathcal E(\widehat F_{\mathcal D_n})-\mathcal E(F_\rho) = \|\widehat F_{\mathcal D_n}-F_\rho\|_{L^2(\rho_{\mathcal K})}^2.
\]
}

\second{
For notational convenience, write
\[
\Delta(F):=\mathcal E(F)-\mathcal E(F_\rho),
\qquad
\Delta_n(F):=\mathcal E_n(F)-\mathcal E_n(F_\rho).
\]
We consider the general clipped function class $\pi_B\mathcal H\subseteq\{F:\mathcal K\to[-B,B]\}$ given in \eqref{eqn:clip}. 
Let $\widehat F_{\mathcal D_n}$ be an empirical risk minimizer in this clipped class and let the offset complexity of $\pi_B\mathcal H$ be
\begin{equation}\label{eq:offset-complexity}
\mathcal C_{n}(\pi_B\mathcal H)
:=
\mathbb E_{\mathcal D_n}
\left[
\Delta(\widehat F_{\mathcal D_n})-2\Delta_n(\widehat F_{\mathcal D_n})
\right].
\end{equation}

\begin{proposition}[Oracle inequality]
\label{thm:oracle-point-evaluation}
Suppose that $|Y|\leq B$ almost surely. Let $\widehat F_{\mathcal D_n}$ be an empirical risk
minimizer over $\pi_B\mathcal H$. Then
\begin{equation}
\label{eq:abstract-oracle-inequality}
\mathbb E_{\mathcal D_n}
\|\widehat F_{\mathcal D_n}-F_\rho\|_{L^2(\rho_{\mathcal K})}^2
\leq  \mathcal C_{n}(\pi_B\mathcal H) +
2\inf_{F\in \pi_B\mathcal H}
\|F-F_\rho\|_{L^2(\rho_{\mathcal K})}^2.
\end{equation}
\end{proposition}

\noindent
{\bf Proof}.
Consider the error decomposition
$$\Delta(\widehat F_{\mathcal D_n})
=
\Delta(\widehat F_{\mathcal D_n})
-2\Delta_n(\widehat F_{\mathcal D_n})
+2\Delta_n(\widehat F_{\mathcal D_n}).$$
Since  $\Delta_n(\widehat F_{\mathcal D_n}) \leq \Delta_n(\pi_B\circ\widehat G_{\mathcal D_n}\circ S_N)$ from \eqref{clipERM}, we have
\begin{eqnarray*}
\mathbb E_{\mathcal D_n} [2 \Delta_n(\widehat F_{\mathcal D_n})] 
\leq 2 E_{\mathcal D_n}\left[\inf_{F \in \pi_B\mathcal H} \Delta_n(F)\right]
\leq 2 \inf_{F \in \pi_B\mathcal H} E_{\mathcal D_n} [\Delta_n(F)]
\leq 2 \inf_{F \in \pi_B\mathcal H} \left\|
F-F_\rho
\right\|_{L^2(\rho_{\mathcal K})}^2.
\end{eqnarray*}
The statement follows directly after.\hfill \BlackBox}

\second{
\subsection{Learning Rates for Functionals on a Gaussian RKHS}
\label{subsec:gaussian-learning-rates}

We now combine the oracle inequality above with the explicit
approximation bounds for functionals on Gaussian RKHSs established in
Theorem~\ref{thm:Gaussian}. Recall that $n$ denotes the number of
independent training pairs. For illustration, we present explicit learning rates for learning functionals
on the Gaussian RKHS; the corresponding results for the other RKHSs considered
in this paper follow by an analogous argument.

\begin{theorem}[Learning rates for Gaussian RKHS functionals]
\label{thm:gaussian-learning-rates}
    Consider the ReLU network architecture in
Theorem~\ref{thm:Gaussian}\textup{(ii)} with
\[
L=1,
\qquad
M=M_n:=\left\lceil n^\alpha\right\rceil,
\qquad
0<\alpha<\frac12,
\]
and choose $m$ and $N$ accordingly as specified in that theorem. Let
$\widehat F_{\mathcal D_n}$ be an empirical risk minimizer over the
corresponding clipped ReLU class.
Let
C$_0
=
\left(
\frac{1}{2^{d-1}\sigma^2\pi^2d}
\right)^{\frac{1}{d+2}}$.
Then, for any 
\[
0<\kappa<
\frac{csC_0}{2(d+2)\sqrt d}\,
\alpha^{\frac{1}{d+2}},
\]
we have
\[
\mathbb E_{\mathcal D_n}
\left\|
\widehat F_{\mathcal D_n}-F_\rho
\right\|_{L^2(\rho_{\mathcal K})}^2
\leq
C
\exp\left\{
-\kappa
(\log n)^{\frac{1}{d+2}}\log\log n
\right\},
\]
where $C>0$ is independent of $n$.
\end{theorem}}

\second{The proof of Theorem \ref{thm:gaussian-learning-rates} is given at the end of Appendix \ref{sec:appendB}.
This theorem shows that empirical risk minimization over the proposed
ReLU class consistently learns the regression functional on Gaussian RKHS. The
convergence is faster than any negative power of $\log n$, although
slower than any polynomial rate in $n$. 
The overall learning rate is governed by
the approximation rate for Gaussian RKHS functionals established in
Theorem~\ref{thm:Gaussian}\textup{(ii)}.

Since the generalization error is measured in the
$L^2(\rho_{\mathcal K})$ norm, it may at first seem natural to assume
that the regression functional $F_\rho$ is $s$-H\"older continuous with
respect to the same norm. 
In our setting, $\rho_{\mathcal K}$ is unspecified; we only assume that it is a Borel probability measure.
As previously discussed in Remark~\ref{rem:holder}, \[ \sup_{\mu\in\mathcal P(\mathcal X)} \|f-\widetilde f\|_{L^2(\mu)} = \|f-\widetilde f\|_\infty, \] where $\mathcal P(\mathcal X)$ denote the collection of Borel probability measures on $\mathcal X$. Consequently, Assumption~\ref{assump:holder-F} can equivalently be written as \[ \bigl|F_\rho(f)-F_\rho(\widetilde f)\bigr| \leq C_{F_\rho} \sup_{\mu\in\mathcal P(\mathcal X)} \|f-\widetilde f\|_{L^2(\mu)}^s, \qquad f,\widetilde f\in\mathcal K. \] Thus, the supremum-norm assumption is an $s$-H\"older condition with respect to the worst-case $L^2(\mu)$ discrepancy over all Borel measures on $\mathcal X$ and does not require the choice of a particular reference measure.}

\section{Overview of Method and  Proof Strategy } \label{sec:method}

This section outlines our proposed method and the key proof ideas underlying the main approximation results.

After rescaling the kernel, we may assume without loss of generality that
    $\sup_{x\in\mathcal{X}} K(x,x)\leq 1$.
By the reproducing property, this implies
$\|f\|_{\infty}\leq \|f\|_{\mathcal{H}_K}$ for every
$f\in\mathcal{H}_K$.

The construction and analysis of our operator-learning framework are based on three main ideas:
\begin{enumerate}
    \item projection of functions $\in \mathcal{H}_K$  onto finite-dimensional space via kernel interpolation;
    \item estimates of the power function, which controls the the uniform error between a function and its projection; and
    \item  approximation of H\"{o}lder continuous function by classical neural networks.
\end{enumerate}

Let $\Bar{t}=\{t_i\}_{i=1}^N\subset\mathcal{X}$ denote a set of sampling points. We begin with the error decomposition
\begin{eqnarray}\label{decomposition}
         \sup_{f\in \mathcal{K}}\left| F(f) - \widehat{G}(f(\Bar{t}))\right| &\leq& 
         \underbrace{\sup_{f\in \mathcal{K}}| F(f) - F(Pf)|}_{\text{error term (I)}} +  \underbrace{\sup_{f\in \mathcal{K}}\left|F(Pf) - \widehat{G}(f(\Bar{t}))\right|}_{\text{error term (II)}}, 
\end{eqnarray}
where $P: \mathcal{H}_K \rightarrow \mathcal{H}_K$ is a projection operator introduced below. The first term measures the error incurred by replacing $f$ with its kernel interpolant $Pf$ and is controlled by the power function. To control the second term, we express $F(Pf)$ as a function of the finite-dimensional vector $f(\Bar{t})$, estimate the H\"{o}lder regularity of this function using $\|(K[\Bar{t}])^{-1}\|_{op}$, and then apply neural-network approximation theory.

\subsection{Kernel Interpolation via Nodal Functions}\label{subsec:nodal}

Choose $\Bar{t}=\{t_i\}_{i=1}^N\subset\mathcal{X}$ such that the Gram matrix
\[
    K[\Bar{t}]=\bigl(K(t_i,t_j)\bigr)_{i,j=1}^N
\]
is invertible. The associated nodal functions $\{\psi_i\}_{i=1}^N$ are defined by
\begin{equation}\label{psi}
    \begin{bmatrix}
        \psi_1\\
        \vdots\\
        \psi_N
    \end{bmatrix}
    =
    \bigl(K[\Bar{t}]\bigr)^{-1}
    \begin{bmatrix}
        K_{t_1}\\
        \vdots\\
        K_{t_N}
    \end{bmatrix}.
\end{equation}
By construction, $\psi_i(t_j)=\delta_{ij}$. We can check that the set $\{\psi_i\}_{i=1}^N$ are linearly independent.
With the chosen $\Bar{t}$, we define a projection operator $P: \mathcal{H}_K \rightarrow \mathcal{H}_K$ as 
\begin{equation}
    Pf:=\sum_{i=1}^N f(t_i)\psi_i,
    \qquad f\in\mathcal{H}_K.
\end{equation}
The following lemma records the interpolation and projection properties of $P$.

\begin{lemma}\label{lemma:ortho}
For every $f\in\mathcal{H}_K$ and every
$\Bar{t}\subset\mathcal{X}$ for which $K[\Bar{t}]$ is invertible,
\begin{enumerate}
    \item $Pf(t_\ell)=f(t_\ell)$ for $\ell=1,\ldots,N$; \label{item1}
    \item $\langle f-Pf,K_{t_\ell}\rangle_{\mathcal{H}_K}=0$ for $\ell=1,\ldots,N$. \label{item2}
\end{enumerate}
Consequently, $P$ is the orthogonal projection onto  finite-dimensional space $\text{span}\{K_{t_i}\}_{i=1}^N$.
\end{lemma}

In particular, $\|Pf\|_{\mathcal{H}_K}\leq
\|f\|_{\mathcal{H}_K}$; hence $Pf\in\mathcal{K}$ whenever
$f\in\mathcal{K}$, so $F(Pf)$ is well-defined. The proof of
Lemma~\ref{lemma:ortho} is provided in Appendix \ref{subsec:project}.

The quality of the interpolation is measured by the power function given previously in (\ref{power}):
\begin{eqnarray*}
     \epsilon_K(\Bar{t})&:=& \max_{x\in \mathcal{X}} \min_{c\in\RR^N} \left\|K_x -  \sum_{i=1}^N c_i  K_{t_i}\right\|_{\mathcal{H}_K} 
\end{eqnarray*}
The following estimate (see, for example, \citep[Section~6.3]{cucker2007learning}) connects the power function to the uniform interpolation error.

\begin{lemma}\label{f-Pf}
For every $f\in\mathcal{H}_K$,
\begin{equation}
    \|f-Pf\|_\infty
    \leq
    \|f\|_{\mathcal{H}_K}\,\epsilon_K(\Bar{t}).
\end{equation}
\end{lemma}

For completeness, the proof of Lemma~\ref{f-Pf} is given in
Appendix~\ref{subsec:project}. Since $F$ is $s$-H\"{o}lder continuous with constant $C_F$ by Assumption~\ref{assump:holder-F}, Lemma~\ref{f-Pf} implies, for every $f\in\mathcal{K}$,
\[
    |F(f)-F(Pf)|
    \leq C_F\|f-Pf\|_\infty^s
    \leq C_F\bigl(\epsilon_K(\Bar{t})\bigr)^s.
\]
Therefore,
\begin{equation}\label{eq:firstterm}
    \sup_{f\in\mathcal{K}}|F(f)-F(Pf)|
    \leq C_F\bigl(\epsilon_K(\Bar{t})\bigr)^s.
\end{equation}
This controls error term~\textup{(I)} in \eqref{decomposition} and gives the first term in the general approximation bound \eqref{decomposition}. It remains to choose the sampling set $\Bar{t}$ and estimate $\epsilon_K(\Bar{t})$ for each kernel class under consideration.

\subsection{Estimation of the Power Function}\label{subsec:power}

The preceding argument reduces the control of error term~\textup{(I)}
in \eqref{decomposition} to estimating the power function
$\epsilon_K(\Bar{t})$. These estimates can be expressed in terms of
the fill distance
\begin{equation}\label{fill}
    h_{\Bar{t}}
    :=
    \sup_{x\in\mathcal{X}}
    \min_{t\in\Bar{t}} d_{\mathcal{X}}(x,t),
\end{equation}
which measures how well the sampling set $\Bar{t}$ covers
$\mathcal{X}$.

The following estimate applies to both the Euclidean and toroidal
settings considered in this paper.
\begin{lemma}\label{lemma:power}
Suppose that $K$ is an $\alpha$-H\"{o}lder continuous Mercer kernel
with $\alpha\in(0,1]$ and H\"{o}lder constant $C_K$. For any sampling
set $\Bar{t}\subset\mathcal{X}$ such that $K[\Bar{t}]$ is invertible,
we have
\begin{equation}
    \epsilon_K(\Bar{t})
    \leq
    \sqrt{2C_K}\,h_{\Bar{t}}^{\alpha/2}.
\end{equation}
Consequently,
\begin{equation}
    \|f-Pf\|_\infty
    \leq
    \sqrt{2C_K}\,h_{\Bar{t}}^{\alpha/2}
    \|f\|_{\mathcal{H}_K},
    \qquad f\in\mathcal{H}_K.
\end{equation}
\end{lemma}

The proof of Lemma~\ref{lemma:power} is provided in
Appendix~\ref{subsec:project}.  It remains only
to estimate the fill distance of the sampling sets used in our main
results.
\paragraph{Euclidean domain.}
For $\mathcal{X}=[0,1]^d$, we use the uniform grid
\begin{equation*}
    \Bar{t}
    =
    \left\{0,\frac{1}{m},\ldots,\frac{m}{m}\right\}^{d},
    \qquad N=(m+1)^d.
\end{equation*}
This is the sampling set used in Theorems~\ref{thm:Sobolev},
\ref{thm:multi}, and \ref{thm:Gaussian}. 
Its fill distance satisfies
\begin{equation}\label{fill_tbar}
    h_{\Bar{t}}\leq\frac{\sqrt{d}}{m}.
\end{equation}
It therefore follows from Lemma~\ref{lemma:power} that
\begin{equation}\label{power_euclidean}
    \epsilon_K(\Bar{t})
    \leq
    \sqrt{2C_K}
    \left(\frac{\sqrt{d}}{m}\right)^{\alpha/2}, \quad \hbox{and hence} \quad \|f-Pf\|_\infty
    \leq
    \sqrt{2C_K}
    \left(\frac{\sqrt{d}}{m}\right)^{\alpha/2}
    \|f\|_{\mathcal{H}_K}.
\end{equation}

\paragraph{Toroidal domain.}
For $\mathcal{X}=\mathbb{T}^p$, we use the uniform grid
\begin{eqnarray}  \bar{t}&=&\left(t_k:=e^{i\frac{2\pi}{m}k}\right)_{k\in \{0,1,\ldots, m-1\}^p} \nonumber\\
     &=&\left( e^{i\frac{2\pi}{m}k_1}, \cdots, e^{i\frac{2\pi}{m}k_p}\right), \qquad \text{for } k_1,\ldots, k_p \in \{0,1,\ldots, m-1\}, 
\end{eqnarray}
which contains a total of $N=m^p$ sample points. 
We equip $\mathbb{T}^1$ with the geodesic metric
\begin{equation*}
    d_{\mathbb{T}^1}(e^{i\xi},e^{i\eta})
    :=
    \min_{\ell\in\mathbb{Z}}
    |\xi-\eta+2\pi\ell|
\end{equation*}
and $\mathbb{T}^p$ with the corresponding product metric
    $d_{\mathbb{T}^p}(u,v)
    :=
    \left(
        \sum_{k=1}^p
        d_{\mathbb{T}^1}(u_k,v_k)^2
    \right)^{1/2}$.
Each coordinate of a point in $\mathbb{T}^p$ lies within
$\pi/m$ of a grid point. Therefore,
\begin{equation}\label{fill_tbar_torus}
    h_{\Bar{t}}
    \leq
    \frac{\sqrt{p}\pi}{m}.
\end{equation}
Applying Lemma~\ref{lemma:power}, we obtain
\begin{equation}\label{power_torus}
    \epsilon_K(\Bar{t})
    \leq
    \sqrt{2C_K}
    \left(\frac{\sqrt{p}\pi}{m}\right)^{\alpha/2}, \quad \hbox{and hence} \quad  \|f-Pf\|_\infty
    \leq
    \sqrt{2C_K}
    \left(\frac{\sqrt{p}\pi}{m}\right)^{\alpha/2}
    \|f\|_{\mathcal{H}_K}.
\end{equation}
Thus, the Euclidean and toroidal cases follow from the same general
power-function estimate; the geometry of the underlying domain enters
only through the corresponding fill distance.

\subsection{The Finite-Dimensional Surrogate $G$ and Its Regularity}
\label{subsec:regularity_G}

We now turn to error term~\textup{(II)} in
\eqref{decomposition}. Define the reconstruction map
\begin{equation*}
    R_{\Bar t}c
    :=\sum_{i=1}^N c_i\psi_i,
    \qquad c=(c_1,\ldots,c_N)\in[-1,1]^N,
\end{equation*}
and the finite-dimensional function
\begin{equation}\label{G}
    G(c):=F(R_{\Bar t}c)
    =F\left(\sum_{i=1}^N c_i\psi_i\right).
\end{equation}
For every $f\in\mathcal{K}$,  we know 
$f(\Bar{t})\in[-1,1]^N$, and the interpolation identity gives
\begin{equation*}
    G(f(\Bar t))=F(Pf).
\end{equation*}
Consequently, error term~\textup{(II)} is bounded by
\begin{equation}\label{eq:second_term_reduction}
    \sup_{f\in\mathcal K}
    \left|F(Pf)-\widehat G(f(\Bar t))\right|
    \leq
    \|G-\widehat G\|_{L^\infty([-1,1]^N)}.
\end{equation}
Thus, controlling error term~\textup{(II)} reduces to approximating
the finite-dimensional function $G$. The next lemma quantifies the
regularity of $G$.

\begin{lemma}[Regularity of $G$]\label{lemma:reg_G}
Suppose that Assumption~\ref{assump:holder-F} holds with exponent
$s\in(0,1]$ and constant $C_F$. Then $G$  defined in \eqref{G} is $s$-H\"{o}lder
continuous on $[-1,1]^N$ with respect to the Euclidean norm, with
H\"{o}lder constant
\begin{equation}\label{eq:C_G_general}
    C_G
    =C_F\left(
        1+
        \bigl\|(K[\Bar t])^{-1}\bigr\|_{op}
        \sqrt N\,C_K h_{\Bar t}^{\alpha}
    \right)^s.
\end{equation}
\end{lemma} 
The proof of Lemma~\ref{lemma:reg_G} is given in Appendix~\ref{subsec:regularity}. Since
\begin{equation*}
    \bigl\|(K[\Bar t])^{-1}\bigr\|_{op}
    =\frac{1}{\lambda_{\min}(K[\Bar t])},
\end{equation*}
estimating $C_G$ reduces to finding a lower bound for the smallest
eigenvalue of the Gram matrix $K[\Bar t]$. For translation-invariant kernels on
$[0,1]^d$, this is achieved through the Fourier transform of the
kernel; for kernels on $\mathbb T^p$, it is achieved through their
Fourier coefficients. The precise spectral estimates, the resulting
kernel-specific constants, and their proofs are provided in
Appendix~\ref{subsec:regularity} and
Appendix~\ref{subsec:kernel_specific_regularity}.

Having established the regularity of $G$, we next approximate it  using either tanh
or ReLU neural networks. The tanh result below extends the
approximation theorem of \citet{de2021approximation}, originally
established for Lipschitz continuous functions, to
$s$-H\"{o}lder continuous functions with $s\in(0,1]$. The ReLU
result is a direct restatement of a result in \citet{shen2022optimal}. The proof of the tanh extension
is provided in Appendix~\ref{sec:appendB}.

\begin{theorem}
\label{thm:extension_tim}
Let $G:[-1,1]^N\to\mathbb{R}$ be $s$-H\"{o}lder continuous with
exponent $s\in(0,1]$ and H\"{o}lder constant $C_G$.

\begin{enumerate}
    \item[\textnormal{(i)}]
    \textbf{Tanh networks.}
    For every $M\in\mathbb{N}$ satisfying $M>5N^2$, there exists a
    tanh network $\widehat{G}:[-1,1]^N\to\mathbb{R}$ with two hidden
    layers of widths at most
    \[
        N(M-1)
        \qquad\text{and}\qquad
        3\left\lceil\frac{N+1}{2}\right\rceil(5M)^N,
    \]
    respectively, such that
    \begin{equation}
        \|\widehat{G}-G\|_{L^\infty([-1,1]^N)}
        \leq
        \frac{(14N^2+1)\sqrt{N}\,C_G}{M^s}.
    \end{equation}

    \item[\textnormal{(ii)}]
    \textbf{ReLU networks.}
    For every $M,L\in\mathbb{N}$, there exists a ReLU network
    $\widehat{G}:[-1,1]^N\to\mathbb{R}$ with depth
    \[
        11L+2N+18
    \]
    and width
    \[
        3^{N+3}
        \max\left\{
            N\left\lfloor M^{1/N}\right\rfloor,
            M+2
        \right\}
    \]
    such that
    \begin{equation}
        \|\widehat{G}-G\|_{L^\infty([-1,1]^N)}
        \leq
        131\cdot 2^s\sqrt{N}\,C_G
        \left(
            M^2L^2\log_3(M+2)
        \right)^{-s/N}.
    \end{equation}
\end{enumerate}
\end{theorem}

Combining Theorem~\ref{thm:extension_tim} with the kernel-specific
H\"{o}lder constants of $G$ yields explicit tanh and ReLU approximation
bounds for each kernel class considered in this paper. These results are
stated in Corollaries~\ref{G-Ghat_Sob}--\ref{G-Ghat_coef2} in
Appendix~\ref{sec:appendcor}.


\section{Remarks, Conclusions and Future Work}\label{sec:conclusion}
Although neural networks are well known as universal approximators of
finite-dimensional functions, their approximation and learning
capabilities on infinite-dimensional input spaces remain less
understood. In this paper, we study the approximation and learning of
nonlinear functionals $F:\mathcal K\to\mathbb R$, where $\mathcal K$ is
a subset of an RKHS.

In summary, we establish a general approximation theorem for
H\"older-continuous functionals on RKHS unit balls induced by Mercer
kernels. We also derive explicit error bounds for Sobolev, inverse multiquadric, and
Gaussian RKHSs, as well as RKHSs associated with kernels on the $p$-torus. 
Both tanh and ReLU networks are shown to achieve arbitrary approximation accuracy with sufficiently many parameters. In particular, the infinitely smooth inverse multiquadric and Gaussian kernels lead to substantially more favorable approximation guarantees than those obtained for finitely smooth Sobolev kernels. 
\second{Beyond approximation, we study empirical risk minimization over the proposed neural-network classes, establish an oracle inequality showing how the generalization error is controlled jointly by the approximation and estimation errors, and derive explicit finite-sample learning rates for Gaussian RKHSs.}

A key feature of our approach is the use of interpolating orthogonal projections in RKHSs. Unlike functional neural-network constructions based on basis expansions or kernel integral transformations \citep{rossi2005functional,song2023approximationarxiv}, our networks use only finitely many point evaluations of the input function. The sampling locations need not be uniformly spaced, and neither a prescribed basis nor an explicit kernel integral representation is required in the network architecture. This point-evaluation design provides a simple and flexible finite-dimensional representation of functional inputs, while preserving approximation guarantees and accommodating both Euclidean and non-Euclidean domains.

As we mentioned, this paper gives a general error bound for approximating functionals on the RKHS induced by a general Mercer kernel. An important consideration in applying this general result is the choice of $N$, the number of sampling points used to evaluate the input function. \second{The error bound reveals a trade-off associated with $N$: When $N$ is small, the discretization error is larger, as quantified by the power function measuring the discrepancy between a function and its interpolant. Increasing $N$ improves the accuracy of this finite-dimensional representation, but also increases the dimension of the neural network input and hence the network complexity required by our construction. The choices of $N$ in our kernel-specific results are obtained by balancing these two effects.} The specific choices of $N$ used in our main theorems are derived in the corresponding proofs.

Several directions remain open. First, our analysis uses fully connected tanh and ReLU networks and does not exploit additional architectural structure. Extending the framework to networks with local connectivity or parameter sharing, such as convolutional neural networks (CNN) \citep{zhou2024learning}, may yield substantially sparser representations for functional inputs exhibiting local or translation-invariant structure. Developing approximation and learning guarantees for RKHS functionals under such structured architectures would therefore be of interest.

\second{Second, our statistical analysis assumes access to an empirical risk
minimizer and does not address the optimization problem arising in practical training.
Understanding whether practical training algorithms can efficiently
attain the guarantees established here is essential for implementing
the proposed framework and remains an important direction for future
work.}

\newpage
\appendix
 \section*{Appendix}

\section{Additional Remarks on Reproducing Kernel on the $p$-torus}\label{sec:append_note}

In this section, we provide supplementary proofs related to the reproducing kernels on the 
$p$-torus, specifically concerning the positive semidefiniteness and regularity of these kernels. While these results are used in the main text, we include the proofs here for the sake of conciseness.

\begin{remark}[Positive semidefiniteness of kernel on $\mathbb{S}^1$]\label{remark:PSD}
Recall the kernel $K$ on the unit circle $\mathbb{S}^1$, defined in \eqref{kernel_S1} by 
\begin{equation*}
 K(e^{i\xi},e^{i\eta}) =  \sum_{j=-\infty}^\infty \theta_j \left(e^{i(\xi-\eta)}\right)^j, \qquad \xi, \eta \in \RR,
\end{equation*}
where the coefficients $\theta_j \in \RR_+$ satisfy $\theta_j = \theta_{-j}$ and $\theta_j \in\RR_+$ for all $j\in \ZZ$.
To verify the positive semidefiniteness of $K$, let $c\in \RR^N, \xi_\ell\in \RR$, and consider $$\sum_{\ell,k=1}^{N} c_\ell  K(e^{i\xi_\ell},e^{i\xi_k}) c_k.$$ 
We have 
\begin{eqnarray*}
    \sum_{\ell,k=1}^{N} c_\ell  K(e^{i\xi_\ell},e^{i\xi_k}) c_k &=&  \sum_{\ell,k=1}^{N} c_\ell \sum_{j=-\infty}^\infty \theta_j  e^{i(\xi_\ell-\xi_k)j}  c_k \\
    &=& \sum_{j=-\infty}^\infty \theta_j \left(\sum_{\ell=1}^{N}  c_\ell e^{i\xi_\ell j}\right) \left(\sum_{k=1}^{N} c_k e^{-i\xi_k j }\right)\\
     &=& \sum_{j=-\infty}^\infty \theta_j  \left|\sum_{\ell=1}^{N} c_\ell e^{i\xi_\ell j }\right|^2\\
     &\geq& 0, 
\end{eqnarray*}
which confirms the positive semidefiniteness of $K$.
\end{remark}

\begin{remark}[Positive semidefiniteness of kernel on $\mathbb{T}^p$]\label{remark:PSD_torus}
Recall the kernel $K$ on $\mathbb{T}^p$, defined in \eqref{kernel_torus} by  
\begin{equation*}
    K(e^{i\xi},e^{i\eta})  = \sum_{j\in\ZZ^p} \theta_j e^{ij\cdot (\xi-\eta)}, \qquad \xi, \eta \in \RR^p,
\end{equation*}
where $\theta_j \in\RR_+$ for all $j\in\ZZ_+^p$, $\theta_{j} = \theta_{\varepsilon \circ j}$ with $\varepsilon \circ j := (\varepsilon_1 j_1, \ldots, \varepsilon_p j_p)$ for all $\varepsilon \in \{-1,1\}^p$, and $j\cdot x =  j_1x_1 + \cdots + j_px_p$.
 
 For any $c\in \RR^{m^p}$ and $ \xi_\ell \in \RR$ for $\ell \in \{0,\ldots, m-1\}^p$, we have 
{\allowdisplaybreaks
\begin{eqnarray*}
    \sum_{k, \ell \in \{0,\ldots, m-1\}^p} c_\ell c_k K(e^{i\xi_\ell}, e^{i\xi_k}) &=&  \sum_{k, \ell \in \{0,\ldots, m-1\}^p} c_\ell c_k\sum_{j\in\ZZ^p} \theta_j  e^{i(\xi_\ell-\xi_k)\cdot j}  \\
    &=& \sum_{j\in\ZZ^p} \theta_j \left(\sum_{\ell \in \{0,\ldots, m-1\}^p}  c_\ell e^{i\xi_\ell \cdot j}\right) \left(\sum_{k \in \{0,\ldots, m-1\}^p}c_k e^{-i\xi_k \cdot j }\right)\\
     &=& \sum_{j\in\ZZ^p} \theta_j  \left|\sum_{\ell \in \{0,\ldots, m-1\}^p} c_\ell e^{i\xi_\ell \cdot j }\right|^2\\
     &\geq& 0. 
\end{eqnarray*}}
This shows that kernel $K$ is positive semidefinite.
\end{remark}

\begin{remark}[Regularity of kernel on $\mathbb{T}^p$ with $\theta_j = a^{\|j\|}$]\label{remark:holder_item1}
     We consider $K$ defined in \eqref{kernel_torus} by  
\begin{equation*}
    K(e^{i\xi},e^{i\eta})  = \sum_{j\in\ZZ^p} \theta_j e^{ij\cdot (\xi-\eta)}, \qquad \xi, \eta \in \RR^p,
\end{equation*} with coefficients $\theta_j = a^{\|j\|}$ with some $a \in (0,1)$ (as in item \ref{coefficient1}). For $u = e^{i\xi}$, $v= e^{i\eta}$ and $\tilde{v}= e^{i\tilde{\eta}}$ with $\xi, \eta , \tilde{\eta} \in \RR^p$, we have 
{\allowdisplaybreaks
\begin{eqnarray*}
   && |K(u,v) - K(u,\tilde{v})|\\
    &=& \left|\sum_{j\in\ZZ^p} a^{\|j\|}e^{ij\cdot (\xi-\eta)} - \sum_{j\in\ZZ^p} a^{\|j\|} e^{ij\cdot (\xi-\tilde \eta)}\right| \\
    &\leq& \sum_{j\in\ZZ^p} a^{\|j\|} \left| 
e^{ij\cdot (\xi-\eta)} - e^{ij\cdot (\xi-\tilde \eta)} \right| \\
&=&  \sum_{j\in\ZZ^p} a^{\|j\|} |\cos(j\cdot (\xi-\eta)) - \cos(j\cdot (\xi-\tilde \eta)) + i (\sin(j\cdot (\xi-\eta)) - \sin(j\cdot (\xi-\tilde \eta)) )|\\
&\leq& \sum_{j\in\ZZ^p} a^{\|j\|} \left|2\sin\left(\frac{j\cdot (\tilde \eta-\eta)}{2}\right)\right|\\
&\leq& \sum_{j\in\ZZ^p} a^{\|j\|} |j| |\eta -\tilde \eta |.
\end{eqnarray*}}
We see that this kernel is $\alpha$-H\"{o}lder continuous for $\alpha = 1$ with constant $C_K =  \sum_{j\in\ZZ^p} a^{\|j\|}|j|$, where  $a\in (0,1)$. 
\end{remark}

\begin{remark}[Regularity of kernel on $\mathbb{T}^p$ with $\theta_j = \left(1+\|j\|^2\right)^{-\gamma}$]\label{remark:holder_item2}
   Here, we consider the kernel $K$ defined in \eqref{kernel_torus} with coefficients $\theta_j = \left(1+\|j\|^2\right)^{-\gamma}$, for $\gamma >(p+1)/2$ (as in item \ref{coefficient2}).
   We have, for $u = e^{i\xi}$, $v= e^{i\eta}$ and $\tilde{v}= e^{i\tilde{\eta}}$ with $\xi, \eta , \tilde{\eta} \in \RR^p$,
\begin{eqnarray*}
   |K(u,v) - K(u,\tilde{v})|
    &=& \left|\sum_{j\in\ZZ^p} \left(1+\|j\|^2\right)^{-\gamma} e^{ij\cdot (\xi-\eta)} - \sum_{j\in\ZZ^p} \left(1+\|j\|^2\right)^{-\gamma} e^{ij\cdot (\xi-\tilde \eta)}\right| \\
    &\leq& \sum_{j\in\ZZ^p} \left(1+\|j\|^2\right)^{-\gamma} \left|
e^{ij\cdot (\xi-\eta)} - e^{ij\cdot (\xi-\tilde \eta)} \right| \\
&\leq& \sum_{j\in\ZZ^p}  \left(1+\|j\|^2\right)^{-\gamma} |j| |\eta -\tilde \eta |.
\end{eqnarray*}
We see that this kernel is $\alpha$-H\"{o}lder continuous for $\alpha = 1$ with constant $C_K =  \sum_{j\in\ZZ^p} \left(1+\|j\|^2\right)^{-\gamma}|j|$, where $\gamma>(p+1)/2$. 
\end{remark}

\section{Approximate
 Regression Map in Functional Regression}\label{sec: regression}
In this section, we apply our findings to approximate regression maps in generalized functional linear regression models. 
Functional data arise in many scientific applications, including air
pollution studies, fMRI analysis, and growth-curve modeling
\citep{yao2005functional,leng2006classification,chiou2012dynamical}.
Scalar-on-function regression, in which a scalar response is predicted
from a random function, is a central problem in functional data analysis
\citep{morris2015functional,wang2016functional}. The classical
functional linear model relate the response $Y\in \RR$ to a  square-integrable random function $X\in L^2([0,1])$  through
\begin{equation}\label{flm}
    Y
    =\int_0^1X(t)\beta(t)\,dt+\varepsilon
    =\langle X,\beta\rangle_{L^2([0,1])}+\varepsilon,
\end{equation}
where  the unknown functional parameter $\beta\in L^2([0,1])$ is referred to as {\it slope function}, and $\varepsilon$ is centered noise. A
generalized functional linear model replaces the linear conditional
mean with a link function $g:\RR\to\RR$
\citep{muller2005generalized}:
\begin{equation}\label{gflm}
    Y
    =g\!\left(\langle X,\beta\rangle_{L^2([0,1])}\right)
    +\varepsilon.
\end{equation}
We shall assume $g$ is Lipschitz continuous in $\RR$. 
A fundamental problem in statistics is estimating the slope function $\beta$ in the FLM based on a training sample 
consisting of $n$ independent copies of $(X, Y)$ (see, e.g., \citep{muller2005generalized, chen2011single, yuan2010reproducing}). An efficient estimator of $\beta$ is crucial
for subsequently retrieving
the regression mean 
    $\mathbb{E}[Y|X]:= g\left(\int_0^1 X(t) \beta(t) dt\right)$.
Here, we focus on approximating the regression map of generalized FLM by neural networks and derive the rate of convergence of approximation error.

For a function covariate $X\in L^2([0,1])$, let $F:L^2([0,1]) \to \RR$ be the regression map of generalized FLM given by
\begin{equation}\label{func_gflm}
    F: X \mapsto g\left(\int_0^1 X(t) \beta(t) dt\right), \qquad X\in L^2([0,1]).
\end{equation}   
We shall restrict the domain of $F$ onto a subset of  an RKHS $\mathcal{H}_{K_0}$ induced by a Mercer kernel $K_0$ on $[0,1] \times [0,1]$ \citep{yuan2010reproducing}.
Consider the unit ball $\mathcal{K}_0 := \{f\in \mathcal{H}_{K_0} \subset L^2([0,1]): \|f\|_{\mathcal{H}_{K_0}} \leq 1\}$. 
We assume $\mathcal{H}_{K_0}$ is dense in $L^2([0,1])$.
It is known that an RKHS induced by an inverse multiquadric kernel or a Gaussian kernel is dense in $L^2$ space, as well as the Sobolev space in $\RR$ with order $r>1/2$. 
We define an RKHS functional 
$F|_{\mathcal{H}_{K_0}}:\mathcal{K}_0 \to \RR$ by 
\begin{equation}\label{regmap}
    F|_{\mathcal{H}_{K_0}}: X \mapsto g\left(\int_0^1 X(t) \beta(t) dt\right), \qquad X\in \mathcal{K}_0 \subset L^2([0,1]).
\end{equation}

Applying our previous theoretical findings, we prove that two-layer tanh neural networks can approximate the generalized functional regression map well. Concretely, for any function covariate $X\in \mathcal{K}_0$ and some discrete points  $\bar{t} = \{t_1, \cdots, t_N\}\subset [0,1]$ (even spacing is not necessary), there exists a network $\widehat G:\RR^N \to \RR$ that takes $X(\bar{t})=(X(t_1), \cdots, X(t_N))\in\RR^N$ as input such that 
$\sup_{X\in \mathcal{K}_0}\left|F|_{\mathcal{H}_{K_0}}(X) - \widehat{G}(X(\Bar{t}))\right| \to 0$  as network widths increases.

For a function $f:[-T, T] \to \RR$, we define its Lipschitz seminorm as $$\|f\|_{Lip[-T,T]}:= \sup_{x,y\in [-T, T], x \neq y}\frac{|f(x) - f(y)|}{|x-y|}.$$
 With a slight abuse of notation, we use $\|\cdot\|_{L^2}:= \|\cdot\|_{L^2([0,1])}$ denote the $L^2$ norm over $[0,1]$.
The following result follows from Theorem \ref{thm:general} and Theorem \ref{thm:Gaussian}.
Its proof is relegated to Appendix \ref{sec:appendB}. 
\begin{corollary}[Approximation of Regression Map in Generalized FLM] \label{corollary:FLM}
Let $d=1$ and $\mathcal{X} = [0,1]$.
Assume $\mathcal{H}_{K_0} $ is dense in $L^2([0,1])$, where $\mathcal{H}_{K_0}$ is the RKHS induced by some Mercer kernel $K_0$ which is $\alpha$-H\"{o}lder continuous for some $\alpha \in (0, 1]$ with constant $C_{K_0} \geq 0$.  
Let $\kappa=\sup_{t\in[0,1]}\sqrt{K_0(t,t)}$. 
Consider the regression map $F|_{\mathcal{H}_{K_0}}$ defined in \eqref{regmap}.
 There exists some $M_0\in \NN$ such that for every $M\in \NN$ with $M>M_0$, by taking some
    $\Bar{t}=\{t_i\}_{i=1}^N \subset \mathcal{X}$ with $N\in \NN$, we have a tanh neural network $\widehat{G}: \RR^N \to \RR$ with two hidden layers of widths at most $N(M-1)$ and $3\left\lceil\frac{N+1}{2}\right\rceil (5M)^N$ satisfying 
\begin{equation}
    \sup_{X\in \mathcal{K}_0}\left|F|_{\mathcal{H}_{K_0}}(X) - \widehat{G}(X(\Bar{t}))\right|\leq  C_F (\epsilon_{K_0}(\Bar{t})) + \highlight{\frac{(14 N^2 +1) \sqrt{N}C_G}{M^s}},
\end{equation}
with $$C_F= \|g\|_{Lip([-\kappa \|\beta\|_{L^2}, \kappa \|\beta\|_{L^2}])} \|\beta\|_{L^2}, C_G = C_F\left(1+ \|(K_0[\Bar{t}])^{-1 }\|_{op}\sqrt{N} C_{K_0} \left(h_{\Bar{t}}\right)^\alpha\right),$$ 
where the Gram matrix $K_0[\Bar{t}]$ is defined in (\ref{gram}), the fill distance $h_{\Bar{t}}$ is given in (\ref{fill}), and the power function $\epsilon_{K_0}(\Bar{t})$ is defined in (\ref{power}).

In particular, if we take $K_0$ to be a Gaussian kernel with $\sigma >0$,  \\
        $m = \left\lfloor \highlight{\frac{1}{\sigma \pi}\left(\sqrt{\log M} - c (\log M)^{\frac{1}{4}}\right) }\right\rfloor$,
$\Bar{t}=\{0, \frac{1}{m},\ldots, \frac{m}{m}\}$, and $N= m+1$, we have 
\begin{equation}
    \sup_{X\in \mathcal{K}_0}\left|F|_{\mathcal{H}_{K_0}}(X) - \widehat{G}(X(\Bar{t}))\right|\leq C_{\sigma, c, s, F} \ \highlight{\exp\left(-\frac{cs}{8\sigma \pi} \sqrt{\log M}  \log (\log (M))\right)},
\end{equation}
where $C_{\sigma, c, s, F} $ is a constant depending on $\sigma, c,s$ and $C_F$, $c$ is a positive constant.
\end{corollary}

Note that we can also take $K_0$ to be an inverse multiquadric kernel or a Sobolev kernel in $\RR$. Then, by applying Theorem \ref{thm:multi} or Theorem \ref{thm:Sobolev}, we can obtain the corresponding approximation error bound. 

By the assumption that $\mathcal{H}_{K_0}$ is a dense subset, for any $X\in cl(\mathcal{K}_0)$ (i.e., the closure of $\mathcal{K}_0$), there exists a sequence of functions $\{X_n\in \mathcal{K}_0, n\in \NN\}$ uniformly convergent to $X$ (i.e., $\sup_{t\in [0,1]}|X_n(t) - X(t)| \to 0$ as $n\to \infty$). 
We can see that the functional $F|_{\mathcal{H}_{K_0}}$ can be extended to $F$ on $cl(\mathcal{K}_0)$ as
     \begin{equation*}
         F(X) = \lim_{n\to \infty} F|_{\mathcal{H}_{K_0}} (X_n) = \lim_{n\to \infty} g\left(\int_0^1 X_n(t) \beta(t) dt\right),\qquad X\in cl(\mathcal{K}_0)\subset L^2([0,1]).
     \end{equation*}

   In this paper, we only consider approximating functional regression maps using neural networks. Based on our results, we find that the choice of $\Bar{t}$ and the smoothness of the kernel that induces an RKHS play important roles in improving the approximation abilities. 
   To prove that neural networks can learn functional regression maps well, we shall extend our work to a generalization analysis in the future.
   Mathematically, the generalization error consists of the approximation error and the estimation error (based on training samples of independent copies of $(X,Y)$). 
   We may assume that the slope function $\beta$ resides in RKHS to determine the estimation error. 
   This is, in fact, a common assumption imposed in the statistics literature on the estimation and inference of FLM (see, \citep{yuan2010reproducing, shang2015nonparametric,balasubramanian2025functional}).  

\section{Proof of Supporting Lemmas}\label{sec:appendA}
Here, we present the proofs of supporting lemmas. 

\subsection{Projection and Power-Function Estimates}\label{subsec:project}

\noindent
{\bf Proof of Lemma \ref{lemma:ortho}}.
   The nodal functions  $\{\psi_i\}_{i=1}^N$ is given by
\begin{equation}
       \begin{bmatrix}
           \psi_1\\
           \vdots\\
           \psi_N
       \end{bmatrix}= ( K[\Bar{t} ])^{-1} \begin{bmatrix}
           K_{t_1}\\
           \vdots\\
           K_{t_N}
       \end{bmatrix}. 
    \end{equation}
    In this way, we have
    \begin{eqnarray*}
       \begin{bmatrix}
           \psi_1\\
           \vdots\\
           \psi_N
       \end{bmatrix}(t_\ell) 
       &=& ( K[\Bar{t} ])^{-1} \begin{bmatrix}
           K(t_1,t_\ell)\\
           \vdots\\
            K(t_N,t_\ell)
       \end{bmatrix}\\
       &=&\left((K[\Bar{t} ])^{-1} K[\Bar{t} ]\right)_{\cdot , \ell}\qquad (\text{the $\ell$-th column of }(K[\Bar{t} ])^{-1} K[\Bar{t} ])\\
       &=&I_{\cdot , \ell}. \qquad (\text{the $\ell$-th column of identity matrix }I)
     \end{eqnarray*}
    We can see that 
    \begin{equation*}
    \psi_i(t_\ell) = \begin{cases}
      1, & \text{if } i=\ell,\\
      0, & \text{if } i \neq \ell.
    \end{cases}
\end{equation*}
Then we have
\begin{eqnarray*}
    Pf(t_\ell) = \sum_{i=1}^N f(t_i)\psi_i(t_\ell) = f(t_\ell)\psi_\ell(t_\ell) = f(t_\ell).
\end{eqnarray*}
The proof of statement \ref{item1} is complete.

Recall the reproducing property of RKHS:  $f(x) = \langle f, K_x \rangle_{\mathcal{H}_K}$. 
Since $f(t_\ell)- Pf(t_\ell) = 0$, we know 
\begin{equation*}
    \langle f,K_{t_\ell} \rangle_{\mathcal{H}_K} - \langle Pf,K_{t_\ell} \rangle_{\mathcal{H}_K}= f(t_\ell) - Pf(t_\ell) =0 \iff  \langle f-Pf,K_{t_\ell} \rangle_{\mathcal{H}_K}=0.
\end{equation*}
The proof of statement \ref{item2} is complete. 
\hfill\BlackBox

Next, we present the proof of Lemma \ref{f-Pf}.

\noindent
{\bf Proof of Lemma \ref{f-Pf}}.
    With $\{\psi_i\}_{i=1}^N$ defined in (\ref{psi}), for  $x\in \mathcal{X}$, by the reproducing property and the Cauchy–Schwarz inequality, we have 
    \begin{eqnarray*}
        |f(x) - Pf(x)| = \left|f(x)-\sum_{i=1}^N f(t_i)\psi_i(x)\right| &=& \left|\langle f, K_x \rangle_{\mathcal{H}_K} -\sum_{i=1}^N \langle f, K_{t_i}\rangle_{\mathcal{H}_K} \psi_i(x) \right|\\
        &=& \left|\langle f, K_x -  \sum_{i=1}^N \psi_i(x)  K_{t_i} \rangle_{\mathcal{H}_K}\right|\\
        &\leq&  \|f\|_{\mathcal{H}_K} \left\|K_x -  \sum_{i=1}^N \psi_i(x)  K_{t_i}\right\|_{\mathcal{H}_K}.
    \end{eqnarray*}
    Let the function $H = H_x$ on $\RR^N$ be $$H(c) : = \left\|K_x -  \sum_{i=1}^N c_i  K_{t_i}\right\|^2_{\mathcal{H}_K} = K(x,x) -2  \sum_{i=1}^N c_i K(t_i,x)+ \sum_{i,j=1}^N c_i c_j K(t_i, t_j).$$ 
    Then \begin{eqnarray*}
        \frac{\partial H}{\partial c} = 0 \iff 2 K[\Bar{t}] c = 2 \begin{bmatrix}
            K(x, t_1)\\
            \vdots\\
            K(x,t_N)
        \end{bmatrix} \iff c = (K[\Bar{t}])^{-1 } \begin{bmatrix}
            K(x, t_1)\\
            \vdots\\
            K(x,t_N)
        \end{bmatrix} = \begin{bmatrix}
           \psi_1(x)\\
           \vdots\\
           \psi_N(x)
       \end{bmatrix}.
    \end{eqnarray*}
    Thus, $\|K_x -  \sum_{i=1}^N \psi_i(x)  K_{t_i}\|_{\mathcal{H}_K} = \min_{c\in \RR^N} \sqrt{H(c)} \leq \max_{x\in \mathcal{X}} \min_{c\in\RR^N}  \sqrt{H(c)} =  \epsilon_K(\Bar{t})$. This is valid for every $x \in \mathcal{X}$. The proof is complete.
    \hfill\BlackBox

Next, we give the proof of Lemma \ref{lemma:power}.

\noindent
{\bf Proof of Lemma \ref{lemma:power}}.
For each $x\in\mathcal{X}$, let $t_{j_x}\in\Bar{t}$ be a closest
sampling point to $x$. Recall the definition of the power function: $$\epsilon_K(\Bar{t})= \max_{x\in \mathcal{X}} \min_{c\in\RR^N} \left\|K_x -  \sum_{i=1}^N c_i  K_{t_i}\right\|_{\mathcal{H}_K}.$$ 
We choose
$c$ such that 
\begin{equation}
    c_i =  \begin{cases}
      1, & \text{if $i=j_x$ },\\
      0, & \text{otherwise}.
    \end{cases}  
\end{equation}
 By the symmetry and $\alpha$-H\"{o}lder continuity of $K$,
\begin{align*}
    \min_{c\in\mathbb{R}^N}
    \left\|
        K_x-\sum_{i=1}^N c_iK_{t_i}
    \right\|_{\mathcal{H}_K}^2
    &\leq
    \|K_x-K_{t_{j_x}}\|_{\mathcal{H}_K}^2 \\
    &=
    K(x,x)-2K(x,t_{j_x})
    +K(t_{j_x},t_{j_x})\\
    &\leq
    |K(x,x)-K(x,t_{j_x})| +
    |K(t_{j_x},t_{j_x})-K(t_{j_x},x)| \\
    &\leq
    2C_K\,d_{\mathcal{X}}(x,t_{j_x})^\alpha \\
    &\leq
    2C_K\,h_{\Bar{t}}^\alpha.
\end{align*}
Taking square roots and then the supremum over
$x\in\mathcal{X}$ gives
\[
    \epsilon_K(\Bar{t})
    \leq
    \sqrt{2C_K}\,h_{\Bar{t}}^{\alpha/2}.
\]
Finally, Lemma~\ref{f-Pf} implies
\[
    \|f-Pf\|_\infty
    \leq
    \|f\|_{\mathcal{H}_K}\epsilon_K(\Bar{t})
    \leq
    \sqrt{2C_K}\,h_{\Bar{t}}^{\alpha/2}
    \|f\|_{\mathcal{H}_K},
\]
which completes the proof.
 \hfill\BlackBox

\subsection{Regularity of the Finite-Dimensional Function $G$}\label{subsec:regularity}

For completeness, we restatement of Lemma \ref{lemma:reg_G}. 
\begin{lemma*}[Restate Lemma \ref{lemma:reg_G}]
Suppose that Assumption~\ref{assump:holder-F} holds with exponent
$s\in(0,1]$ and constant $C_F$. Then $G$  defined in \eqref{G} is $s$-H\"{o}lder
continuous on $[-1,1]^N$ with respect to the Euclidean norm, with
H\"{o}lder constant
\begin{equation}
    C_G
    =C_F\left(
        1+
        \bigl\|(K[\Bar t])^{-1}\bigr\|_{op}
        \sqrt N\,C_K h_{\Bar t}^{\alpha}
    \right)^s.
\end{equation}
\end{lemma*} 

\noindent
{\bf Proof  of Lemma \ref{lemma:reg_G}}.
For any $c, \Tilde{c} \in [-1,1]^N$, we have 
\begin{equation}
   | G(c) - G(\Tilde{c})| = \left|F\left(\sum_{i=1}^N c_i \psi_i\right) - F\left(\sum_{i=1}^N \Tilde{c_i} \psi_i\right)\right| \leq C_F \left\|\sum_{i=1}^N (c_i - \Tilde{c_i}) \psi_i\right\|^s_\infty.
\end{equation}
We will proceed to bound $\|\sum_{i=1}^N a_i\psi_i\|_\infty$ for any $a\in \RR^N$. 
For $x\in \mathcal{X}$, let $t_{j_x}$ be the closest point in $\bar{t}$ to $x$. Observe that $\left(\sum_{i=1}^N a_i\psi_i\right)(t_{j_x}) = a_{j_x}$ and thus 
\begin{equation*}
    \left|\sum_{i=1}^N a_i\psi_i(x)\right| = \left|\sum_{i=1}^N a_i\psi_i(t_{j_x}) + \sum_{i=1}^N a_i (\psi_i(x) - \psi_i(t_{j_x}))\right| = \left|a_{j_x} + \sum_{i=1}^N a_i (\psi_i(x) - \psi_i(t_{j_x}))\right|.
\end{equation*}
Next, we have \begin{eqnarray*}
    \left|\sum_{i=1}^N a_i (\psi_i(x) - \psi_i(t_{j_x}))\right| &=& \left|a^T (K[\Bar{t}])^{-1 } \begin{bmatrix}
        K(t_1, x) - K(t_1, t_{j_x})\\
        \vdots\\
        K(t_N, x) - K(t_N, t_{j_x})
    \end{bmatrix}\right|\\
    &=& \left|((K[\Bar{t}])^{-1 }a)^T \begin{bmatrix}
        K(t_1, x) - K(t_1, t_{j_x})\\
        \vdots\\
        K(t_N, x) - K(t_N, t_{j_x})
    \end{bmatrix}\right|\\
    &\leq& \|(K[\Bar{t}])^{-1 }\|_{op} \|a\| \sqrt{\sum_{i=1}^N | K(t_i, x) - K(t_i, t_{j_x})|^2}.
    \end{eqnarray*}  
    Since $K$ is $\alpha$-H\"older continuous, we know that  
      $| K(t_i, x) - K(t_i, t_{j_x})| \leq  C_K\left(d_\mathcal{X}(x,t_{j_x})\right)^\alpha. $ Hence, we know 
 \begin{eqnarray*}
    \left|\sum_{i=1}^N a_i (\psi_i(x) - \psi_i(t_{j_x}))\right| 
     &\leq& \|(K[\Bar{t}])^{-1 }\|_{op} \|a\| \sqrt{N} C_K\left(\sup_{x\in \mathcal{X}}d_\mathcal{X}(x,t_{j_x})\right)^\alpha. 
    \end{eqnarray*}     
         Recall $h_{\Bar{t}} = \sup_{x\in \mathcal{X}} d_\mathcal{X}(x,t_{j_x})$. Thus, we have
\begin{eqnarray*}
    | G(c) - G(\Tilde{c})| 
    &\leq& C_F \left\{\left|c_{j_x} - \Tilde{c}_{j_x}\right| + \|(K[\Bar{t}])^{-1 }\|_{op} \|c - \Tilde{c}\|\sqrt{N} C_K\left(h_{\Bar{t}}\right)^\alpha\right\}^s\\
    &\leq& C_F \|c - \Tilde{c}\|^s \left\{1 + \|(K[\Bar{t}])^{-1 }\|_{op} \sqrt{N} C_K\left(h_{\Bar{t}}\right)^\alpha\right\}^s.
\end{eqnarray*}
The proof is complete. 
\hfill\BlackBox

\vspace{20pt}

Since $K[\Bar{t}]$ is a real, symmetric matrix, it is diagonalizable, and so is $(K[\Bar{t}])^{-1 }$. Let $\{\lambda_1, \ldots, \lambda_N\}$ denote the non-increasing sequence of eigenvalues of $K[\Bar{t}]$. We  have  \begin{equation}\label{operatornorm}
    \|(K[\Bar{t}])^{-1}\|_{op} = \frac{1}{\min_{1 \leq j \leq N} \lambda_j} =: \frac{1}{\lambda_{N}},
\end{equation}
where $\lambda_{N}$ is the smallest eigenvalue of $K[\Bar{t}]$.

Since $K[\Bar{t}] v = \lambda_{N} v$ for some eignevector $v=(v_1, \cdots, v_N)\in \RR^N$ of norm 1 corresponding to the eigenvalue $\lambda_N$, we have
\begin{eqnarray} \label{lambda_N}
    \lambda_{N} = \lambda_{N} v^T v = v^T K[\Bar{t}] v 
    = \sum_{i=1}^N  \sum_{\ell=1}^N v_i v_\ell K(t_i, t_\ell). 
\end{eqnarray}
In other words, to compute $\|(K[\Bar{t}])^{-1}\|_{op}$, it suffices to compute  $\lambda_{N} = \sum_{i=1}^N  \sum_{\ell=1}^N v_i v_\ell K(t_i, t_\ell)$.

For the Euclidean and toroidal kernels considered in this paper, the
smallest eigenvalue of the Gram matrix can be controlled through the
spectral representation of the kernel. We record the two estimates
used throughout the proofs.

\begin{lemma}\label{lemma:norminverse}
Let $\mathcal{X}=[0,1]^d$ and
    $\Bar{t}=\left\{0,\frac{1}{m},\ldots,\frac{m}{m}\right\}^d$.
Suppose that $K(x,y)=\phi(x-y)$ is a translation-invariant reproducing
kernel and define
\begin{equation*}
    \Gamma_m
    :=\min_{\xi\in[-m/2,m/2]^d}\widehat{\phi}(\xi)>0,
\end{equation*}
where $\widehat{\phi}$ is the Fourier transform of $\phi$.
Then
\begin{equation}
    \bigl\|(K[\Bar{t}])^{-1}\bigr\|_{op} = \frac{1}{\lambda_N}
    \leq\frac{1}{m^d\Gamma_m},
\end{equation}
 where $\lambda_{N}$ is the smallest eigenvalue of $K[\Bar{t}]$.
\end{lemma}

\noindent
{\bf Proof of Lemma \ref{lemma:norminverse}}.
From Lemma \ref{lemma:trans}, we know that the Fourier transform of a translation-invariant and reproducing kernel $K$ is real-valued and non-negative. 
We also know that, from (\ref{operatornorm}) and (\ref{lambda_N}), to estimate  $\|(K[\Bar{t}])^{-1}\|_{op}$ for a translation-invariant $K$ over $\RR^d$, it suffices to estimate the smallest eigenvalue of $K[\Bar{t}]$ given by $$\lambda_{N}
= \sum_{k=1}^N \sum_{\ell=1}^N v_k v_\ell K(t_k, t_\ell),$$ where $v\in \RR^N$ is an eigenvector of $K[\Bar{t}]$ of norm $1$. 
    
By the inverse Fourier transform, we have 
\begin{eqnarray*}
    \lambda_{N}
    = \sum_{k=1}^N \sum_{\ell=1}^N v_k v_\ell K(t_k, t_\ell) &=&  \int_{\RR^d} \widehat{\phi}(\xi)\left|\sum_{k=1}^N v_k e^{2\pi i\xi \cdot t_k}\right|^2 d\xi
   \  \geq \ \int_{[-\frac{m}{2},\frac{m}{2}]^d} \widehat{\phi}(\xi)\left|\sum_{k=1}^N v_k e^{2\pi i\xi \cdot t_k}\right|^2 d\xi.
\end{eqnarray*}
But $\Gamma_m = \min_{\xi \in [-\frac{m}{2},\frac{m}{2}]^d} \widehat{\phi}(\xi)$. 
Then, we have 
\begin{eqnarray}\label{Gamma_m}
     \lambda_{N}  \geq \Gamma_m\int_{[-\frac{m}{2},\frac{m}{2}]^d} \left|\sum_{k=1}^N v_k e^{2\pi i\xi \cdot t_k}\right|^2 d\xi \nonumber 
     &=& \Gamma_m\int_{[-\frac{1}{2},\frac{1}{2}]^d} \left|\sum_{k\in \{0,1,\ldots,m\}^d} v_k e^{2\pi i\xi \cdot k}\right|^2 m^d d\xi \nonumber\\
      &=& m^d \Gamma_m \|v\|^2 \nonumber\\ 
      &=& m^d \Gamma_m.
\end{eqnarray}
Here, we have used the fact that $\{e^{2\pi i\xi \cdot k}\}_{k\in \ZZ^d}$ is an orthonormal basis of $L_2 [(-1/2, 1/2)^d]$.

\hfill\BlackBox

\vspace{10pt}
For kernels on $p$-torus, we can estimate   $\|(K[\Bar{t}])^{-1}\|_{op} $ in terms of the decay of Fourier coefficients. 
\begin{lemma}\label{lemma:norminverse_torus}
Let $\mathcal{X}=\mathbb{T}^p$ and let $\Bar{t}$ be the uniform grid
defined in \eqref{torus_points}, with $N=m^p$. For a kernel of the
form \eqref{kernel_torus},
\begin{equation}
    \bigl\|(K[\Bar{t}])^{-1}\bigr\|_{op}
    \leq\frac{1}{N\Theta_N},
    \qquad
    \Theta_N
    :=\min_{j\in \left[-\frac{m}{2}, \frac{m}{2}\right]^p}\theta_j.
\end{equation}
\end{lemma}

\noindent
{\bf Proof of Lemma \ref{lemma:norminverse_torus}}.
With the chosen $\bar{t}=\left(e^{i\frac{2\pi}{m}k}\right)_{k\in \{0,1,\ldots, m-1\}^p}$, the Gram matrix is given by 
\begin{eqnarray*}
    K[\Bar{t}] &:=& (K(t_k, t_\ell))_{k,\ell \in \{0,1,\ldots, m-1\}^p}\\ 
    &=& \left(\sum_{j\in \ZZ^p}\theta_j e^{i \frac{2\pi}{m}(k-\ell)\cdot j} \right)_{k,\ell \in \{0,1,\ldots, m-1\}^p}\in \RR^{m^p \times m^p}.
\end{eqnarray*}
Note that $N= m^p$ indicates the number of points in the set $\Bar{t}$. From \eqref{operatornorm}, 
we know \begin{equation*}
    \|(K[\Bar{t}])^{-1}\|_{op} = \frac{1}{\min_{1 \leq j \leq N} \lambda_j} =: \frac{1}{\lambda_{N}},
\end{equation*}
where $\lambda_{N}$ is the smallest eigenvalue of $K[\Bar{t}]$.
Thus, to estimate $\|(K[\Bar{t}])^{-1}\|_{op}$, it suffices to estimate the smallest eigenvalue of $K[\Bar{t}]$ given by $$\lambda_{N}
= \sum_{k, \ell \in \{0,\ldots, m-1\}^p}v_k  v_\ell K(t_k, t_\ell),$$ where $v\in \RR^{N=m^p}$ is an eigenvector of $K[\Bar{t}]$ of norm 1 corresponding to $\lambda_N$. 

Let $\Theta_{N}:= \min_{j\in \left[-\frac{m}{2}, \frac{m}{2}\right]^p} \theta_j$.
We have
{\allowdisplaybreaks
\begin{eqnarray*}
    \lambda_{N}
&=&\sum_{k, \ell \in \{0,\ldots, m-1\}^p}v_k v_\ell K(t_k, t_\ell)\\
&=& \sum_{k, \ell \in \{0,\ldots, m-1\}^p}  v_k  v_\ell\sum_{j\in \ZZ^p} \theta_j e^{i \frac{2\pi}{m}(k-\ell)\cdot j} \\
&=&  \sum_{j\in \ZZ^p}\theta_j  \left|\sum_{k \in \{0,\ldots, m-1\}^p} v_k e^{i \frac{2\pi }{m} k \cdot j}\right|^2
\\
&\geq& \sum_{j\in \left[-\frac{m}{2}, \frac{m}{2}\right)^p} \theta_j \left|\sum_{k \in \{0,\ldots, m-1\}^p}  v_k e^{i \frac{2\pi }{m} k \cdot j}\right|^2 \\
&\geq& \Theta_{N}  \sum_{j\in \left[-\frac{m}{2}, \frac{m}{2}\right)^p}  \left|\sum_{k \in \{0,\ldots, m-1\}^p}  v_k e^{i \frac{2\pi }{m} k \cdot j}\right|^2\\
&=& \Theta_{N}  \sum_{j\in \left[-\frac{m}{2}, \frac{m}{2}\right)^p}   \sum_{k,\ell \in \{0,\ldots, m-1\}^p}  v_k  v_\ell e^{i\frac{2\pi}{m} (k-\ell) \cdot j}\\
&=& \Theta_{N} \sum_{k,\ell \in \{0,\ldots, m-1\}^p} v_k  v_\ell \left(\sum_{j\in \left[-\frac{m}{2}, \frac{m}{2}\right)^p}   e^{i \frac{2\pi}{m} (k-\ell)\cdot j}\right).
\end{eqnarray*}}
If $k=\ell$, we have 
    $$\sum_{j\in \left[-\frac{m}{2}, \frac{m}{2}\right)^p} e^{i\frac{2\pi}{m} (k-\ell)\cdot j} = \sum_{j\in \left[-\frac{m}{2}, \frac{m}{2}\right)^p} e^0 = N=m^p. $$
If $k\neq \ell$,
 we have 
\begin{eqnarray*}
    \sum_{j\in \left[-\frac{m}{2}, \frac{m}{2}\right)^p} e^{i\frac{2\pi}{N} (k-\ell)\cdot j} 
    &=& \sum_{j_1\in \left[-\frac{m}{2}, \frac{m}{2}\right)} \sum_{j_2\in \left[-\frac{m}{2}, \frac{m}{2}\right)} \cdots \sum_{j_p\in \left[-\frac{m}{2}, \frac{m}{2}\right)} e^{i\frac{2\pi}{N} ((k_1-\ell_1)j_1 + (k_2-\ell_2)j_2 + \cdots + (k_p-\ell_p)j_p)} \\
    &=& \prod_{s=1}^p \left(\sum_{j_s\in \left[-\frac{m}{2}, \frac{m}{2}\right)} e^{i\frac{2\pi}{N} ((k_s-\ell_s)j_s)}\right)\\
    &=& 0.
\end{eqnarray*}
Thus, we have
\begin{eqnarray*}
     \lambda_{N} &\geq& \Theta_{N} \sum_{k,\ell \in \{0,\ldots, m-1\}^p}  v_k  v_\ell \left(\sum_{j\in \left[-\frac{m}{2}, \frac{m}{2}\right)^p}  e^{ij \frac{2\pi}{N} (k-\ell)}\right)\\
     &\geq& \Theta_{N} \sum_{k\in \{0,\ldots, m-1\}^p} v_k^2 N \\
     &=& \|v\|_2^2   \Theta_{N} N \\
     &=& \Theta_{N} N.\end{eqnarray*}
     The proof is complete
\hfill\BlackBox

\vspace{10pt}

Combining these estimates of $\bigl\|(K[\Bar{t}])^{-1}\bigr\|_{op}$ with
Lemma~\ref{lemma:reg_G} yields the kernel-specific H\"{o}lder
constants summarized in Table~\ref{tab:regularity_G} below. 
Full statements and proofs of these kernel-specific constants are in Appendix \ref{subsec:kernel_specific_regularity}.

\subsection{Kernel-Specific Regularity Bounds for $G$}
\label{subsec:kernel_specific_regularity}

\begin{table}[h]
\centering
\caption{H\"{o}lder constants of the finite-dimensional function $G$
for different kernel classes considered in the paper.}
\label{tab:regularity_G}
\renewcommand{\arraystretch}{1.35}
\begin{tabular}{|p{0.28\textwidth}|p{0.48\textwidth}|p{0.15\textwidth}|}
\hline
Kernel class & H\"{o}lder constant $C_G$ & Supporting result \\
\hline
Sobolev kernel
& $\displaystyle
   C_F C_{r,d,s}
   m^{2rs-\frac{ds}{2}-s}d^{rs+\frac{s}{2}}$
& Lemma~\ref{lemma:C_G_Sob} \\

Inverse multiquadric kernel
& $\displaystyle
   C_F\left(
   1+\frac{4d\beta\sigma^{-2\beta-2}}
   {C_{\sigma,\beta,d}}
   (2m)^{-\beta-\frac12}e^{4\sigma M_dm}
   \right)^s$
& Lemma~\ref{lemma:C_G_multi} \\

Gaussian kernel
& $\displaystyle
   C_F C_{\sigma,d,s}
   e^{\sigma^2\pi^2dsm^2/2}$
& Lemma~\ref{lemma:C_G_Gaussian} \\

Toroidal kernel with $\theta_j=a^{\|j\|}$
& $\displaystyle
   C_F C_{a,p,s}
   a^{-m\sqrt{p}s/2}m^{-ps/2-s}$
& Lemma~\ref{lemma:C_G_coef1} \\

Toroidal kernel with
$\theta_j=(1+\|j\|^2)^{-\gamma}$
& $\displaystyle
   C_F C_{\gamma,p,s}
   m^{s(2\gamma-\frac{p}{2}-1)}$
& Lemma~\ref{lemma:C_G_coef2} \\
\hline
\end{tabular}
\end{table}

\begin{lemma}[Sobolev Space $W_2^r(\mathcal{X})$]\label{lemma:C_G_Sob}
Let $d,m\in\NN$, $\mathcal{X}=[0,1]^d$, and
$r-d/2>1$. Consider the uniform grid
    $\Bar{t}
    =\left\{0,\frac{1}{m},\ldots,\frac{m}{m}\right\}^d,
    N=(m+1)^d$,
and let $\mathcal{H}_K=W_2^r(\mathcal{X})$. Then the corresponding
function $G:[-1,1]^N\to\RR$ is $s$-H\"{o}lder continuous, and one may
take
\begin{equation}
    C_G
    =C_F C_{r,d,s}
    m^{2rs-\frac{ds}{2}-s}d^{rs+\frac{s}{2}},
\end{equation}
where $C_{r,d,s}>0$ depends only on $r$, $d$, and $s$.
\end{lemma}

\noindent
{\bf Proof of Lemma \ref{lemma:C_G_Sob}}.
The Sobolev space $W_2^r(\mathcal{X})$ is an RKHS induced by a
translation-invariant kernel whose Fourier transform is
\[
    \widehat{\phi}(\xi)=(1+\|\xi\|^2)^{-r},
    \qquad \xi\in\RR^d.
\]
From \eqref{fourier_Sobolev}, we know  $\Gamma_m =  \min_{\xi \in [-\frac{m}{2},\frac{m}{2}]^d} \widehat{\phi}(\xi) \geq \left(1+ \frac{dm^2}{4}\right)^{-r}$
for this function $\widehat \phi$.
Lemma~\ref{lemma:norminverse} therefore gives
\begin{equation}\label{eq:inverse_Sob}
    \bigl\|(K[\Bar{t}])^{-1}\bigr\|_{op}
    \leq
    \frac{1}{m^d}
    \left(1+\frac{dm^2}{4}\right)^r.
\end{equation}

By the Sobolev embedding theorem, the kernel satisfies \eqref{reg_K} $\alpha$-H\"{o}lder continuous for $0<\alpha \leq 1$ with constant $C_K \geq 0$
with exponent $\alpha=1$ and a constant $C_K = C_{r,d}$ depending only on
$r$ and $d$. Moreover,
\[
    h_{\Bar{t}}\leq\frac{\sqrt d}{m}
    \qquad\text{and}\qquad
    \sqrt N=(m+1)^{d/2}.
\]
It follows from Lemma \ref{lemma:reg_G} the associated $G$ function  $G(c) = F\left(\sum_{i=1}^N c_i \psi_i\right)$ is $s$-H\"{o}lder continuous for $s \in (0,1]$ with constant 
\begin{eqnarray*}
    & & C_F\left(1+ \|(K[\Bar{t}])^{-1 }\|_{op}\sqrt{N} C_K \left(h_{\Bar{t}}\right)^\alpha\right)^s\\
    &\leq& C_F\left(1+ \frac{1}{m^d}\left(1+ \frac{dm^2}{4}\right)^{r}(m+1)^{\frac{d}{2}}  C_{r,d} \left(\frac{\sqrt{d}}{m}\right)\right)^s\\
     &\leq&
    C_F\left(
        1+C_{r,d}
        m^{2r-\frac d2-1}d^{r+\frac12}
    \right)^s.
\end{eqnarray*}
Since $r-d/2>1$ and $m,d\geq1$, the additive constant can be
absorbed into a constant $C_{r,d,s}$ depending only on $r$, $d$, and $s$.
Therefore,
\[
    C_G
    \leq
    C_F C_{r,d,s}
    m^{2rs-\frac{ds}{2}-s}d^{rs+\frac{s}{2}},
\]
as claimed.\hfill\BlackBox

\begin{lemma}[Inverse Multiquadric Kernel]\label{lemma:C_G_multi}
Let $d,m\in\NN$, $\sigma,\beta>0$, and
$\mathcal{X}=[0,1]^d$. Consider the inverse multiquadric kernel
\[
    K(u,v)=(\sigma^2+\|u-v\|^2)^{-\beta}
\]
and the uniform grid
\[
    \Bar{t}
    =\left\{0,\frac{1}{m},\ldots,\frac{m}{m}\right\}^d,
    \qquad N=(m+1)^d.
\]
Define
\[
    M_d
    :=12\left(
        \frac{\pi\Gamma^2\bigl((d+2)/2\bigr)}{9}
    \right)
    \leq6.38d.
\]
Then the corresponding function $G:[-1,1]^N\to\RR$ is
$s$-H\"{o}lder continuous, and one may take
\begin{equation}\label{C_G_multi}
    C_G
    =C_F\left(
        1+
        \frac{4d\beta\sigma^{-2\beta-2}}
             {C_{\sigma,\beta,d}}
        (2m)^{-\beta-\frac12}e^{4\sigma M_dm}
    \right)^s,
\end{equation}
where $C_{\sigma,\beta,d}>0$ depends only on $\sigma$, $\beta$, and
$d$.
\end{lemma}

\noindent
{\bf Proof of Lemma \ref{lemma:C_G_multi}}.
First, the inverse multiquadric kernel is Lipschitz continuous on
$\mathcal{X}$. Indeed, for $u,v,\widetilde v\in\mathcal{X}$, the Mean Value Theorem gives
\begin{eqnarray*}     
\  | K(u,v) - K(u,\Tilde{v})| &= &\left|(\sigma^2 + |u-v|^2)^{-\beta}- (\sigma^2 + |u-\Tilde{v}|^2)^{-\beta}\right| \\
&\leq& \beta \sigma^{-2\beta-2} \left|\sigma^2 +|u-v|^2- \sigma^2 -|u-\Tilde{v}|^2\right|\\
&\leq& \beta   \sigma^{-2\beta-2} (|u-v|+|u-\Tilde{v}|)||u-v|-|u-\Tilde{v}||\\
&\leq&2\sqrt{d}\beta  \sigma^{-2\beta-2}|v-\Tilde{v}|.
\end{eqnarray*}
We see that this kernel is $\alpha$-H\"{o}lder continuous for $\alpha = 1$ with constant $C_K = 2\sqrt{d}\beta  \sigma^{-2\beta-2}$. 

Let
\[
    q_{\Bar{t}}
    :=\frac12\min_{u\neq v\in\Bar{t}}\|u-v\|
\]
denote the separation distance. By
\citep[Lemma~8.2]{fasshauer2005meshfree}, the smallest eigenvalue of
$K[\Bar{t}]$ satisfies
\begin{equation}\label{eq:lambda_multi}
    \lambda_{\min}(K[\Bar{t}])
    \geq
    C_{\sigma,\beta,d} \ 
    q_{\Bar{t}}^{-\beta+\frac{1-d}{2}}
    e^{-2\sigma M_d/q_{\Bar{t}}}.
\end{equation}
For the uniform grid, $q_{\Bar{t}}=1/(2m)$, and hence
\[
    \lambda_{\min}(K[\Bar{t}])
    \geq
    C_{\sigma,\beta,d}
    (2m)^{\beta+\frac{d-1}{2}}
    e^{-4\sigma M_dm}.
\]
It follows from \eqref{operatornorm} that
\begin{equation}\label{eq:inverse_multi}
    \bigl\|(K[\Bar{t}])^{-1}\bigr\|_{op}
    \leq
    \frac{1}{C_{\sigma,\beta,d}}
    (2m)^{-\beta-\frac{d-1}{2}}
    e^{4\sigma M_dm}.
\end{equation}

Since $h_{\Bar{t}}\leq\sqrt d/m$, and
$\sqrt N=(m+1)^{d/2}$ in Lemma~\ref{lemma:reg_G}, we obtain
\begin{align*}
    C_G &= C_F\left(1+ \|(K[\Bar{t}])^{-1 }\|_{op}\sqrt{N} C_K \left(h_{\Bar{t}}\right)^\alpha\right)^s\\
&\leq C_F\left(1+ \frac{1}{C_{\sigma, \beta,d}} (2m)^{-\beta -\frac{d-1}{2}}e^{4\sigma M_d m}(m+1)^{\frac{d}{2}} 2\sqrt{d}\beta  \sigma^{-2\beta-2} \frac{\sqrt{d}}{m}\right)^s\\
    &\leq
    C_F\left(
        1+
        \frac{2d\beta\sigma^{-2\beta-2}}
             {C_{\sigma,\beta,d}}
        (2m)^{-\beta-\frac{d-1}{2}}
        e^{4\sigma M_dm}
        \frac{(m+1)^{d/2}}{m}
    \right)^s \\
    &\leq
    C_F\left(
        1+
        \frac{4d\beta\sigma^{-2\beta-2}}
             {C_{\sigma,\beta,d}}
        (2m)^{-\beta-\frac12}
        e^{4\sigma M_dm}
    \right)^s,
\end{align*}
where the last inequality uses $(m+1)^{d/2}\leq(2m)^{d/2}$.
This proves the result.\hfill\BlackBox

\begin{lemma}[Gaussian Kernel]\label{lemma:C_G_Gaussian}
Let $d,m\in\NN$, $\sigma>0$, and $\mathcal{X}=[0,1]^d$.
Consider the Gaussian kernel
\[
    K(u,v)=\exp\left(-\frac{\|u-v\|^2}{2\sigma^2}\right)
\]
and the uniform grid
\[
    \Bar{t}
    =\left\{0,\frac{1}{m},\ldots,\frac{m}{m}\right\}^d,
    \qquad N=(m+1)^d.
\]
Then the corresponding function $G:[-1,1]^N\to\RR$ is
$s$-H\"{o}lder continuous, and one may take
\begin{equation}\label{C_G_Gaussian}
    C_G
    =C_FC_{\sigma,d,s}
    e^{\sigma^2\pi^2dsm^2/2},
\end{equation}
where $C_{\sigma,d,s}>0$ depends only on $\sigma$, $d$, and $s$.
\end{lemma}

\noindent
{\bf Proof of Lemma \ref{lemma:C_G_Gaussian}}.
For $u,v,\widetilde v\in\mathcal{X}$, the Mean Value Theorem gives
\begin{eqnarray*}     
\  | K(u,v) - K(u,\Tilde{v})| &=& \left|e^{-\frac{|u-v|^2}{2\sigma^2}} - e^{-\frac{|u-\Tilde{v}|^2}{2\sigma^2}}\right| \\
&\leq& 1 \cdot \left|\frac{|u-v|^2}{2\sigma^2} - \frac{|u-\Tilde{v}|^2}{2\sigma^2}\right|\\
&\leq&\frac{|u-v|+|u-\Tilde{v}|}{2\sigma^2} ||u-v|-|u-\Tilde{v}||\\
&\leq& \frac{\sqrt{d}}{\sigma^2}|v-\Tilde{v}|.
\end{eqnarray*}
Thus, \eqref{reg_K} holds with $\alpha=1$ and
\begin{equation}\label{eq:CK_Gaussian}
    C_K=\frac{\sqrt d}{\sigma^2}.
\end{equation}

Under the Fourier-transform convention used in this paper,
\[
    \widehat\phi(\xi)
    =(2\sigma^2\pi)^{d/2}
    e^{-2\sigma^2\pi^2\|\xi\|^2}.
\]
Since $\|\xi\|^2\leq dm^2/4$ on $[-m/2,m/2]^d$,
\begin{equation*}
    \Gamma_m
    \geq
    (2\sigma^2\pi)^{d/2}
    e^{-\sigma^2\pi^2dm^2/2}.
\end{equation*}
Lemma~\ref{lemma:norminverse} then yields
\begin{equation}\label{eq:inverse_Gaussian}
    \bigl\|(K[\Bar{t}])^{-1}\bigr\|_{op}
    \leq
    \frac{(2\sigma^2\pi)^{-d/2}}{m^d}
    e^{\sigma^2\pi^2dm^2/2}.
\end{equation}

Substituting \eqref{eq:CK_Gaussian},
$h_{\Bar{t}}\leq\sqrt d/m$, and
$\sqrt N=(m+1)^{d/2}$ into Lemma~\ref{lemma:reg_G} gives
\begin{align*}
    C_G
    &\leq
    C_F\left(
        1+
        (2\sigma^2\pi)^{-d/2}
        e^{\sigma^2\pi^2dm^2/2}
        \frac{d}{\sigma^2}
        \frac{(m+1)^{d/2}}{m^{d+1}}
    \right)^s \\
    &\leq
    C_FC_{\sigma,d,s}
    e^{\sigma^2\pi^2dsm^2/2},
\end{align*}
where we used $m\geq1$ to absorb the remaining polynomial factor
and the additive constant into $C_{\sigma,d,s}$. This proves the
claim.\hfill\BlackBox

\begin{lemma}[Toroidal Kernel with $\theta_j=a^{\|j\|}$]\label{lemma:C_G_coef1}
Let $p,m\in\NN$, $\mathcal{X}=\mathbb{T}^p$, and $a\in(0,1)$.
Consider the kernel \eqref{kernel_torus} with
$\theta_j=a^{\|j\|}$ and the uniform grid
$\bar{t}=\left(e^{i\frac{2\pi}{m}k}\right)_{k\in \{0,1,\ldots, m-1\}^p} \in \mathcal{X}$ for some $m\in \NN$, and $N=m^p$.
Then the corresponding function $G:[-1,1]^N\to\RR$ is
$s$-H\"{o}lder continuous, and one may take
\begin{equation}
    C_G
    =C_FC_{a,p,s}
    a^{-\frac{m\sqrt p s}{2}}m^{-\frac{ps}{2}-s},
\end{equation}
where $C_{a,p,s}>0$ depends only on $a$, $p$, and $s$.
\end{lemma}

\noindent
{\bf Proof of Lemma \ref{lemma:C_G_coef1}}.
By Remark~\ref{remark:holder_item1}, the kernel is Lipschitz
continuous with
\begin{equation}\label{eq:CK_coef1}
    C_K=\sum_{j\in\ZZ^p}a^{\|j\|}\|j\|<\infty.
\end{equation}
We have
\begin{eqnarray*}
\Theta_{N}= \min_{j\in \left[-\frac{m}{2}, \frac{m}{2}\right]^p} a^{\|j\|}  =  a^{\frac{m\sqrt{p}}{2}} ,\qquad \forall a\in (0,1). 
\end{eqnarray*}
It follows from Lemma \ref{lemma:norminverse_torus} that  $\|(K[\Bar{t}])^{-1}\|_{op} \leq \frac{1}{\Theta_{N} \cdot N} = a^{-\frac{m\sqrt{p}}{2}} m^{-p}$. 
Then, we know
{\allowdisplaybreaks
 \begin{eqnarray*}
&&C_F\left(1+ \|(K[\Bar{t}])^{-1 }\|_{op}\sqrt{N} C_K \left(h_{\Bar{t}}\right)^\alpha\right)^s \\
&\leq& C_F\left(1+ a^{-\frac{m\sqrt{p}}{2}} m^{-p}m^{\frac{p}{2}}\left(\sum_{j\in\ZZ^p} a^{\|j\|}|j| \right)\frac{\sqrt{p}\pi}{m}\right)^s\\
&=& C_F\left(1+ a^{-\frac{m\sqrt{p}}{2}}m^{-\frac{p}{2}-1}\left(\sum_{j\in\ZZ^p} a^{\|j\|}|j| \right)\sqrt{p}\pi\right)^s\\
&=& C_F\left(1+ C_{a,p}a^{-\frac{m\sqrt{p}}{2}}m^{-\frac{p}{2}-1}\right)^s,
\end{eqnarray*}}where $C_{a,p}= \left(\sum_{j\in\ZZ^p} a^{\|j\|}|j| \right)\sqrt{p}\pi $. The sequence $a^{-m\sqrt p/2}m^{-p/2-1}$ has a positive lower
bound over $m\in\NN$. Thus, the additive constant can be absorbed
into a constant depending only on $a$, $p$, and $s$, which gives
\begin{equation*}
    C_G =  C_F C_{a,p,s} a^{-\frac{m\sqrt{p}s}{2}}m^{-\frac{ps}{2}-s}, 
\end{equation*}
where $C_{a,p,s}$ is a positive constant depends only on $a, p$, and $s$.
\hfill \BlackBox

\begin{lemma}[Toroidal Kernel with
$\theta_j=(1+\|j\|^2)^{-\gamma}$]\label{lemma:C_G_coef2}
Let $p,m\in\NN$, $\mathcal{X}=\mathbb{T}^p$, and
$\gamma>(p+1)/2$. Consider the kernel \eqref{kernel_torus} with
\[
    \theta_j=(1+\|j\|^2)^{-\gamma}
\]
and the uniform grid
$\bar{t}=\left(e^{i\frac{2\pi}{m}k}\right)_{k\in \{0,1,\ldots, m-1\}^p} \in \mathcal{X}$ for some $m\in \NN$, and $N=m^p$.
Then the corresponding function $G:[-1,1]^N\to\RR$ is
$s$-H\"{o}lder continuous, and one may take
\begin{equation}
    C_G
    =C_FC_{\gamma,p,s}
    m^{s(2\gamma-\frac p2-1)},
\end{equation}
where $C_{\gamma,p,s}>0$ depends only on $\gamma$, $p$, and $s$.
\end{lemma}

\noindent
{\bf Proof of Lemma \ref{lemma:C_G_coef2}}.
By Remark~\ref{remark:holder_item2}, the kernel is Lipschitz
continuous with
\begin{equation}\label{eq:CK_coef2}
    C_K
    =\sum_{j\in\ZZ^p}
      (1+\|j\|^2)^{-\gamma}\|j\|<\infty.
\end{equation}
We have
\begin{eqnarray*}
\Theta_{N}= \min_{j\in \left[-\frac{m}{2}, \frac{m}{2}\right]^p} \left(1+\|j\|^2\right)^{-\gamma}  =  \left(1+\frac{m^2p}{4}\right)^{-\gamma} ,\qquad \forall \gamma>(p+1)/2, 
\end{eqnarray*}
and Lemma~\ref{lemma:norminverse_torus} yields
\begin{equation}\label{eq:inverse_coef2}
    \bigl\|(K[\Bar{t}])^{-1}\bigr\|_{op}
    \leq
    \left(1+\frac{m^2p}{4}\right)^\gamma m^{-p}.
\end{equation}

Using $\sqrt N=m^{p/2}$,
$h_{\Bar{t}}\leq\sqrt p\,\pi/m$, and \eqref{eq:CK_coef2}, we obtain
\begin{align*}
    C_G &=C_F\left(1+ \|(K[\Bar{t}])^{-1 }\|_{op}\sqrt{N} C_K \left(h_{\Bar{t}}\right)^\alpha\right)^s \\
    &\leq C_F\left(1+ \left(1+\frac{m^2p}{4}\right)^{\gamma} m^{-p}m^{\frac{p}{2}}\left(\sum_{j\in\ZZ^p} (1+\|j\|^2)^{-\gamma}|j|\right)\frac{\sqrt{p}\pi}{m}\right)^s\\
    &\leq C_F\left(1+ m^{2\gamma -\frac{p}{2}-1} p^{\gamma+\frac{1}{2}}  2^{-\gamma}\pi\left(\sum_{j\in\ZZ^p} (1+\|j\|^2)^{-\gamma}|j|\right)\right)^s\\
    &\leq
    C_F\left(
        1+C_{\gamma,p}m^{2\gamma-\frac{p}{2}-1}
    \right)^s.
\end{align*}
Because $\gamma>(p+1)/2$, the exponent
$2\gamma-\frac{p}{2}-1$ is positive. Hence the additive constant can be
absorbed into $C_{\gamma,p,s}$, giving
\[
    C_G
    \leq
    C_FC_{\gamma,p,s}
    m^{s(2\gamma-\frac{p}{2}-1)}.
\]
This completes the proof.
\hfill \BlackBox

\vspace{10pt}

\section{Supporting Corollaries}\label{sec:appendcor}

The following corollaries are obtained by substituting the
kernel-specific H\"{o}lder constants from
Lemmas~\ref{lemma:C_G_Sob}--\ref{lemma:C_G_coef2} into
Theorem~\ref{thm:extension_tim}. Each corollary provides explicit error bounds for approximating the 
$G$ functions using different reproducing kernels: the Sobolev space kernel, the inverse multiquadric kernel, the Gaussian kernel, and kernels on $p$-torus with different coefficients. These results play a key role in deriving the final theorems of the paper.
To avoid repeating the same network
architectures, let $\widehat G_{\mathrm{tanh}}$ and
$\widehat G_{\mathrm{ReLU}}$ denote, respectively, the tanh and ReLU
networks specified in that theorem, and set
\begin{equation*}
    A_{M,L}:=M^2L^2\log_3(M+2).
\end{equation*}
In every result below, the tanh assertion holds for
$M\in\NN$ with $M>5N^2$, whereas the ReLU assertion holds for all
$M,L\in\NN$.

For kernels on $\mathcal{X}=[0,1]^d$, we use
\begin{equation}\label{eq:euclidean_grid_corollaries}
    \Bar t
    =\left\{0,\frac1m,\ldots,\frac{m}{m}\right\}^d,
    \qquad N=(m+1)^d.
\end{equation}
For kernels on $\mathcal{X}=\mathbb T^p$, we use the uniform grid
\begin{equation}\label{eq:torus_grid_corollaries}
    \bar{t}=\left(e^{i\frac{2\pi}{m}k}\right)_{k\in \{0,1,\ldots, m-1\}^p},
    \qquad N=m^p.
\end{equation}

\begin{corollary}[Sobolev kernel]\label{G-Ghat_Sob}
Let $\mathcal H_K=W_2^r([0,1]^d)$ with $r-d/2>1$, and use the
sampling grid \eqref{eq:euclidean_grid_corollaries}. Then
\begin{eqnarray*}
     \|\widehat G_{\mathrm{tanh}}-G\|_{L^\infty([-1,1]^N)} &=&  \left\|\widehat G_{\mathrm{tanh}}-G\right\|_{L^\infty [-1,1]^{(m+1)^{d}}}\\
    &\leq& \frac{(14 (m+1)^{2d} +1)(m+1)^{\frac{d}{2}}}{M^s}C_F C_{r,d,s} m^{2rs-\frac{ds}{2}-s} d^{rs+\frac{s}{2}}\\
     &\leq&  \frac{15 \cdot 2^{\frac{5d}{2}} C_F C_{r,d,s}}{M^s}\   m^{s\left(2r-\frac{d}{2}-1\right) + \frac{5d}{2}} \ d^{s\left(r+\frac{1}{2}\right)},
\end{eqnarray*}
and 
\begin{eqnarray*}
    \|\widehat G_{\mathrm{ReLU}} - G\|_{L^\infty ([-1, 1]^N)} 
    &\leq& 131 \cdot 2^s (m+1)^{\frac{d}{2}}(A_{M,L})^{-\frac{s}{(m+1)^d}}  C_F C_{r,d,s} m^{2rs-\frac{ds}{2}-s}d^{rs+\frac{s}{2}}\\
    &\leq& 131   C_F C_{r,d,s} \cdot 2^{s+\frac{d}{2}} (A_{M,L})^{-\frac{s}{(m+1)^d}}   m^{s(2r-\frac{d}{2}-1) +\frac{d}{2}}d^{s\left(r+\frac{1}{2}\right)}.
\end{eqnarray*}
\end{corollary}

\begin{corollary}[Inverse multiquadric kernel]
\label{G-Ghat_multi}
Let  $K(u,v)=(\sigma^2+\|u-v\|^2)^{-\beta}$ for $\sigma,\beta>0$, 
and use the sampling grid \eqref{eq:euclidean_grid_corollaries}.
Let $M_d=12\left(
        \frac{\pi\Gamma^2((d+2)/2)}{9}
    \right)
    \leq6.38d$.
Then
\begin{align*}
    \|\widehat G_{\mathrm{tanh}}-G\|_{L^\infty([-1,1]^N)}
    &=  \left\|\widehat{G}-G\right\|_{L^\infty [-1,1]^{(m+1)^{d}}} \\
    &\leq \frac{(14 (m+1)^{2d} +1)(m+1)^{\frac{d}{2}}}{M^s} (2m)^{-s\beta-s/2}  C_FC_{\sigma, \beta,d,s}e^{4\sigma M_d sm}\\
    &\leq \frac{15 C_F C_{\sigma, \beta,d,s}}{M^s} (2m)^{-s\beta-\frac{s}{2} + \frac{5d}{2}} e^{4\sigma M_d sm},\\
 \hbox{and}  \quad  \|\widehat G_{\mathrm{ReLU}}-G\|_{L^\infty([-1,1]^N)}
    &\leq 131 \cdot 2^s (m+1)^{\frac{d}{2}}(A_{M,L})^{-\frac{s}{(m+1)^d}}C_FC_{\sigma, \beta,d,s}(2m)^{-s\beta -\frac{s}{2}}e^{4\sigma M_d sm}  \\
    &\leq 131 C_FC_{\sigma, \beta,d,s} \cdot 2^{\frac{d}{2}+s(\frac{1}{2}-\beta)}(A_{M,L})^{-\frac{s}{(m+1)^d}}(m)^{-s\beta -\frac{s}{2}+\frac{d}{2}}e^{4\sigma M_d sm}.
\end{align*}
\end{corollary}

\begin{corollary}[Gaussian kernel]\label{G-Ghat_Gaussian}
Let
\[
    K(u,v)=\exp\left(-\frac{\|u-v\|^2}{2\sigma^2}\right),
    \qquad \sigma>0,
\]
and use the sampling grid \eqref{eq:euclidean_grid_corollaries}.
Then
\begin{align*}
    \|\widehat G_{\mathrm{tanh}}-G\|_{L^\infty([-1,1]^N)}
    &\leq \frac{(14 (m+1)^{2d} +1) (m+1)^{\frac{d}{2}}}{M^s} C_FC_{\sigma,d,s}e^{\sigma^2\pi^2 dsm^2/2} \\
     &\leq \frac{(2m)^{\frac{5d}{2}}}{M^s} 15C_FC_{\sigma,d,s}e^{\sigma^2\pi^2 dsm^2/2},\\
   \hbox{and} \quad \|\widehat G_{\mathrm{ReLU}}-G\|_{L^\infty([-1,1]^N)}
    &\leq  131 \cdot 2^s (m+1)^{\frac{d}{2}}(A_{M,L})^{-\frac{s}{(m+1)^d}}C_FC_{\sigma,d,s}e^{\sigma^2\pi^2 dsm^2/2}\\
     &\leq  131 C_FC_{\sigma,d,s} \cdot 2^{s+\frac{d}{2}} m^{\frac{d}{2}}(A_{M,L})^{-\frac{s}{(m+1)^d}}e^{\sigma^2\pi^2 dsm^2/2}.
\end{align*}
\end{corollary}

\begin{corollary}[Toroidal kernel with
$\theta_j=a^{\|j\|}$]\label{G-Ghat_coef1}
Let $a\in(0,1)$, consider the kernel \eqref{kernel_torus} with
$\theta_j=a^{\|j\|}$, and use the sampling grid
\eqref{eq:torus_grid_corollaries}. Then
\begin{align*}
    \|\widehat G_{\mathrm{tanh}}-G\|_{L^\infty([-1,1]^N)}
    &\leq \frac{(14m^{2p}+1)m^{\frac{p}{2}}}{M^s}C_F C_{a,p,s} a^{-\frac{m\sqrt{p}s}{2}}m^{-\frac{ps}{2}-s}\\
   &\leq \frac{15C_F C_{a,p,s}}{M^s} m^{\frac{p}{2}(5-s) -s }a^{-\frac{m\sqrt{p}s}{2}}\\
 \hbox{and} \quad  \|\widehat G_{\mathrm{ReLU}}-G\|_{L^\infty([-1,1]^N)}
 &\leq 131 \cdot 2^s  m^{\frac{p}{2}}  (A_{M,L})^{-\frac{s}{m^p}}C_FC_{a,p,s} a^{-\frac{m\sqrt{p}s}{2}}m^{-\frac{ps}{2}-s}\\
    &\leq 131 C_FC_{a,p,s} \cdot 2^s \ a^{-\frac{m\sqrt{p}s}{2}}  \ m^{\frac{p}{2}(1-s)-s}   ((A_{M,L})^{-\frac{s}{m^p}}.
\end{align*}
\end{corollary}

\begin{corollary}[Toroidal kernel with
$\theta_j=(1+\|j\|^2)^{-\gamma}$]\label{G-Ghat_coef2}
Let $\gamma>(p+1)/2$, consider the kernel \eqref{kernel_torus} with
$\theta_j=(1+\|j\|^2)^{-\gamma}$, and use the sampling grid
\eqref{eq:torus_grid_corollaries}. Then
\begin{align*}
    \|\widehat G_{\mathrm{tanh}}-G\|_{L^\infty([-1,1]^N)}
    &\leq \frac{(14 m^{2p}+1)m^{\frac{p}{2}}}{M^s}C_F C_{\gamma,p,s} m^{s(2\gamma-\frac{p}{2}-1)}\\
   &\leq \frac{15C_F C_{\gamma,p,s}}{M^s} m^{\frac{5p}{2}+s(2\gamma-\frac{p}{2}-1)},\\
 \hbox{and} \quad  \|\widehat G_{\mathrm{ReLU}}-G\|_{L^\infty([-1,1]^N)}
 &\leq131 \cdot 2^s  m^{\frac{p}{2}}    (A_{M,L})^{-\frac{s}{m^p}} C_F C_{\gamma,p,s} m^{2\gamma s -\frac{ps}{2}-s}\\
    &\leq 131 C_F C_{\gamma,p,s} \cdot 2^s \  m^{\frac{p}{2}+s(2\gamma-\frac{p}{2}-1)}  (A_{M,L})^{-\frac{s}{m^p}}.
\end{align*}
\end{corollary}

\section{Proof of Main Results}\label{sec:appendB}

We are in a position to prove our main results. 
We first present the proof of Theorem \ref{thm:Sobolev} below.

\noindent
{\bf Proof of Theorem \ref{thm:Sobolev}}.

    Recall that from the Sobolev Embedding Theorem, we know that the reproducing kernel $K$ in $W^r_2(\mathcal{X})$ is $\alpha$-H\"{o}lder continuous with $\alpha = r-d/2$ and the constant $C_K = C_{r,d}>0$ depends only on $r$ and $d$. In our analysis, we require  $\alpha \leq 1$, which is why we assume  $\alpha \leq 1$ throughout the paper. Thus, we take $\alpha =\min\{1,r-d/2\}$.
    
    We consider the error decomposition in (\ref{decomposition}), that is,
    \begin{eqnarray*}
         \sup_{f\in \mathcal{K}}\left| F(f) - \widehat{G}(f(\Bar{t}))\right| &\leq  \sup_{f\in \mathcal{K}}| F(f) - F(Pf)| +  \sup_{f\in \mathcal{K}}|F(Pf) - \widehat{G}(f(\Bar{t}))| \\
        &=  \sup_{f\in \mathcal{K}}| F(f) - F(Pf)| +  \sup_{f\in \mathcal{K}}|G(f(\Bar{t})) - \widehat{G}(f(\Bar{t}))|.
    \end{eqnarray*}
Here, we have used $G(f(\Bar{t})) = F\left(\sum_{i=1}^N f(t_i) \psi_i\right) = F(Pf)$, where the function $G$ is defined previously in equation (\ref{G}). 

According to \citep[Theorem 6.5.3]{schaback1997reconstruction}, when $r>d/2$ and $r-d/2 \notin \NN$, an upper bound of the power function is derived from the well-known Sobolev Embedding Theorem as 
\begin{equation*}
\epsilon_K(\Bar{t})  \leq C^\prime_{r,d}d^{r-\frac{d}{2}}m^{d-2r},
\end{equation*}
where $C^\prime_{r,d}$ is some positive constant depending on $r$ and $d$. 
Thus, we have 
   \begin{equation}\label{f-Pf_Sobolev}
      \|f-Pf\|_\infty \leq \epsilon_K(\Bar{t})\|f\|_{\mathcal{H}_K} \leq C^\prime_{r,d}d^{r-\frac{d}{2}}m^{d-2r}
   \end{equation} 
   and 
   \begin{equation*}
      \sup_{f\in \mathcal{K}}| F(f) - F(Pf)| \leq  C_F \|f-Pf\|^s_\infty \leq  C_F  (C^\prime_{r,d})^s d^{s(r-\frac{d}{2})}m^{s(d-2r)}. 
   \end{equation*}

\vspace{10pt}
\noindent \underline{\textbf{Tanh network} }  

From Corollary \ref{G-Ghat_Sob}, for every $M\in \NN$ with $M> 5N^2 = 5(m+1)^{2d}$, there exists a tanh neural network $\widehat{G}$ of the stated size such that
   \begin{eqnarray*}
       \sup_{f\in \mathcal{K}}\left|G(f(\Bar{t})) - \widehat{G}(f(\Bar{t}))\right| \leq \left\|\widehat{G}-G\right\|_{L^\infty [-1,1]^{(m+1)^{d}}}
        &\leq&\highlight{\frac{15 \cdot 2^{\frac{5d}{2}} C_F C_{r,d,s}}{M^s}\   m^{s\left(2r-\frac{d}{2}-1\right) + \frac{5d}{2}} \ d^{s\left(r+\frac{1}{2}\right)}.}
   \end{eqnarray*}

   Now combining the two error bounds together, we get 
   \begin{eqnarray*}  
   && \sup_{f\in  \mathcal{K}}\left| F(f) - \widehat{G}(f(\Bar{t}))\right| \\
  &\leq& C_F (C^\prime_{r,d})^s d^{s(r-\frac{d}{2})}m^{s(d-2r)} + \highlight{\frac{15 \cdot 2^{\frac{5d}{2}} C_F C_{r,d,s}}{M^s}\   m^{s\left(2r-\frac{d}{2}-1\right) + \frac{5d}{2}} \ d^{s\left(r+\frac{1}{2}\right)}}\\
  &\leq&  \highlight{C_F\max\{(C^\prime_{r,d})^s, 15 \cdot 2^{\frac{5d}{2}} C_{r,d,s} \}\left(d^{s(r-\frac{d}{2})}m^{s(d-2r)} + \frac{1}{M^s}m^{s\left(2r-\frac{d}{2}-1\right) + \frac{5d}{2}}d^{s(r+\frac{1}{2})}\right).} 
   \end{eqnarray*}
   \highlight{We see that $m^{s(d-2r)}$ is a decreasing function of $m$, whereas $m^{s\left(2r-\frac{d}{2}-1\right) + \frac{5d}{2}}$ is an increasing one. We choose the smallest integer $m$ such that $m^{s(d-2r)} \leq \frac{1}{M^s}m^{s\left(2r-\frac{d}{2}-1\right)+ \frac{5d}{2}}$; that is }
   \highlight{
   \begin{eqnarray*}
       &&m^{s(d-2r-2r+\frac{d}{2} +1)-\frac{5d}{2}} \leq \frac{1}{M^s} \\
       &\Leftrightarrow& m^{s(4r  -\frac{3d}{2}-1)+\frac{5d}{2}} \geq M^s\\
       &\Leftrightarrow& m \geq M^{\frac{s}{s(4r -\frac{3d}{2}-1)+\frac{5d}{2}}}.
   \end{eqnarray*}}
   So we take $m=\left\lceil M^{\frac{s}{s(4r -\frac{3d}{2}-1)+\frac{5d}{2}}}\right\rceil \in \NN$. Notice that $$s(4r -\frac{3d}{2}-1)+\frac{5d}{2} \geq s(4r +d-1) >3s(d+1).$$ 
   We see that $m < \left\lceil M^{\frac{1}{3(d+1)}}\right\rceil$ and
   $5N^2 = 5(m+1)^{2d} < 5\left(M^{\frac{1}{3(d+1)}}+2\right)^{2d}$. Here $M>5N^2$ is satisfied when $M> (5\cdot 4^d)^3$.
   Then,
   \begin{eqnarray*}
       \frac{1}{M^s}m^{s\left(2r-\frac{d}{2}-1\right)+ \frac{5d}{2}} \leq 2M^{\frac{s^2(2r-\frac{d}{2}-1)+ \frac{5}{2}ds}{s(4r -\frac{3d}{2}-1)+\frac{5d}{2}}-s} 
       &=& 2M^{\frac{s^2(d-2r)}{s(4r -\frac{3d}{2}-1)+\frac{5d}{2}}}\\
       &=& 2\left(\frac{1}{M}\right)^{\frac{s^2(2r-d)}{s(4r -\frac{3d}{2}-1)+\frac{5d}{2}}}.
   \end{eqnarray*}
   We also know that $d^{s(r-\frac{d}{2})} \leq d^{s(r+\frac{1}{2})}$. \highlight{Set $C_{r,s,d,F} = C_Fd^{s(r+\frac{1}{2})}\max\{(C^\prime_{r,d})^s, 15 \cdot 2^{\frac{5d}{2}} C_{r,d,s} \}$.} The stated bound for the tanh network is proved.

   \vspace{10pt}
  \second{ \noindent \underline{\textbf{ReLU network} }
  
  From Corollary \ref{G-Ghat_Sob}, 
for any $M, L \in \NN \setminus\{0\}$, there exists a ReLU network $\widehat{G}$ of the stated size such that
  \begin{eqnarray*}
    \|\widehat{G} - G\|_{L^\infty ([-1, 1]^N)} 
    &\leq& 131   C_F C_{r,d,s} \cdot 2^{s+\frac{d}{2}} (M^2 L^2 \log_3(M+2))^{-\frac{s}{(m+1)^d}}   m^{s(2r-\frac{d}{2}-1) +\frac{d}{2}}d^{s\left(r+\frac{1}{2}\right)}.
\end{eqnarray*}

Now combining the two error bounds together, we get 
\begin{eqnarray*}  
   && \sup_{f\in  \mathcal{K}}\left| F(f) - \widehat{G}(f(\Bar{t}))\right| \\
   &\leq& C_F (C^\prime_{r,d})^s d^{s(r-\frac{d}{2})}m^{s(d-2r)} \\
   &&+ 131   C_F C_{r,d,s} \cdot 2^{s+\frac{d}{2}} (M^2 L^2 \log_3(M+2))^{-\frac{s}{(m+1)^d}}   m^{s(2r-\frac{d}{2}-1) +\frac{d}{2}}d^{s\left(r+\frac{1}{2}\right)}\\
  &\leq& C_F\max\{(C^\prime_{r,d})^s, 131 \cdot 2^{s+\frac{d}{2}} C_{r,d,s} \}\\
  && \cdot \left(d^{s(r-\frac{d}{2})}m^{s(d-2r)} + (M^2 L^2 \log_3(M+2))^{-\frac{s}{(m+1)^d}} m^{s\left(2r-\frac{d}{2}-1\right) + \frac{d}{2}}d^{s(r+\frac{1}{2})}\right).
   \end{eqnarray*}

   Let
\[
A:=M^2L^2\log_3(M+2)
\]
and define
\[
\alpha:=s(2r-d),
\qquad
\beta:=s\left(2r-\frac d2-1\right)+\frac d2.
\]
Since \(r>d/2\), we have \(\alpha>0\). We also have $\beta>0$ under $r>d/2$ and $s \in (0,1]$. 
The total approximation error is bounded by
\begin{eqnarray*}
&& \sup_{f\in\mathcal K}\left|F(f)-\widehat G(f(\bar t))\right| \\
&\leq&
C_F\max\left\{(C'_{r,d})^s,\;131\cdot 2^{s+\frac d2}C_{r,d,s}\right\}
\left(
d^{s(r-\frac d2)}m^{-\alpha}
+
A^{-\frac{s}{(m+1)^d}}m^\beta d^{s(r+\frac12)}
\right).
\end{eqnarray*}
Since $d^{s(r-\frac d2)}\leq d^{s(r+\frac12)}$,
we obtain
\begin{eqnarray*}
\sup_{f\in\mathcal K}\left|F(f)-\widehat G(f(\bar t))\right| 
\leq
C_F d^{s(r+\frac12)}
\max\left\{(C'_{r,d})^s,\;131\cdot 2^{s+\frac d2}C_{r,d,s}\right\}
\left(
m^{-\alpha}
+
A^{-\frac{s}{(m+1)^d}}m^\beta
\right).
\end{eqnarray*}
Next, we define
\[
\gamma:=\alpha+\beta
=
s\left(4r-\frac{3d}{2}-1\right)+\frac d2.
\]
We can check that $\gamma>0$.
We now choose \(m\) to balance the two terms: $m^{-\alpha}$ and $A^{-\frac{s}{(m+1)^d}}m^\beta$. Let \(m\in\mathbb N\) be an integer satisfying
\begin{eqnarray*}
  && m^{-\alpha} \geq A^{-\frac{s}{(m+1)^d}}m^\beta \\
&\Leftrightarrow& m^\gamma = m^{\alpha+\beta}
\leq
A^{\frac{s}{(m+1)^d}}.
\end{eqnarray*}
Taking logarithms, this is equivalent to
\[
(m+1)^d\log m
\leq
\frac{s}{\gamma}\log A.
\]

Since $\log_3(M+2) \geq 1$ and $\log m \leq \log(m+1)$, it is sufficient to require $(m+1)^d \log(m+1) \leq \frac{2s}{\gamma} \log(ML)$. Observe that for $B>1$, the number $x= \left(\frac{B}{\log B}\right)^{\frac{1}{d}}$ satisfies $\log x \leq \frac{1}{d} \log B$ and the inequality $x^d \log x \leq \frac{B}{\log B} \cdot \frac{1}{d} \log B =\frac{B}{d}$. Thus, when $ML > e^{\frac{\gamma}{2sd}}$, we can take $B= \frac{2sd}{\gamma} \log(ML)>1$ and see that 
\begin{eqnarray*}
     m = \left\lfloor\left(\frac{B}{\log B}\right)^{\frac{1}{d}} \right\rfloor -1
\end{eqnarray*}
satisfies
\begin{eqnarray*}
    (m+1)^d \log(m+1) \leq \frac{2s}{\gamma} \log(ML) \qquad \hbox{and thereby} \qquad m^{-\alpha}
\geq A^{-\frac{s}{(m+1)^d}}m^\beta.
\end{eqnarray*}
Moreover, when $B \geq 4^d$, we have $m \geq 1$ and $$m^{-\alpha} \leq \left(\frac{m+2}{3}\right)^{-\alpha} \leq 3^\alpha \left(\frac{B}{\log B}\right)^{-\frac{\alpha}{d}}.$$
This proves the stated bound with 
$$C_{r,s,d,F} = 2\cdot 3^{\alpha} C_F d^{s(r+\frac{1}{2})} \max\{(C^\prime_{r,d})^s, 131 \cdot 2^{s+\frac{d}{2}} C_{r,d,s}\}.$$
   \hfill\BlackBox}

Next, we present the proof of Theorem \ref{thm:multi}.

\noindent
{\bf Proof of Theorem \ref{thm:multi}}.

We consider the error decomposition in (\ref{decomposition}), that is, 
\begin{eqnarray*}
    \sup_{f\in \mathcal{K}}\left| F(f) - \widehat{G}(f(\Bar{t}))\right| &\leq  \sup_{f\in \mathcal{K}}| F(f) - F(Pf)| +  \sup_{f\in \mathcal{K}}|F(Pf) - \widehat{G}(f(\Bar{t}))| \\
    &=  \sup_{f\in \mathcal{K}}| F(f) - F(Pf)| +  \sup_{f\in \mathcal{K}}|G(f(\Bar{t})) - \widehat{G}(f(\Bar{t}))|.
\end{eqnarray*}
Here, we have used $G(f(\Bar{t})) = F\left(\sum_{i=1}^N f(t_i) \psi_i\right) = F(Pf)$, where the function $G$ is defined previously in equation (\ref{G}).

For the inverse multiquadric kernel given in (\ref{multi}), an upper bound of $\epsilon_K(\Bar{t})$ and $\|f-Pf\|_\infty$ are established in  \citep[Equation (15)]{fasshauer2005meshfree} for some positive constant $c$: 
\begin{equation}\label{f-Pf_multi}
\epsilon_K(\Bar{t}) \leq e^{\frac{-cm}{\sqrt{d}}} \quad \text{and thus }\quad  
    \|f-Pf\|_\infty \leq e^{\frac{-cm}{\sqrt{d}}}\|f\|_{\mathcal{H}_K}, \qquad f\in  \mathcal{H}_K.
\end{equation}
With the assumption that $F$ is  $s$-H\"{o}lder continuous with constant $C_F \geq 0$, we have
\begin{equation*}
      \sup_{f\in \mathcal{K}}| F(f) - F(Pf)| \leq  C_F \|f-Pf\|^s_\infty \leq  C_F e^{\frac{-csm}{\sqrt{d}}}\|f\|^s_{\mathcal{H}_K} \leq C_F e^{\frac{-csm}{\sqrt{d}}}. 
\end{equation*}

\vspace{10pt}
\noindent \underline{\textbf{Tanh network} }  

From Corollary \ref{G-Ghat_multi}, \highlight{for every $M\in \NN$ with $M>5N^2 = 5(m+1)^{2d}$,} we get 
\begin{eqnarray*}
       \sup_{f\in \mathcal{K}}\left|G(f(\Bar{t})) - \widehat{G}(f(\Bar{t}))\right| &\leq& \left\|\widehat{G}-G\right\|_{L^\infty [-1,1]^{(m+1)^{d}}}\\
       &\leq& \highlight{ \frac{15 C_F C_{\sigma, \beta,d,s}}{M^s} (2m)^{\frac{5d}{2}-\frac{s}{2} } e^{4\sigma M_d sm}}.
\end{eqnarray*}
Then, combining the above two error bounds together, we get
\begin{eqnarray*}
       \sup_{f\in \mathcal{K}}\left| F(f) - \widehat{G}(f(\Bar{t}))\right|
       & \leq& C_F e^{\frac{-csm}{\sqrt{d}}} +  \highlight{ \frac{15 C_F C_{\sigma, \beta,d,s}}{M^s} (2m)^{\frac{5d}{2}-\frac{s}{2} } e^{4\sigma M_d sm}}\\
        & \leq& \highlight{15C_F C_{\sigma, \beta,d,s} \left( e^{\frac{-csm}{\sqrt{d}}} +   \frac{(2m)^{\frac{5d}{2}-\frac{s}{2} }}{M^s}  e^{4\sigma M_d sm}\right).}
\end{eqnarray*}
We can see that there is a trade-off of $m$ between $e^{\frac{-csm}{\sqrt{d}}}$ and $\frac{(2m)^{\highlight{\frac{5d}{2}-\frac{s}{2}}}}{M^\highlight{s}}e^{4M_d\sigma sm}$. To balance the trade-off, we choose the smallest integer $m\in \NN$ such that $e^{\frac{-csm}{\sqrt{d}}} \leq  \frac{1}{M^\highlight{s}} e^{4M_d\sigma sm}$, which means
\highlight{\begin{eqnarray} \label{ineq:logM}
s\log (M) \leq 4M_d\sigma sm + \frac{csm}{\sqrt{d}} = m \left(4M_d\sigma s + \frac{cs}{\sqrt{d}}\right).
    \end{eqnarray}}
We take $m=\left\lceil\frac{\log (M)}{4M_d\sigma + \frac{c}{\sqrt{d}}}\right\rceil \in \NN$. 
\highlight{We know $m-1$ does not satisfy the above inquality \eqref{ineq:logM}, meaning that $e^{\frac{-cs(m-1)}{\sqrt{d}}} >  \frac{1}{M^\highlight{s}} e^{4M_d\sigma s(m-1)}$. It follows that 
\begin{eqnarray*}
    \frac{1}{M^\highlight{s}} e^{4M_d\sigma sm} = \frac{1}{M^\highlight{s}} e^{4M_d\sigma s(m-1)} \cdot  e^{4M_d\sigma s}  < e^{4M_d\sigma s} \cdot e^{\frac{-cs(m-1)}{\sqrt{d}}}
\end{eqnarray*}
and 
{\allowdisplaybreaks
\begin{eqnarray*}
     e^{\frac{-csm}{\sqrt{d}}} +   \frac{(2m)^{\frac{5d}{2}-\frac{s}{2} }}{M^s}  e^{4\sigma M_d sm}&\leq&  \frac{1}{M^\highlight{s}} e^{4M_d\sigma sm} \left(1+(2m)^{\frac{5d}{2}-\frac{s}{2} }\right)\\
     &\leq& e^{4M_d\sigma s} \cdot e^{\frac{-cs(m-1)}{\sqrt{d}}}  \left(1+(2m)^{\frac{5d}{2}-\frac{s}{2} }\right)\\
      &\leq& e^{4M_d\sigma s+ \frac{cs}{\sqrt{d}}} \cdot e^{-\frac{csm}{\sqrt{d}}}  \left(1+(2m)^{\frac{5d}{2}-\frac{s}{2} }\right)\\
      &\leq& e^{4M_d\sigma s+ \frac{cs}{\sqrt{d}}} \cdot e^{-\frac{cs}{\sqrt{d}} \left(\frac{\log (M)}{4M_d\sigma + \frac{c}{\sqrt{d}}}\right)}\left(1+\left(\frac{4\log (M)}{4M_d\sigma + \frac{c}{\sqrt{d}}}\right)^{\frac{5d}{2}-\frac{s}{2} }\right)    \\
      &=& M^{-\frac{cs}{\sqrt{d}}\frac{1}{4M_d \sigma+\frac{c}{\sqrt{d}}}} \cdot e^{4M_d\sigma s+ \frac{cs} {\sqrt{d}}} \left(1+\left(\frac{4\log(M)}{4M_d\sigma + \frac{c}{\sqrt{d}}}\right)^{\frac{5d}{2}-\frac{s}{2} }\right),
\end{eqnarray*}}
which yields the desired bound with $$C_{\sigma,\beta,s,d,F}=15C_F C_{\sigma, \beta,d,s} e^{4M_d\sigma s+ \frac{cs} {\sqrt{d}}} \left(1+  \left(\frac{4}{4M_d\sigma + \frac{c}{\sqrt{d}}}\right)^{\frac{5d}{2}-\frac{s}{2} } \right).$$

Additionally, the restriction $M> 5(m+1)^{2d}$ is satisfied when $M>M_0$, where $M_0$ is the largest positive solution to the inequality $
\log x \geq \frac{4M_d\sigma +\frac{c}{\sqrt{d}} }{3}\left(\frac{x}{5}\right)^{\frac{1}{2d}}$.
}

\second{
\vspace{10pt}
\noindent \underline{\textbf{ReLU network}} 

From Corollary \ref{G-Ghat_multi}, for any $M, L \in \NN$, there exists a ReLU network $\widehat{G}$ of the stated size such that 
\begin{eqnarray*}
   && \|\widehat{G} - G\|_{L^\infty ([-1, 1]^N)} \\   
    &\leq& 131 C_FC_{\sigma, \beta,d,s} \cdot 2^{\frac{d}{2}+s(\frac{1}{2}-\beta)}(M^2 L^2 \log_3(M+2))^{-\frac{s}{(m+1)^d}}m^{-s\beta -\frac{s}{2}+\frac{d}{2}}e^{4\sigma M_d sm}
\end{eqnarray*}
Combining two error bounds together,
\begin{eqnarray*}
       &&\sup_{f\in \mathcal{K}}\left| F(f) - \widehat{G}(f(\Bar{t}))\right|\\
       & \leq& C_F e^{\frac{-csm}{\sqrt{d}}} +  131 C_FC_{\sigma, \beta,d,s} \cdot 2^{\frac{d}{2}+s(\frac{1}{2}-\beta)}(M^2 L^2 \log_3(M+2))^{-\frac{s}{(m+1)^d}}m^{-s\beta -\frac{s}{2}+\frac{d}{2}}e^{4\sigma M_d sm}\\
       & \leq& 131 C_FC_{\sigma, \beta,d,s} \cdot 2^{\frac{d}{2}+s(\frac{1}{2}-\beta)} \left( e^{\frac{-csm}{\sqrt{d}}} +(M^2 L^2 \log_3(M+2))^{-\frac{s}{(m+1)^d}}m^{-s\beta -\frac{s}{2}+\frac{d}{2}}e^{4\sigma M_d sm}\right).
\end{eqnarray*}
Let $A:=M^2L^2\log_3(M+2)$.
For notational simplicity, define
\[
a:=\frac{cs}{\sqrt d},
\qquad
b:=4\sigma M_ds,
\qquad
\kappa:=a+b
=s\left(4\sigma M_d+\frac{c}{\sqrt d}\right),
\]
and
\[
\rho:=s\beta+\frac{s}{2}+\frac{d}{2}=s\left(\beta+\frac12\right)+\frac{d}{2}.
\]
Then the approximation error can be written as
\begin{eqnarray*}
\sup_{f\in\mathcal K}
\left|F(f)-\widehat G(f(\bar t))\right|
\leq
131 C_FC_{\sigma,\beta,d,s}
\cdot 2^{\frac d2+s(\frac12-\beta)}\left(
T_1(m)+T_2(m)
\right),
\end{eqnarray*}
where
\[
T_1(m):=e^{-am}
\qquad \hbox{and} \qquad
T_2(m):=
A^{-\frac{s}{(m+1)^d}}m^{-\rho}e^{bm}.
\]

We choose \(m\) to be an integer such that
$T_1(m)\geq T_2(m)$, that is 
\begin{eqnarray*}
  &&  e^{-am}
\geq
A^{-\frac{s}{(m+1)^d}}m^{-\rho}e^{bm}\\
&\Leftrightarrow& -am
\geq
-\frac{s\log A}{(m+1)^d}
-\rho\log m
+bm\\
&\Leftrightarrow& (m+1)^d (\kappa m-\rho\log m)
\leq
s\log A\\ 
&\Leftrightarrow& (m+1)^d
\left[
s\left(4\sigma M_d+\frac{c}{\sqrt d}\right)m
-
s\left(\beta+\frac12\right)\log m
\right]
\leq
s\log A\\
&\Leftrightarrow& (m+1)^d
\left[
\left(4\sigma M_d+\frac{c}{\sqrt d}\right)m
-
\left(\beta+\frac12\right)\log m
\right]
\leq
\log A.
\end{eqnarray*}
Since $\log_3(M+2) \geq 1$, it is sufficient to find an integer $m\in\NN$ satisfying $$(m+1)^d  \left(4\sigma M_d+\frac{c}{\sqrt d}\right)(m+1) \leq 2 \log(ML).$$ 
So, we take
\begin{eqnarray*}
   m
= \left\lfloor \left(\frac{2\log(ML)}{4\sigma M_d +c/\sqrt{d}}\right)^{\frac{1}{d+1}}\right\rfloor -1.
\end{eqnarray*}
When $ML \geq e^{2^d \left(4\sigma M_d+\frac{c}{\sqrt d}\right)}$, we have $m\in\NN$, $m \geq \frac{m+1}{2} \geq \frac{1}{4}\left(\frac{2\log(ML)}{4\sigma M_d +c/\sqrt{d}}\right)^{\frac{1}{d+1}}$, and 
\begin{eqnarray*}
\sup_{f\in\mathcal K}
\left|F(f)-\widehat G(f(\bar t))\right|
\leq 262 C_F C_{\sigma,\beta,d,s} 2^{\frac{d}{2}+ s(\frac{1}{2}-\beta)} e^{-am}.
\end{eqnarray*}
The desired bound is verified.}
The proof is complete. \hfill\BlackBox

Next, we present the proof of Theorem \ref{thm:Gaussian}. 

\noindent
{\bf Proof of Theorem \ref{thm:Gaussian}}.

     Again, we use the error decomposition in (\ref{decomposition}).
   Recall that $F$ is  $s$-H\"{o}lder continuous with constant $C_F \geq 0$.
   For the Gaussian kernel given in (\ref{Gaussian}), an upper bound of $\epsilon_K(\Bar{t})$ and $\|f-Pf\|_\infty$ are established in  \citep[Equation (14)]{fasshauer2005meshfree} for some positive constant $c$: 
\begin{equation*}
\epsilon_K(\Bar{t}) \leq e^{\frac{-cm\left|\log \left(\frac{\sqrt{d}}{m}\right)\right|}{\sqrt{d}}} \quad \text{and thus }\quad
    \|f-Pf\|_\infty \leq e^{\frac{-cm\left|\log \left(\frac{\sqrt{d}}{m}\right)\right|}{\sqrt{d}}}\|f\|_{\mathcal{H}_K}, \qquad f\in  \mathcal{H}_K.
\end{equation*}
Suppose $m>\sqrt{d}$, we have $\left|\log \left(\frac{\sqrt{d}}{m}\right)\right| = \log \left(\frac{m}{\sqrt{d}}\right)$ and thus $\|f-Pf\|_\infty \leq e^{\frac{-cm\log \left(\frac{m}{\sqrt{d}}\right)}{\sqrt{d}}}\|f\|_{\mathcal{H}_K}$, for all  $f\in  \mathcal{H}_K$.
It follows that
       \begin{equation*}
      \sup_{f\in \mathcal{K}}| F(f) - F(Pf)| \leq  C_F \|f-Pf\|^s_\infty \leq  C_F e^{\frac{-csm\log \left(\frac{m}{\sqrt{d}}\right)}{\sqrt{d}}}\|f\|^s_{\mathcal{H}_K} \leq C_F e^{\frac{-csm\log \left(\frac{m}{\sqrt{d}}\right)}{\sqrt{d}}}. 
   \end{equation*}

\vspace{10pt}
\noindent \underline{\textbf{Tanh network} }  

   With Corollary \ref{G-Ghat_Gaussian}, \highlight{for every $M\in \NN$ with $M>5N^2 = 5(m+1)^{2d}$,} we have 
   \begin{eqnarray*}
       \sup_{f\in \mathcal{K}}\left|G(f(\Bar{t})) - \widehat{G}(f(\Bar{t}))\right| \leq \left\|\widehat{G}-G\right\|_{L^\infty [-1,1]^{(m+1)^{d}}}
        &\leq& \highlight{\frac{(2m)^{\frac{5d}{2}}}{M^s} 15 C_FC_{\sigma,d,s}e^{\sigma^2\pi^2 dsm^2/2}.}
   \end{eqnarray*}
   
Combining the above two error bounds, we get
 \begin{eqnarray*}
       \sup_{f\in \mathcal{K}}\left| F(f) - \widehat{G}(f(\Bar{t}))\right|
            \leq  \highlight{15} C_F (1+C_{\sigma,d,s}) \left(e^{\frac{-csm\log \left(\frac{m}{\sqrt{d}}\right)}{\sqrt{d}}} + \frac{(2m)^{\highlight{\frac{5d}{2}}}}{M^\highlight{s}} e^{\sigma^2\pi^2 dsm^2/2}\right).
   \end{eqnarray*}
  \highlight{To trade-off the two terms in the sum, we take 
   $$m = \left\lfloor \frac{\sqrt{2}}{\sigma \pi \sqrt{d}}\left(\sqrt{\log M} - c (\log M)^{\frac{1}{4}}\right) \right\rfloor.$$
   Then, there exists some $M_0\in\NN $ with $M_0 \geq e^{(2\sigma \pi d)^4}$ such that for every $M>M_0$, we have $m>\sqrt{d}$, $M> 5(m+1)^{2d}$, and $m \geq \frac{1}{2\sigma \pi } \sqrt{\frac{\log M}{d}}$. It follows that
   {\allowdisplaybreaks
   \begin{eqnarray*}
        \frac{1}{M^s}e^{\sigma^2\pi^2 dsm^2/2} &\leq&  \frac{1}{M^s}e^{\sigma^2\pi^2 ds \cdot \left(\frac{\sqrt{2}}{\sigma \pi \sqrt{d}}\right)^2\left(\sqrt{\log M} - c (\log M)^{\frac{1}{4}}\right)^2}\\
       &=& \frac{1}{M^s}e^{s\left(\sqrt{\log M} - c (\log M)^{\frac{1}{4}}\right)^2}\\
       &=& e^{s\left(-2c(\log M)^{\frac{3}{4}}+ c^2 \sqrt{\log M}\right)}
   \end{eqnarray*}}
   and 
   \begin{eqnarray*}
       (2m)^{\frac{5d}{2}} \leq \left(\frac{2\sqrt{2}}{\sigma \pi \sqrt{d}} \sqrt{\log M}\right)^{\frac{5d}{2}}.
   \end{eqnarray*}
   Then, 
   \begin{eqnarray*}
   \frac{(2m)^{\frac{5d}{2}}}{M^s} e^{\sigma^2\pi^2 dsm^2/2} \leq \Tilde{C}_{\sigma,s,d}\ e^{-cs (\log M)^{\frac{3}{4}}}, \qquad \forall M >M_0,
     \end{eqnarray*}
     where $\Tilde{C}_{\sigma,s,d}$ is a constant depending only on $\sigma,s,d$. 
     
     With the chosen $m$, we also have 
     \begin{eqnarray*}
         e^{\frac{-csm\log \left(\frac{m}{\sqrt{d}}\right)}{\sqrt{d}}} &\leq& \exp\left(-\frac{cs}{\sqrt{d}} \frac{\sqrt{2}}{2\sigma \pi } \sqrt{\frac{\log M}{d}} \log \left( \frac{1}{2\sigma \pi } \frac{\sqrt{\log M}}{d}\right)\right)\\
         &=&\exp\left(-\frac{cs}{\sqrt{2}\sigma \pi d} \sqrt{\log M}  \left(\frac{1}{2}\log (\log (M)) - \log (\sqrt{2}\sigma \pi d)\right)\right).
     \end{eqnarray*}
     Since $\log (2\sigma \pi d) \leq \frac{1}{4} \log (\log (M))$ for $M > M_0$, we see that
     \begin{equation*}
         e^{\frac{-csm\log \left(\frac{m}{\sqrt{d}}\right)}{\sqrt{d}}} \leq \exp\left(-\frac{cs}{8\sigma \pi d} \sqrt{\log M}  \log (\log (M))\right).
     \end{equation*}
     Finally, we have 
     {\allowdisplaybreaks
      \begin{eqnarray*}
       &&\sup_{f\in \mathcal{K}}\left| F(f) - \widehat{G}(f(\Bar{t}))\right|\\
            &\leq&  15 C_F (1+C_{\sigma,d,s}) \left(e^{\frac{-csm\log \left(\frac{m}{\sqrt{d}}\right)}{\sqrt{d}}} + \frac{(2m)^{\frac{5d}{2}}}{M^s} e^{\sigma^2\pi^2 dsm^2/2}\right)\\
            &\leq& 15 C_F (1+C_{\sigma,d,s})  \left(\exp\left(-\frac{cs}{8\sigma \pi d} \sqrt{\log M}  \log (\log (M))\right) + \Tilde{C}_{\sigma,s,d}\ e^{-cs (\log M)^{\frac{3}{4}}} \right)\\
            &\leq& C_{\sigma,c,d,s, F} \exp\left(-\frac{cs}{8\sigma \pi d} \sqrt{\log M}  \log (\log (M))\right),
   \end{eqnarray*}}
   where $C_{\sigma,c,d,s, F}$ is a constant depending only on $\sigma,c,d,s$ and $C_F$.
   }

\second{
   \vspace{10pt}
\noindent \underline{\textbf{ReLU network} }  

   With Corollary \ref{G-Ghat_Gaussian}, for any $M, L \in \NN$, there exists a ReLU network $\widehat{G}$ of the stated size such that
   \begin{eqnarray*}
    \|\widehat{G} - G\|_{L^\infty ([-1, 1]^N)} 
     \leq  131 C_FC_{\sigma,d,s} \cdot 2^{s+\frac{d}{2}} m^{\frac{d}{2}}(M^2 L^2 \log_3(M+2))^{-\frac{s}{(m+1)^d}}e^{\sigma^2\pi^2 dsm^2/2}.
\end{eqnarray*}

Combining the two error terms with $\log_3(M+2) \geq 1$, we get
 \begin{eqnarray*}
    &&   \sup_{f\in \mathcal{K}}\left| F(f) - \widehat{G}(f(\Bar{t}))\right|\\   
           & \leq& 131 C_F (1+C_{\sigma,d,s} \cdot 2^{s+\frac{d}{2}}) \left(e^{\frac{-csm\log \left(\frac{m}{\sqrt{d}}\right)}{\sqrt{d}}} + m^{\frac{d}{2}}(ML)^{-\frac{2s}{(m+1)^d}}e^{\sigma^2\pi^2 dsm^2/2}\right).
   \end{eqnarray*}

Let
\[
A:=A_{M,L} :=\log\!\left(ML\right).
\]
Take
$C_0:=\left(\frac{1}{2^{d-1}\sigma^2\pi^2 d}\right)^{\frac{1}{d+2}}$
and take
\begin{equation}\label{choice_m_relu_Gau}
m=\left\lfloor C_0 A^{\frac{1}{d+2}}\right\rfloor.
\end{equation}
For $A$ sufficiently large, specifically for
\begin{equation}\label{lower_A}
    A>
\left(\frac{2\sqrt d}{C_0}\right)^{d+2},
\end{equation}
we have
\[
m=\left\lfloor C_0 A^{\frac{1}{d+2}}\right\rfloor
\geq
\frac{C_0}{2}A^{\frac{1}{d+2}}
>
\sqrt d.
\]
This is to ensure $\log(m/\sqrt{d})>0$ and thus $\frac{-csm\log \left(\frac{m}{\sqrt{d}}\right)}{\sqrt{d}} <0$.
Also note that $m+1\leq 2C_0A^{\frac{1}{d+2}}$.

Since
$\left(ML\right)^{-\frac{2s}{(m+1)^d}}
=
\exp\left(
-\frac{2sA}{(m+1)^d}
\right)$,
we have
\begin{eqnarray*}
m^{\frac d2}
\left(ML\right)^{-\frac{2s}{(m+1)^d}}
e^{\sigma^2\pi^2 dsm^2/2} 
=
m^{\frac d2}
\exp\left(
-\frac{2sA}{(m+1)^d}
+\sigma^2\pi^2 dsm^2/2
\right).
\end{eqnarray*}
By our choice of $m$ in \eqref{choice_m_relu_Gau} and the lower bound of A in \eqref{lower_A}, we have 
\begin{eqnarray*}
    \frac{A}{(m+1)^d}
\geq
\frac{1}{(2C_0)^d}
A^{\frac{2}{d+2}} \qquad \hbox{and} \qquad m^2
\leq
C_0^2 A^{\frac{2}{d+2}}.
\end{eqnarray*}
Therefore,
\begin{eqnarray*}
-\frac{2sA}{(m+1)^d}
+\sigma^2\pi^2 dsm^2/2
&\leq&
-s\left[
\frac{2}{(2C_0)^d}
-\frac{\sigma^2\pi^2 d C_0^2}{2}
\right]
A^{\frac{2}{d+2}}.
\end{eqnarray*}
Now, by the definition of $C_0$, we know 
$\sigma^2\pi^2 d C_0^2
=
\frac{1}{2^{d-1}C_0^d}$.
Hence
\[
\frac{2}{(2C_0)^d}
-\frac{\sigma^2\pi^2 d C_0^2}{2}
=
\frac{2}{2^d C_0^d}
-
\frac{1}{2^{d}C_0^d}
=
\frac{1}{2^{d}C_0^d}.
\]
Thus,
\[
-\frac{2sA}{(m+1)^d}
+\sigma^2\pi^2 dsm^2/2
\leq
-\frac{s}{2^d C_0^d}
A^{\frac{2}{d+2}}.
\]
Also, since
$m=\left\lfloor C_0 A^{\frac{1}{d+2}}\right\rfloor
\leq C_0 A^{\frac{1}{d+2}}$,
we have
$m^{\frac d2}
\leq
C_0^{\frac d2} A^{\frac{d}{2(d+2)}}$.
Absorbing this polynomial factor into the exponential, we get
\[
m^{\frac d2}
\left(ML\right)^{-\frac{2s}{(m+1)^d}}
e^{\sigma^2\pi^2 dsm^2/2}
\leq
C_1
\exp\left(
-c_1 A^{\frac{2}{d+2}}
\right),
\]
where $C_1,c_1>0$ depend only on $\sigma,d,s$.

Next, we bound the term $e^{\frac{-csm\log \left(\frac{m}{\sqrt{d}}\right)}{\sqrt{d}}}$. Since
$m\geq \frac{C_0}{2}A^{\frac{1}{d+2}}$,
we have
\begin{eqnarray*}
e^{\frac{-csm\log \left(\frac{m}{\sqrt{d}}\right)}{\sqrt{d}}}
&\leq&
\exp\left(
-\frac{csC_0}{2\sqrt d}
A^{\frac{1}{d+2}}
\log\left(
\frac{C_0}{2\sqrt d}
A^{\frac{1}{d+2}}
\right)
\right).
\end{eqnarray*}
For $A$ sufficiently large, specifically for
\begin{equation}\
    A>
\left(\frac{2\sqrt d}{C_0}\right)^{2(d+2)},
\end{equation}
we have
\[
\log\left(
\frac{C_0}{2\sqrt d}
A^{\frac{1}{d+2}}
\right)
=
\frac{1}{d+2}\log A
-
\log\left(\frac{2\sqrt d}{C_0}\right)
\geq
\frac{1}{2(d+2)}\log A.
\]
Notice that the condition  A$>
\left(\frac{2\sqrt d}{C_0}\right)^{2(d+2)}$ automatically implies the previous lower bound of A we required in \eqref{lower_A}.
It follows that 
\[
e^{-\frac{csm\log(m/\sqrt d)}{\sqrt d}}
\leq
\exp\left(
-\frac{csC_0}{4(d+2)\sqrt d}
A^{\frac{1}{d+2}}
\log A
\right).
\]

Combining the two bounds, we obtain
{\allowdisplaybreaks
\begin{eqnarray*}
&&
\sup_{f\in\mathcal K}
\left|F(f)-\widehat G(f(\bar t))\right|\\
&\leq&
131 C_F\left(1+C_{\sigma,d,s}2^{s+\frac d2}\right)
\left[
e^{-\frac{csm\log(m/\sqrt d)}{\sqrt d}}
+
m^{\frac d2}
\left(ML\right)^{-\frac{2s}{(m+1)^d}}
e^{\sigma^2\pi^2 dsm^2/2}
\right]\\
&\leq&
C_2
\left[
\exp\left(
-\frac{csC_0}{4(d+2)\sqrt d}
A^{\frac{1}{d+2}}\log A
\right)
+
\exp\left(
-c_1 A^{\frac{2}{d+2}}
\right)
\right],
\end{eqnarray*}}
where $C_2>0$ depends only on the fixed parameters.

Since
\[
A^{\frac{2}{d+2}}
\gg
A^{\frac{1}{d+2}}\log A
\qquad
\text{as } A\to\infty,
\]
the second exponential term decays faster than the first one. Hence, there exists
$C_{\sigma,c,d,s,F}>0$ such that
\[
\sup_{f\in\mathcal K}
\left|F(f)-\widehat G(f(\bar t))\right|
\leq
C_{\sigma,c,d,s,F}
\exp\left(
-\frac{csC_0}{4(d+2)\sqrt d}
A^{\frac{1}{d+2}}\log A
\right).
\]
Recall that
$A=\log\!\left(ML\right)$,
we finally obtain
\begin{eqnarray*}
&&\sup_{f\in\mathcal K}
\left|F(f)-\widehat G(f(\bar t))\right|\\
&\leq&
C_{\sigma,c,d,s,F}
\exp\left(
-\frac{csC_0}{4(d+2)\sqrt d}
\left(\log\!\left(ML\right)\right)^{\frac{1}{d+2}}
\log\log\!\left(ML\right)
\right).
\end{eqnarray*}}

The proof is complete. \hfill\BlackBox

Next, we present the proof of Theorem \ref{thm:aj}.


\noindent
{\bf Proof of Theorem \ref{thm:aj}.}

We consider the error decomposition in (\ref{decomposition}). 
Note that $0<a<1$. Also, recall that the kernel $K$ given in \eqref{kernel_torus} with $\theta_j = a^{\|j\|}$ is $\alpha$-H\"{o}lder continuous for $\alpha = 1$ with constant $C_K =  \sum_{j\in\ZZ^p} a^{\|j\|}|j|$, as previously proven in Remark \ref{remark:holder_item1}.
With the choice of $\Bar{t}$ in Theorem \ref{thm:aj}, it follows from Lemma \ref{power_torus} that
the corresponding power function is 
\begin{equation}\label{power_coef1}
    \epsilon_K(\Bar{t}) \leq \sqrt{2 C_K} \left( \frac{\sqrt{p}\pi}{m}\right)^\frac{\alpha}{2} \leq \left(2 \sum_{j\in\ZZ^p} a^{\|j\|}|j|\right)^{\frac{1}{2}} \left( \frac{\sqrt{p}\pi}{m}\right)^{\frac{1}{2}} ,
\end{equation} and thus
\begin{eqnarray*}
    |F(f) - F(Pf)| \leq C_F \|f-Pf\|^s_\infty  &\leq& C_F \|f\|^s_{\mathcal{H}_K}  (\epsilon_K(\Bar{t}))^s \\
    &\leq& C_F\left(2  \sum_{j\in\ZZ^p} a^{\|j\|}|j|\right)^{\frac{s}{2}} \left( \frac{\sqrt{p}\pi}{m}\right)^{\frac{s}{2}}, \qquad \forall f\in  \mathcal{H}_K.
\end{eqnarray*}
It follows that
$$\sup_{f\in \mathcal{K}}| F(f) - F(Pf)| \leq C_F\left(2  \sum_{j\in\ZZ^p} a^{\|j\|}|j|\right)^\frac{s}{2} \left( \frac{\sqrt{p}\pi}{m}\right)^\frac{s}{2} .$$

\vspace{10pt}
\noindent \underline{\textbf{Tanh network} }  

From Corollary \ref{G-Ghat_coef1}, we know for every $M\in \NN$ with $M>5N^2 = 5m^{2p}$,
\begin{eqnarray*}
   \sup_{f\in \mathcal{K}}\left|G(f(\Bar{t})) - \widehat{G}(f(\Bar{t}))\right| \leq  \left \|\widehat{G}-G\right\|_{L^\infty ([-1,1]^{m^p})} 
   \leq  \frac{15C_F C_{a,p,s}}{M^s} m^{\frac{p}{2}(5-s) -s }a^{-\frac{m\sqrt{p}s}{2}}. 
\end{eqnarray*}
Combining the above two error bounds, we get
 \begin{eqnarray*}
       \sup_{f\in \mathcal{K}}\left| F(f) - \widehat{G}(f(\Bar{t}))\right|
            \leq 15C_F C_{a,p,s} \left(\left(2  \sum_{j\in\ZZ^p} a^{\|j\|}|j|\right)^\frac{s}{2} \left(\frac{\sqrt{p}\pi}{m}\right)^\frac{s}{2} + \frac{1}{M^s}m^{\frac{p}{2}(5-s) -s }a^{-\frac{m\sqrt{p}s}{2}}\right).
   \end{eqnarray*}
   We take \begin{equation}\label{choice_of_m}
       m=\left \lfloor \frac{2}{\sqrt{p}s \log(1/a
       )}\log\left(\frac{M^s}{(\log(M))^{\frac{p}{2}(5-s)
       }}\right)\right \rfloor \in\NN.
   \end{equation}
   Then, we see that 
   \begin{eqnarray*}
       m^{-s} &\leq& 2^s
       \left(\frac{2}{\sqrt{p}s \log(1/a
       )}\log\left(\frac{M^s}{(\log(M))^{\frac{p}{2}(5-s)
       }}\right)\right)^{-s}\\
       &\leq& \left(\sqrt{p}s \log\left(\frac{1}{a}
       \right)\right)^s \left(\frac{s}{2}\log(M)\right)^{-s} \\
       &\leq&  \left(2\sqrt{p}\log\left(\frac{1}{a}
       \right)\right)^s (\log
       (M))^{-s},
   \end{eqnarray*}
   if $ \frac{s}{2}\log(M) > \log \left((\log(M))
       ^{\frac{p}{2}(5-s)
       }\right)$. Hence, $$\left(\frac{\sqrt{p}\pi}{m}\right)^{\frac{s}{2}} \leq p^{\frac{s}{4}}\pi^{\frac{s}{2}} \left(2\sqrt{p}\log\left(\frac{1}{a}
       \right)\right)^{\frac{s}{2}}  (\log(M))^{-\frac{s}{2}}.$$ 
       Also, from the choice of $m$ given in \eqref{choice_of_m}, we find
       \begin{eqnarray*}
           \frac{m\sqrt{p}s}{2} \leq \log_{1/a}\left(\frac{M^s}{(\log(M))^{\frac{p}{2}(5-s)
       }}\right).
       \end{eqnarray*}
       Thus, we have
       \begin{eqnarray*}
           && \frac{1}{M^s}m^{\frac{p}{2}(5-s) -s }a^{-\frac{m\sqrt{p}s}{2}} \\
           &\leq&  \frac{1}{M^s} \left(\frac{2}{\sqrt{p}s \log(1/a
       )}\log\left(\frac{M^s}{(\log(M))^{\frac{p}{2}(5-s)
       }}\right)\right )^{\frac{p}{2}(5-s) -s }\frac{M^s}{(\log(M))
       ^{\frac{p}{2}(5-s)
       }}\\
       &\leq& \left(\frac{2}{\sqrt{p}s \log(1/a
       )}\right)^{\frac{p}{2}(5-s) -s } (s\log(M))^{\frac{p}{2}(5-s) -s }\frac{1}{(\log(M))
       ^{\frac{p}{2}(5-s)
       }} \\
       &\leq&  \left(\frac{2}{\sqrt{p}\log(1/a
       ) }\right)^{\frac{p}{2}(5-s) -s }(\log(M))^{-s}.
       \end{eqnarray*}
Then, we have 
{\allowdisplaybreaks
\begin{eqnarray*}
      && \sup_{f\in \mathcal{K}}\left| F(f) - \widehat{G}(f(\Bar{t}))\right|\\
            &\leq& 15C_F C_{a,p,s} \left(\left(2  \sum_{j\in\ZZ^p} a^{\|j\|}|j|\right)^\frac{s}{2} \left(\frac{\sqrt{p}\pi}{m}\right)^\frac{s}{2}  + \frac{1}{M^s}m^{\frac{p}{2}(5-s) -s }a^{-\frac{m\sqrt{p}s}{2}}\right)\\
             &\leq& 15C_F C_{a,p,s} \left(2  \sum_{j\in\ZZ^p} a^{\|j\|}|j|\right)^\frac{s}{2} \\
             && \left(p^{\frac{s}{4}}\pi^{\frac{s}{2}} \left(2\sqrt{p}\log\left(\frac{1}{a}
       \right)\right)^{\frac{s}{2}}  (\log(M))^{-\frac{s}{2}} +  \left(\frac{2}{\sqrt{p}\log(1/a
       ) }\right)^{\frac{p}{2}(5-s) -s }(\log(M))^{-s}\right)\\
             &\leq& C_{a,p,s,F} \ (\log(M))^{-\frac{s}{2}},
   \end{eqnarray*}}
where $ C_{a,p,s,F}$ is a positive constant depending on $a,p,s, C_F$. 

\vspace{10pt}

\second{\noindent \underline{\textbf{ReLU network} }

From Corollary \ref{G-Ghat_coef1}, we know for any $M, L \in \NN$, there exists a ReLU network $\widehat{G}: [-1, 1]^{m^p} \to {\mathbb R}$ such that
\begin{eqnarray*}
    \sup_{f\in \mathcal{K}}\left|G(f(\Bar{t})) - \widehat{G}(f(\Bar{t}))\right| &\leq&  \|\widehat{G} - G\|_{L^\infty ([-1, 1]^{m^p})}  \\
    &\leq& 131 C_FC_{a,p,s} \cdot 2^s \ a^{-\frac{m\sqrt{p}s}{2}}  \ m^{\frac{p}{2}(1-s)-s}   (M^2 L^2 \log_3(M+2))^{-\frac{s}{m^p}}.
\end{eqnarray*}
Combining the two terms, we get
 \begin{eqnarray*}
      && \sup_{f\in \mathcal{K}}\left| F(f) - \widehat{G}(f(\Bar{t}))\right| \\
            &\leq& 131C_F C_{a,p,s} 2^s\left(\left(\sum_{j\in\ZZ^p} a^{\|j\|}|j|\right)^\frac{s}{2} \left(\frac{\sqrt{p}\pi}{m}\right)^\frac{s}{2} + a^{-\frac{m\sqrt{p}s}{2}}  m^{\frac{p}{2}(1-s)-s}   (M^2 L^2 \log_3(M+2))^{-\frac{s}{m^p}}\right).
   \end{eqnarray*}

   Let
\[
A:=A_{M,L}:=\log\!\left(ML\right).
\]
Since $0<a<1$, we have $\log(1/a)>0$. We take 
$C_0:=\left(\frac{1}{\sqrt p\log(1/a)}\right)^{\frac{1}{p+1}}$ and take
\begin{equation}\label{choice_of_m_relu_coef1}
m=\left\lfloor C_0 A^{\frac{1}{p+1}}\right\rfloor .
\end{equation}
For $A$ sufficiently large, specifically for
\[
A\geq \left(\frac{2}{C_0}\right)^{p+1}
=2^{p+1}\sqrt p\log(1/a),
\]
we have
$C_0A^{\frac{1}{p+1}}\geq 2$ and thus
$m\geq 1$.

Since $m\geq \frac{C_0}{2}A^{\frac{1}{p+1}}$, we get  \begin{eqnarray*}
\left(\frac{\sqrt p\pi}{m}\right)^{\frac{s}{2}}
&\leq&
\left(\sqrt p\pi\right)^{\frac{s}{2}}
\left(\frac{2}{C_0}\right)^{\frac{s}{2}}
A^{-\frac{s}{2(p+1)}}.
\end{eqnarray*} 
It follows that
\[
\left(\sum_{j\in\mathbb Z^p}a^{\|j\|}|j|\right)^{\frac{s}{2}}
\left(\frac{\sqrt p\pi}{m}\right)^{\frac{s}{2}}
\leq
C_{a,p,s}
A^{-\frac{s}{2(p+1)}}
\]
with some positive constant $C_{a,p,s}$ depending only on $a,p,$ and $s$.

Next, we write
\begin{eqnarray*}
a^{-\frac{m\sqrt p s}{2}}
\left(M^2L^2\log_3(M+2)\right)^{-\frac{s}{m^p}}
\leq 
\exp\left(
\frac{m\sqrt p s}{2}\log\frac1a
-
\frac{sA}{m^p}
\right).
\end{eqnarray*}
By our choice of $m$ in \eqref{choice_of_m_relu_coef1}, we know 
$m\leq C_0A^{\frac{1}{p+1}}$,
so
\begin{eqnarray*}
    \frac{m\sqrt p s}{2}\log\frac1a
\leq
\frac{s\sqrt p \log(1/a)}{2}C_0
A^{\frac{1}{p+1}} \qquad \hbox{and} \qquad \frac{A}{m^p}
\geq
\frac{1}{C_0^p}
A^{\frac{1}{p+1}}.
\end{eqnarray*}
Since $C_0^{p+1}= \frac{1}{\sqrt p\log(1/a)}$, we have 
$\frac{s\sqrt p\log(1/a)}{2}C_0
=
\frac{s}{2C_0^p}$.
It follows that
\begin{eqnarray*}
\frac{m\sqrt p s}{2}\log\frac1a
-
\frac{sA}{m^p}
\leq
\frac{s}{2C_0^p}A^{\frac{1}{p+1}}
-
\frac{s}{C_0^p}A^{\frac{1}{p+1}}
=
-\frac{s}{2C_0^p}
A^{\frac{1}{p+1}}.
\end{eqnarray*}
Therefore,
\[
a^{-\frac{m\sqrt p s}{2}}
\left(M^2L^2\log_3(M+2)\right)^{-\frac{s}{m^p}}
\leq
\exp\left(
-\frac{s}{2C_0^p}
A^{\frac{1}{p+1}}
\right).
\]

Now consider the term $m^{\frac p2\left(1-s\right)-s}$. Let  
$\gamma:=\frac p2\left(1-s\right)-s$.
If $\gamma\leq 0$, then $m^\gamma\leq 1$ for $m\geq 1$. If $\gamma>0$, then
\[
m^\gamma
\leq
C_0^\gamma A^{\frac{\gamma}{p+1}}.
\]
In either case, there exists a constant $C_{a,p,s}>0$ such that
\[
2^s m^\gamma
\exp\left(
-\frac{s}{2C_0^p}
A^{\frac{1}{p+1}}
\right)
\leq
C_{a,p,s}
\exp\left(
-\frac{s}{4C_0^p}
A^{\frac{1}{p+1}}
\right),
\]
and since exponential decay dominates polynomial decay, we further have
\[
\exp\left(
-\frac{s}{4C_0^p}
A^{\frac{1}{p+1}}
\right)
\leq
C_{a,p,s}
A^{-\frac{s}{2(p+1)}}.
\]
Hence
\[
2^s a^{-\frac{m\sqrt p s}{2}}
m^{\frac p2(1-\frac s2)-s}
\left(M^2L^2\log_3(M+2)\right)^{-\frac{s}{m^p}}
\leq
C_{a,p,s}
A^{-\frac{s}{2(p+1)}}.
\]

Finally, we obtain
\begin{eqnarray*}
\sup_{f\in \mathcal K}
\left|F(f)-\widehat G(f(\bar t))\right|
\leq
C_{a,p,s,\pi,F}
A^{-\frac{s}{2(p+1)}} = C_{a,p,s,F}
\left(
\log\left(ML\right)
\right)^{-\frac{s}{2(p+1)}}.
    \end{eqnarray*}}
The proof is complete. \hfill\BlackBox

Next, we present the proof of Theorem \ref{thm:1+j}.

\noindent
{\bf Proof of Theorem \ref{thm:1+j}}.

We consider the error decomposition in (\ref{decomposition}). Recall that the kernel $K$  given in \eqref{kernel_torus} with $\theta_j = \left(1+\|j\|^2\right)^{-\gamma}$ for $\gamma>(p+1)/2$ is $\alpha$-H\"{o}lder continuous for $\alpha = 1$ with constant  $C_K =  \sum_{j\in\ZZ^p} \left(1+\|j\|^2\right)^{-\gamma}|j|$, as previously proven in Remark \ref{remark:holder_item2}.
With the choice of $\Bar{t}$ in Theorem \ref{thm:1+j}, it follows from Lemma \ref{power_torus} that
the corresponding power function is 
\begin{equation}\label{power_coef2}
    \epsilon_K(\Bar{t}) \leq \sqrt{2 C_K} \left( \frac{\sqrt{p}\pi}{m}\right)^\frac{\alpha}{2} \leq \left(2 \sum_{j\in\ZZ^p} \left(1+\|j\|^2\right)^{-\gamma} |j|\right)^{\frac{1}{2}} \left( \frac{\sqrt{p}\pi}{m}\right)^{\frac{1}{2}},
\end{equation} and thus
$$|F(f) - F(Pf)| \leq C_F\left(2  \sum_{j\in\ZZ^p} \left(1+\|j\|^2\right)^{-\gamma}|j|\right)^{\frac{s}{2}} \left( \frac{\sqrt{p}\pi}{m}\right)^{\frac{s}{2}}, \qquad \forall f\in  \mathcal{H}_K.$$
It follows that
$$\sup_{f\in \mathcal{K}}| F(f) - F(Pf)| \leq  C_F\left(2  \sum_{j\in\ZZ^p} \left(1+\|j\|^2\right)^{-\gamma}|j|\right)^{\frac{s}{2}}  \left( \frac{\sqrt{p}\pi}{m}\right)^{\frac{s}{2}}.$$

\noindent \underline{\textbf{Tanh network} }  

From Corollary \ref{G-Ghat_coef2}, we know for every $M\in \NN$ with $M>5N^2 = 5m^{2p}$, there exists a tanh network $\widehat{G}$ such that
\begin{eqnarray*}
   \sup_{f\in \mathcal{K}}\left|G(f(\Bar{t})) - \widehat{G}(f(\Bar{t}))\right| \leq  \left \|\widehat{G}-G\right\|_{L^\infty [-1,1]^{m^p}} 
   \leq   \frac{15C_F C_{\gamma,p,s}}{M^s} m^{\frac{5p}{2}+s(2\gamma-\frac{p}{2}-1)}.
\end{eqnarray*}
Combining the two error bounds, we have 
\begin{eqnarray*}
       \sup_{f\in \mathcal{K}}\left| F(f) - \widehat{G}(f(\Bar{t}))\right|
            \leq C_F  \left(\left(2  \sum_{j\in\ZZ^p} \left(1+\|j\|^2\right)^{-\gamma}|j|\right)^{\frac{s}{2}}\left(\frac{\sqrt{p}\pi}{m}\right)^{\frac{s}{2}} + 15 C_{\gamma,p,s}\frac{1}{M^s}m^{\frac{5p}{2}+s(2\gamma-\frac{p}{2}-1)}\right).
   \end{eqnarray*}
We can see that $m^{-\frac{s}{2}}$ is a decreasing function of $m$, whereas $m^{\frac{5p}{2}+s(2\gamma-\frac{p}{2}-1)}$ is an increasing function of $m$ (note that $\gamma>(p+1)/2$). We choose the smallest integer $m$ such that $m^{-\frac{s}{2}} \leq \frac{1}{M^s}m^{\frac{5p}{2}+s(2\gamma-\frac{p}{2}-1)}$; that is
\begin{eqnarray*}
   &&m^{-\frac{s}{2}-\frac{5p}{2}-s(2\gamma-\frac{p}{2}-1)} \leq \frac{1}{M^s}\\
   &\Leftrightarrow& m^{\frac{5p}{2}+s(2\gamma-\frac{p}{2} -\frac{1}{2} )} \geq M^s\\
   &\Leftrightarrow& m \geq M^{\frac{s}{\frac{5p}{2}+s(2\gamma-\frac{p}{2}-\frac{1}{2})}}.
\end{eqnarray*}
Then we take \begin{equation}
    m = \left\lceil M^{\frac{s}{\frac{5p}{2}+s(2\gamma-\frac{p}{2}-\frac{1}{2})}}\right\rceil \in \NN.
\end{equation}
The requirement $M>5m^{2p}$ is satisfied when $M> 6 M^{\frac{2ps}{\frac{5p}{2}+s(2\gamma-\frac{p}{2}-\frac{1}{2})}}$, that is, $M > M_0 := 6^{\frac{5p+s(4\gamma -p-1
)}{5p(1-s)+s(4\gamma -1)}}$. Then we know
\begin{eqnarray*}
    \frac{1}{M^s}m^{\frac{5p}{2}+s(2\gamma-\frac{p}{2}-1)} 
    &\leq& 2^{\frac{5p}{2} + 2s\gamma}\ M^{\frac{s\left(\frac{5p}{2}+s(2\gamma-\frac{p}{2}-1)\right)}{\frac{5p}{2}+s(2\gamma-\frac{p}{2} -\frac{1}{2})} -s} \\
    &=& 2^{\frac{5p}{2} + 2s\gamma}\ M^{-\frac{\frac{s^2}{2}}{\frac{5p}{2}+s(2\gamma-\frac{p}{2} - \frac{1}{2})}}\\
    &=&2^{\frac{5p}{2} + 2s\gamma}\ M^{-\frac{s^2}{5p+s(4\gamma-p - 1)}}.
\end{eqnarray*}

\second{
\noindent \underline{\textbf{ReLU network} }  

From Corollary \ref{G-Ghat_coef2}, we know for any $M, L \in \NN$, there exists a ReLU network of the stated size $\widehat{G}: [-1, 1]^{m^p} \to {\mathbb R}$ such that
\begin{eqnarray*}
\sup_{f\in \mathcal{K}}\left|G(f(\Bar{t})) - \widehat{G}(f(\Bar{t}))\right| &\leq&  \left \|\widehat{G}-G\right\|_{L^\infty [-1,1]^{m^p}} \\
    &\leq& 131 C_F C_{\gamma,p,s} \cdot 2^s \  m^{\frac{p}{2}+s(2\gamma-\frac{p}{2}-1)}  (M^2 L^2 \log_3(M+2))^{-\frac{s}{m^p}}.
\end{eqnarray*}

Then, we obtain 
\begin{eqnarray*}
       &&\sup_{f\in \mathcal{K}}\left| F(f) - \widehat{G}(f(\Bar{t}))\right|\\
            &\leq& 131 C_F C_{\gamma,p,s} 2^s \left(\left(  \sum_{j\in\ZZ^p} \left(1+\|j\|^2\right)^{-\gamma}|j|\right)^{\frac{s}{2}}\left(\frac{\sqrt{p}\pi}{m}\right)^{\frac{s}{2}} + m^{\frac{p}{2}+s(2\gamma-\frac{p}{2}-1)}  (M^2 L^2 \log_3(M+2))^{-\frac{s}{m^p}}\right).
   \end{eqnarray*}
   
 Let
\[
A:=A_{M,L}:=\log\left(ML\right).
\]
For notational simplicity, we define 
$\beta:=\frac p2+s\left(2\gamma-\frac p2-1\right)$.
Then the ReLU approximation bound can be written as
\[
\sup_{f\in\mathcal K}
\left|F(f)-\widehat G(f(\bar t))\right|
\leq
C_{F,\gamma,p,s}
\left(
m^{-\frac s2}
+
m^\beta e^{-\frac{2sA}{m^p}}
\right),
\]
where we have absorbed fixed constants, including $2^s$, into the constant
$C_{F,\gamma,p,s}$.

Since $\gamma>(p+1)/2$, we know $\beta + \frac{s}{2}= 
\frac p2+s\left(2\gamma-\frac p2-\frac12\right)>0$. Take \[
C_0:=\left(\frac{sp}{\beta + \frac{s}{2}}\right)^{1/p}
\]
and 
\[
m=\left\lfloor
C_0\left(\frac{A}{\log A}\right)^{1/p}
\right\rfloor.
\]
Choose $A_0> e$ sufficiently large such that $\frac{A_0}{\log A_0}\geq \left(\frac{2}{C_0}\right)^p$. Then for all $A> A_0$, since $A / \log A$ is increasing for $A>e$, 
$\frac{A}{\log A}\geq \left(\frac{2}{C_0}\right)^p$ and thus
$C_0\left(\frac{A}{\log A}\right)^{1/p}\geq 2$.
Using the fact that $\lfloor x\rfloor\geq x/2$ for every $x\geq 2$, we obtain
\[
m
=
\left\lfloor C_0\left(\frac{A}{\log A}\right)^{1/p}\right\rfloor
\geq
\frac{C_0}{2}\left(\frac{A}{\log A}\right)^{1/p}.
\]

We first bound the term $m^{-\frac s2}$. By the lower bound on $m$, we know
\[
m^{-\frac s2}
\leq
\left(\frac{2}{C_0}\right)^{\frac s2}
\left(\frac{A}{\log A}\right)^{-\frac{s}{2p}}
=
\left(\frac{2}{C_0}\right)^{\frac s2}
\left(\frac{\log A}{A}\right)^{\frac{s}{2p}}.
\]

Next, we bound the term $m^\beta e^{-\frac{sA}{m^p}}$. Since
$m^p
\leq
C_0^p\frac{A}{\log A}$,
we have
\[
e^{-\frac{2sA}{m^p}}
\leq
e^{-\frac{2s}{C_0^p}\log A}
=
A^{-\frac{2s}{C_0^p}}.
\]
Also,
\[
m^\beta
\leq
C_0^\beta
\left(\frac{A}{\log A}\right)^{\frac{\beta}{p}}
\leq
C_0^\beta A^{\frac{\beta}{p}}.
\]
Thus,
\[
m^\beta e^{-\frac{sA}{m^p}}
\leq
C_0^\beta
A^{\frac{\beta}{p}-\frac{2s}{C_0^p}}.
\]
By the definition of $C_0$, $\frac{\beta}{p}-\frac{2s}{C_0^p}
= -\frac{\beta+s}{p}.
$
Since $\beta+s>s/2$, it follows that
\[
m^\beta e^{-\frac{sA}{m^p}}
\leq
C_{\gamma,p,s}A^{-\frac{\beta+s}{p}}
\leq
C_{\gamma,p,s}A^{-\frac{s}{2p}}.
\]

Combining the two estimates, we obtain with a constant $C_{\gamma,p,s,F}$ independent of $M,L$ such that
\[
\sup_{f\in\mathcal K}
\left|F(f)-\widehat G(f(\bar t))\right|
\leq
C_{\gamma,p,s,F}
\left(\frac{\log A}{A}\right)^{\frac{s}{2p}}.
\]
Recall
$A=\log\left(ML\right)$,
we obtain 
\[
\sup_{f\in\mathcal K}
\left|F(f)-\widehat G(f(\bar t))\right|
\leq
C_{\gamma,p,s,F}
\left(
\frac{
\log\log\left(ML\right)
}{
\log\left(ML\right)
}
\right)^{\frac{s}{2p}}.
\]}
   
The proof is complete. \hfill\BlackBox

Next, we give the proof of Theorem \ref{thm:general}.

\noindent
{\bf Proof of Theorem \ref{thm:general}}.

    The proof is rather straightforward. Here, we consider  $
    \mathcal{K} := \{f\in \mathcal{H}_K: \|f\|_{\mathcal{H}_K} \leq 1\}
$ with $\mathcal{H}_K$ induced by some Mercer kernel $K$ which is $\alpha$-H\"{o}lder continuous for $\alpha \in (0, 1]$ with constant $C_K \geq 0$. Note that we do not require $K$ to be translation-invariant. Again, we consider the error decomposition 
        \begin{eqnarray*}
         \sup_{f\in \mathcal{K}}\left| F(f) - \widehat{G}(f(\Bar{t}))\right| &\leq  \sup_{f\in \mathcal{K}}| F(f) - F(Pf)| +  \sup_{f\in \mathcal{K}}|F(Pf) - \widehat{G}(f(\Bar{t}))| \\
        &=  \sup_{f\in \mathcal{K}}| F(f) - F(Pf)| +  \sup_{f\in \mathcal{K}}|G(f(\Bar{t})) - \widehat{G}(f(\Bar{t}))|.
    \end{eqnarray*}
    The upper bound of the first term is derived earlier in (\ref{eq:firstterm}), which is a consequence of Lemma \ref{f-Pf}. 
    The theorem follows directly after Lemma \ref{lemma:reg_G} and  Theorem \ref{thm:extension_tim}.
    \hfill\BlackBox

Next, we give the proof of Corollary \ref{corollary:FLM}. 

\noindent
{\bf Proof of Corollary \ref{corollary:FLM}}.
 We first prove the functional $F|_{\mathcal{H}_{K_0}}$ is H\"{o}lder continuous.
 By the Cauchy-Schwarz inequality,
$$|\langle X, \beta \rangle_{L^2([0,1])}| \leq \|\beta\|_{L^2}  \|X\|_{L^2}\leq \|\beta\|_{L^2}\kappa  \|X\|_{\mathcal{H}_{K_0}} \leq \kappa \|\beta\|_{L^2}.$$
For $X, U \in \mathcal{K}_0$, we have 
 \begin{eqnarray*}
    \left |F|_{\mathcal{H}_{K_0}} (X) - F|_{\mathcal{H}_{K_0}} (U)\right| 
     &=& \left|g\left(\langle X, \beta \rangle_{L^2([0,1])}\right) - g\left(\langle U, \beta \rangle_{L^2([0,1])}\right)\right|\\
     &\leq&\|g\|_{Lip([-\kappa \|\beta\|_{L^2}, \kappa \|\beta\|_{L^2}])}\|\beta\|_{L^2} \|X-U\|_{L^2}\\
     &\leq& \|g\|_{Lip([-\kappa \|\beta\|_{L^2}, \kappa \|\beta\|_{L^2}])} \|\beta\|_{L^2}\|X-U\|_{\infty}.
 \end{eqnarray*}
 We see that $F|_{\mathcal{H}_{K_0}}$ is $s$-H\"{o}lder continuous with $s=1$ and constant $$C_F = \|g\|_{Lip([-\kappa \|\beta\|_{L^2}, \kappa \|\beta\|_{L^2}])} \|\beta\|_{L^2}.$$
 Substitute $d=1$, $s=1$, and $C_F = \|g\|_{Lip([-\kappa \|\beta\|_{L^2}, \kappa \|\beta\|_{L^2}])} \|\beta\|_{L^2}$ into Theorem \ref{thm:general} and Theorem \ref{thm:Gaussian}, the proof is complete. 
 \hfill\BlackBox

Lastly, we present the proof of Theorem \ref{thm:extension_tim}.

\noindent
{\bf Proof of Theorem \ref{thm:extension_tim}}.

First, by the classical Whitney extension theorem, the $s$-H\"{o}lder continuous function $G$ on $[-1, 1]^N$ can be extended to a $s$-H\"{o}lder continuous function $g$ on $[-2, 2]^N$ with the same H\"{o}lder constant $C_G$. In fact, we can define $g$ explicitly as $g(x_1, \cdots, x_N) = G(\widetilde x_1, \cdots, \widetilde x_N)$ where 
\begin{equation*}
    \widetilde x_i = \begin{cases}
      x_i, & \text{if $x_i \in [-1,1]$},\\
      2-x_i, & \text{if $x_i \in (1,2]$},\\
      -2-x_i, & \text{if $x_i \in [-2,-1)$}.
    \end{cases} 
\end{equation*}

Then we apply a smoothing technique to construct a function $h_\tau$ on $[-1, 1]^N$ with $\tau=\frac{1}{M} \in (0, 1]$ and $M\in {\mathbb N}$ as 
\begin{eqnarray*}
    h_\tau (x) &=& \frac{1}{\tau^N} \int_0^\tau \cdots \int_0^\tau g(x+(t_1, \ldots, t_N)) d t_1 \ldots d t_N\\
    &=& \frac{1}{\tau^N} \int_{x_1}^{x_1 +\tau} \cdots \int_{x_N}^{x_N +\tau} g(u) d u_1 \ldots d u_N
\end{eqnarray*}
for $x\in [-1, 1]^N$. 
We can see from $$G(x) - h_\tau (x) = \frac{1}{\tau^N} \int_0^\tau \cdots \int_0^\tau g(x) - g(x+(t_1, \ldots, t_N)) d t_1 \ldots d t_N$$ and the $s$-H\"{o}lder continuity of $g$ that 
$$\|G - h_\tau\|_{L^\infty ([-1, 1]^N)} \leq C_G \sqrt{N} \tau^s.$$

Moreover,  from the partial derivative $$\frac{\partial h_\tau}{\partial x_1} =  \frac{1}{\tau^N} \int_{x_2}^{x_2 +\tau} \cdots \int_{x_N}^{x_N +\tau} g(x_1 + \tau, \hat{u}) - g(x_1, \hat{u}) d u_2 \ldots d u_N$$ with $\hat{u} = (u_2, \ldots, u_N)$ and similar expressions for the other partial derivatives, we find that $\|\frac{\partial h_\tau}{\partial x_i}\|_{L^\infty ([-1, 1]^N)} \leq C_G \tau^{s-1}$ for each $i=1, \ldots, N$, and $h_\tau$ is a Lipschitz continuous function on $[-1, 1]^N$ with Lipschitz constant $C_G\sqrt{N} \tau^{s-1}$. 

We now apply Corollary 5.4 in \citep{de2021approximation}, which asserts that for 
 a Lipschitz continuous function $f$ on $[0,1]^N$ with Lipschitz constant $L>0$ and every $M\in \NN$ with $M>5N^2$, there exists a tanh neural network $\widehat{f}$ with two hidden layers of widths at most $N(M-1)$ and $3\left\lceil \frac{N+1}{2} \right\rceil    
\begin{pmatrix}
        2N-1\\
        N
    \end{pmatrix}M^N$ (or $M-1$ and $6M$ for $N=1$), such that 
    \begin{equation}\label{eqn:tim}
      \left\|\widehat{f}-f\right\|_{L^\infty [0,1]^N} \leq \frac{7N^2L}{M}.   
    \end{equation}
Additionally, it is shown in Lemma 2.1 in \citep{de2021approximation} that $ \begin{pmatrix}
        2N-1\\
        N
    \end{pmatrix}< 5^N$. The total number of parameters in the network is at most $3\lceil\frac{N+1}{2}\rceil(5M)^N$ with $M>5N^2$.     
    
We apply the above approximation result \eqref{eqn:tim}  to the function $f(x) = h_\tau (2x -(1, \ldots, 1))$ on $[0, 1]^N$ with Lipschitz constant $2 C_G
\sqrt{N} \tau^{s-1}$, and find for every $M> 5N^2$ a tanh neural network $\widehat{f}$ with two hidden layers of widths at most $N(M-1)$ and $3 \lceil \frac{N+1}{2}\rceil(5M)^N$ such that
$$ \|\widehat{f} - f\|_{L^\infty ([0, 1]^N)} \leq \frac{14 N^2 C_G \sqrt{N} \tau^{s-1}}{M}.$$
It follows the function $\widehat{G} = \widehat{f} \left(\frac{1}{2} \cdot + (1/2, \ldots, 1/2)\right): [-1, 1]^N \to {\mathbb R}$ produced by a tanh neural network with two hidden layers of the same widths satisfies 
$$\|\widehat{G} - G\|_{L^\infty ([-1, 1]^N)}  = \|\widehat{G} - h_\tau + h_\tau -G\|_{L^\infty ([-1, 1]^N)} \leq  \|\widehat{f} - f\|_{L^\infty ([0, 1]^N)} + \|G - h_\tau\|_{L^\infty ([-1, 1]^N)},$$
which can be bounded with $\tau = \frac{1}{M}$ by 
$$ \frac{14 N^2 C_G \sqrt{N} \tau^{s-1}}{M} + C_G \sqrt{N} \tau^s =\frac{(14 N^2 +1) C_G \sqrt{N}}{M^s}. $$
This proves the theorem. 
\hfill\BlackBox

\vspace{10pt}

\second{
\noindent
{\bf Proof of Theorem \ref{thm:gaussian-learning-rates}}.
The proof for this generalization result is similar to that in \citep{shi2025learning}. 

Recall the oracle inequality \eqref{eq:abstract-oracle-inequality} for a general clipped function class $\pi_B\mathcal H\subseteq\{F:\mathcal K\to[-B,B]\}$:
\begin{eqnarray*}
   \mathbb E_{\mathcal D_n}[\|\widehat F_{\mathcal D_n}-F_\rho\|_{L^2(\rho_{\mathcal K})}^2]\leq \mathcal C_{n}(\pi_B\mathcal H) + 2 \inf_{F \in \pi_B\mathcal H} \left\|
F-F_\rho
\right\|_{L^2(\rho_{\mathcal K})}^2.
\end{eqnarray*}
We will focus on estimating the sample error term $\mathcal C_{n}(\pi_B\mathcal H) =
\mathbb E_{\mathcal D_n}
\left[
\Delta(\widehat F_{\mathcal D_n})-2\Delta_n(\widehat F_{\mathcal D_n})
    \right]$. The approximation error $\inf_{F \in \pi_B\mathcal H} \left\|
F-F_\rho
\right\|_{L^2(\rho_{\mathcal K})}^2$ has been studied in the preceding sections. 

 For a set of $n$ fixed functions in $\pi_B\mathcal H$ denoted by 
$\boldsymbol f=(f_1,\ldots,f_n)\in \pi_B\mathcal H$, let
\[
d_{1,\boldsymbol f}(F,G)
:=
\frac1n\sum_{i=1}^n|F(f_i)-G(f_i)|.
\]
For $\eta>0$, denote by $\mathcal N
\bigl(\eta,\mathcal F,d_{1,\boldsymbol f}\bigr)$ the $\eta$-covering number of  $\mathcal F:= \pi_B\mathcal H$ under the metric $d_{1,\boldsymbol f}$. 
Set
\[
 H_{1,n}(\eta,\mathcal F)
:=
\sup_{\boldsymbol f\in \mathcal F}
\log\mathcal N
\bigl(\eta,\mathcal F,d_{1,\boldsymbol f}\bigr).
\]
We apply a concentration inequality given in \citep[Theorem 11.4] {gyorfi2002distribution} with $\delta = \frac{1}{2}$ and $\alpha = \beta = \frac{\eta}{2}$. This concentration inequality asserts that, for any $0 < \eta < \frac{1}{2}$, 
\begin{eqnarray}\label{gyorfi}
   && \mathbb{P}\left\{\exists F \in  \mathcal F: \left\|
F-F_\rho
\right\|_{L^2(\rho_{\mathcal K})}^2- \Delta_n(F) \geq \frac{1}{2}\left(\eta +  \left\|
F-F_\rho
\right\|_{L^2(\rho_{\mathcal K})}^2\right)\right\} \nonumber\\
&\leq& 14 \exp\left( H_{1,n}\left(\frac{\eta}{80B},\mathcal F\right) - \frac{n\eta}{B^4}\right).
\end{eqnarray}
We note that while this result specifically considers elements of $\mathcal F$ defined on $\RR^d$, it is also applicable to situations where the elements of $\mathcal F$ are defined on an
arbitrary set.

We specifically consider the neural network function class $\mathcal G_{\sigma,\boldsymbol p}$ introduced above in \eqref{neural_function} with ReLU activation 
$\sigma(x) = \max\{0,x\}$.
Let
$D$, $W$ denote its depth,
maximum width, and
let
\begin{equation}
\mathcal P_{\boldsymbol p}
:=
\sum_{\ell=1}^D p_\ell(p_{\ell-1}+1)
\end{equation}
be the total number of network parameters. By the pseudodimension
estimate for ReLU network in \citep{bartlett2019nearly},
\[
\operatorname{Pdim}
(\mathcal G_{\mathrm{ReLU},\boldsymbol p})
=
\mathcal{O}(\mathcal P_{\boldsymbol p}D
\log(\mathcal P_{\boldsymbol p})).
\]

Composition with $S_N$ cannot increase pseudodimension, and clipping is
monotone. We further consider the clipped ReLU network function class $\pi_B\mathcal H_{N,\mathrm{ReLU},\boldsymbol p}$.
 Since the clipped class takes values in $[-B,B]$, the standard
empirical entropy estimate therefore yields
\begin{equation}
\label{eq:relu-empirical-covering}
 H_{1,n}
(\eta,\pi_B\mathcal H_{N,\mathrm{ReLU},\boldsymbol p})
\leq
C_1\mathcal P_{\boldsymbol p}D
\log(\mathcal P_{\boldsymbol p})
\log\left(\frac{C_2B}{\eta}\right),
\qquad 0<\eta\leq B,
\end{equation}
where $C_1,C_2>0$ are universal constants. Combining
\eqref{eq:relu-empirical-covering} with the concentration inequality \eqref{gyorfi} gives
\begin{equation}
\label{eq:relu-sample-error}
\mathcal C_n
(\pi_B\mathcal H_{N,\mathrm{ReLU},\boldsymbol p})
\leq
\frac{CB^2}{n}
\left(
\mathcal P_{\boldsymbol p}D
\log(\mathcal P_{\boldsymbol p})\log(n)+1
\right).
\end{equation}
The approximation term follows because clipping
cannot increase the error when $|F_\rho|\leq B$.

An
empirical risk minimizer over
$\pi_B\mathcal H_{N,\mathrm{ReLU},\boldsymbol p}$ satisfies
\begin{align}
\mathbb E_{\mathcal D_n}
\|\widehat F_{\mathcal D_n}-F_\rho\|_{L^2(\rho_{\mathcal K})}^2
\leq
2\inf_{G\in\mathcal G_{\mathrm{ReLU},\boldsymbol p}}
\sup_{f\in\mathcal K}
|F_\rho(f)-G(S_Nf)|^2
\nonumber
+
\frac{CB^2}{n}
\left(
\mathcal P_{\boldsymbol p}D
\log(\mathcal P_{\boldsymbol p})\log(n)+1
\right),
\label{eq:relu-oracle}
\end{align}
where $C>0$ is a universal constant.

Recall that $m$ and $N$ are chosen as prescribed in
Theorem~\ref{thm:Gaussian}\textup{(ii)}. In particular,
$N=(m+1)^d$, and the choice of $m$ gives
$N=(m+1)^d
\leq
C(\log M)^{\frac{d}{d+2}}$.
Moreover,
\[
D=2N+29
\]
and
\[
W
=
3^{N+3}
\max\left\{
N\left\lfloor M^{1/N}\right\rfloor,M+2
\right\}
\leq
C3^N(N+1)M.
\]
After padding the hidden layers to width $W$, the total number of parameters
satisfies
\[
\mathcal P_{\boldsymbol p}
\leq
C(N+1)^3 9^N M^2.
\]
Since $M=\lceil n^\alpha\rceil$ and
$N=\mathcal O\left(
(\log n)^{\frac{d}{d+2}}
\right)
=o(\log n)$,
we obtain
\[
\log\mathcal P_{\boldsymbol p}
\leq
2\alpha\log n+o(\log n).
\]
Consequently, for any $\delta>0$,
\[
\mathcal P_{\boldsymbol p} D\log(\mathcal P_{\boldsymbol p})\log n
\leq
C_\delta n^{2\alpha+\delta}
\]
for all sufficiently large $n$. Therefore, the sample-error term satisfies
\[
\frac{CB^2}{n}
\left(
\mathcal P_{\boldsymbol p} D\log(\mathcal P_{\boldsymbol p})\log n+1
\right)
\leq
C_\delta B^2 n^{-1+2\alpha+\delta}.
\]
In particular, this term converges to zero whenever
$0<\delta<1-2\alpha$.

On the other hand, Theorem~\ref{thm:Gaussian}\textup{(ii)} and the oracle
inequality give the squared approximation term
\[
2C_{\sigma,c,d,s,F}^2
\exp\left\{
-\frac{csC_0}{2(d+2)\sqrt d}
(\log M)^{\frac{1}{d+2}}\log\log M
\right\}.
\]
Since
\[
\frac{\log M}{\log n}\longrightarrow\alpha,
\qquad
\frac{\log\log M}{\log\log n}\longrightarrow 1,
\]
for any fixed 
\[
0<\kappa<
\frac{csC_0}{2(d+2)\sqrt d}\,
\alpha^{\frac{1}{d+2}},
\]
the approximation term is bounded, for all sufficiently large $n$, by
\[
C\exp\left\{
-\kappa
(\log n)^{\frac{1}{d+2}}\log\log n
\right\}.
\]
Combining the approximation and sample-error estimates, we conclude that
\[
\mathbb E_{\mathcal D_n}
\left\|
\widehat F_{\mathcal D_n}-F_\rho
\right\|_{L^2(\rho_{\mathcal K})}^2
\leq
C\exp\left\{
-\kappa
(\log n)^{\frac{1}{d+2}}\log\log n
\right\}
+
C_\delta B^2n^{-1+2\alpha+\delta}.
\]
Since the second term decays faster than the first,
\[
\mathbb E_{\mathcal D_n}
\left\|
\widehat F_{\mathcal D_n}-F_\rho
\right\|_{L^2(\rho_{\mathcal K})}^2
=
\mathcal O\left(
\exp\left\{
-\kappa
(\log n)^{\frac{1}{d+2}}\log\log n
\right\}
\right).
\]
This proves the theorem.
\hfill \BlackBox
}
 
\second{
\section{Derivation of the Parameter Complexity Bounds}
\label{app:parameter-complexity}

\subsection{Parameter Complexity Bound for the Gaussian Kernel}
\label{app:gaussian-parameter-complexity}

We derive the parameter-complexity bounds for the Gaussian kernel in
Table~\ref{tab:compare} from the approximation estimates in
Theorem~\ref{thm:Gaussian}. Throughout this section, let
$\mathcal{P}$ denote the total number of network parameters and let
$\varepsilon>0$ be the prescribed approximation accuracy. 

We first recall the following elementary asymptotic inversion. Suppose
that $x\log x\geq u$ is required, where $u$ is sufficiently large. It is sufficient to take $x=\frac{2u}{\log u}$, since
\[
\log\left(\frac{2u}{\log u}\right)
=
\log u-\log\log u+\log 2
\geq \frac{1}{2}\log u
\]
for all sufficiently large $u$. The choice $x=\frac{Au}{(\log u)^{1+r}}$ does not satisfy the inequality for any fixed $A,r>0$ and sufficiently large $u$
We use this to invert the approximation bounds below.

\paragraph{Tanh network.}
For notational convenience, write
\[
C_{\tanh}=C_{\sigma,c,s,d,F},
\qquad
a_{\tanh}=\frac{cs}{8\sigma\pi d},
\qquad
q=\log M.
\]
The approximation bound in Theorem~\ref{thm:Gaussian}(i) becomes
\[
\sup_{f\in\mathcal K}
\left|F(f)-\widehat G(f(\bar t))\right|
\leq
C_{\tanh}\exp\left(-a_{\tanh}\sqrt q\,\log q\right).
\]
To make the right-hand side at most $\varepsilon$, it is sufficient
that
\[
a_{\tanh}\sqrt q\,\log q
\geq
R_{\tanh,\varepsilon},
\qquad
R_{\tanh,\varepsilon}
:=
\log\left(\frac{C_{\tanh}}{\varepsilon}\right).
\]
Set $x=\sqrt q$. The above condition is
equivalent to
\[
x\log x
\geq
\frac{R_{\tanh,\varepsilon}}{2a_{\tanh}}.
\]
By the elementary inversion above, it suffices to choose some $q$ of the size
\begin{equation}\label{eq:Gaussian-tanh-logM}
q=\log M
=
\mathcal O\left(
\left[
\frac{R_{\tanh,\varepsilon}}
{\log R_{\tanh,\varepsilon}}
\right]^2
\right).
\end{equation}

We next express the size of the resulting network in terms of $q$.
From the choice of $m$ in Theorem~\ref{thm:Gaussian}(i),
\[
m
=
\left\lfloor
\frac{\sqrt{2}}{\sigma\pi\sqrt d}
\left(\sqrt q-cq^{1/4}\right)
\right\rfloor,
\]
and hence
\[
m+1=\mathcal O(\sqrt q),
\qquad
N=(m+1)^d=\mathcal O\left(q^{d/2}\right).
\]
Let
\[
W_1=N(M-1),
\qquad
W_2=3\left\lceil\frac{N+1}{2}\right\rceil(5M)^N
\]
denote the widths of the two hidden layers. Counting all weights and
biases in a fully connected realization gives
\[
\mathcal P_{\tanh}
\leq
(N+1)W_1+(W_1+1)W_2+(W_2+1).
\]
Therefore,
\begin{align*}
\log\mathcal P_{\tanh}=
\mathcal O\left(
N\log(5M)+\log M+\log N
\right)
=
\mathcal O(Nq)
=
\mathcal O\left(q^{\frac d2+1}\right).
\end{align*}
Substituting \eqref{eq:Gaussian-tanh-logM} yields
\[
\log\mathcal P_{\tanh}
=
\mathcal O\left(
\left[
\frac{R_{\tanh,\varepsilon}}
{\log R_{\tanh,\varepsilon}}
\right]^{d+2}
\right).
\]
Since
$R_{\tanh,\varepsilon}
=
\log\left(\frac{C_{\tanh}}{\varepsilon}\right)
=
\Theta\left(\log\frac{1}{\varepsilon}\right)$
as $\varepsilon\to0$, we obtain
\[
\mathcal P_{\tanh,\varepsilon}
=
\exp\left\{
\mathcal O\left(
\left[
\frac{\log(1/\varepsilon)}
{\log\log(1/\varepsilon)}
\right]^{d+2}
\right)
\right\}.
\]

\paragraph{ReLU network.}
Let
\[
C_{\mathrm{ReLU}}=C_{\sigma,c,d,s,F},
\qquad
a_{\mathrm{ReLU}}
=
\frac{csC_0}{4(d+2)\sqrt d},
\qquad
q=\log(ML).
\]
Theorem~\ref{thm:Gaussian}(ii) gives
\[
\sup_{f\in\mathcal K}
\left|F(f)-\widehat G(f(\bar t))\right|
\leq
C_{\mathrm{ReLU}}
\exp\left(
-a_{\mathrm{ReLU}}
q^{\frac{1}{d+2}}\log q
\right).
\]
To make the right-hand side at most $\varepsilon$, it is sufficient
that
\[
a_{\mathrm{ReLU}}
q^{\frac{1}{d+2}}\log q
\geq
R_{\mathrm{ReLU},\varepsilon},
\qquad
R_{\mathrm{ReLU},\varepsilon}
:=
\log\left(\frac{C_{\mathrm{ReLU}}}{\varepsilon}\right).
\]
Set
$x=q^{\frac{1}{d+2}}$.
The above condition becomes
\[
a_{\mathrm{ReLU}}(d+2)x\log x
\geq R_{\mathrm{ReLU},\varepsilon}.
\]
The elementary inversion above shows that it suffices to take some $x$ of the size 
\[
x
=
\mathcal O\left(
\frac{R_{\mathrm{ReLU},\varepsilon}}
{\log R_{\mathrm{ReLU},\varepsilon}}
\right).
\]
Consequently,
\begin{equation}\label{eq:Gaussian-ReLU-logML}
q=\log(ML)
=
\mathcal O\left(
\left[
\frac{R_{\mathrm{ReLU},\varepsilon}}
{\log R_{\mathrm{ReLU},\varepsilon}}
\right]^{d+2}
\right).
\end{equation}
For example, up to integer rounding, one may choose
$M=L=\exp(q/2)$.

We now bound the total number of parameters. From
$m=\left\lfloor
C_0q^{\frac{1}{d+2}}
\right\rfloor$
we have
\[
N=(m+1)^d
=
\mathcal O\left(q^{\frac{d}{d+2}}\right).
\]
Let
\[
D=11L+2N+18
\]
and
\[
W
=
3^{N+3}
\max\left\{
N\left\lfloor M^{1/N}\right\rfloor,M+2
\right\}
\]
denote the depth and maximal width, respectively. The total number of
parameters of a fully connected network with depth $D$ and maximal
width $W$ is bounded by
\[
\mathcal P_{\mathrm{ReLU}}
= \mathcal{O}((D+1)(W+1)^2).
\]Moreover,
\[
\log D
=
\mathcal O\left(\log L+\log(N+1)\right)
=
\mathcal O(q),
\]
because $\log L\leq\log(ML)=q$. Similarly,
\begin{align*}
\log W
\leq
(N+3)\log 3
+
\mathcal O\left(\log N+\log(M+2)\right)
=
\mathcal O(N+q).
\end{align*}
Because
$N=\mathcal O\left(q^{\frac{d}{d+2}}\right)=o(q)$,
it follows that
\[
\log\mathcal P_{\mathrm{ReLU}}
=
\mathcal O(q).
\]
Substituting \eqref{eq:Gaussian-ReLU-logML} gives
\[
\log\mathcal P_{\mathrm{ReLU}}
=
\mathcal O\left(
\left[
\frac{R_{\mathrm{ReLU},\varepsilon}}
{\log R_{\mathrm{ReLU},\varepsilon}}
\right]^{d+2}
\right).
\]
And since
$R_{\mathrm{ReLU},\varepsilon}
=
\Theta\left(\log\frac{1}{\varepsilon}\right)$, we obtain
\[
\mathcal P_{\mathrm{ReLU},\varepsilon}
=
\exp\left\{
\mathcal O\left(
\left[
\frac{\log(1/\varepsilon)}
{\log\log(1/\varepsilon)}
\right]^{d+2}
\right)
\right\}.
\]

Thus, the tanh and ReLU constructions have the same asymptotic
parameter complexity, although their respective depth and width
requirements differ.

\subsection{ Parameter Complexity Bound for Kernels on the $p$-Torus with Polynomially Decaying Fourier Coefficients}
\label{app:torus-polynomial-parameter-complexity}
We derive the parameter-complexity bounds reported in Table~\ref{tab:torus-functional-approximation} for kernels on $\mathbb T^p$ with polynomially decaying Fourier coefficients
\[
\theta_j=(1+\|j\|^2)^{-\gamma},
\qquad \gamma>\frac{p+1}{2},
\]
using the approximation estimates in Theorem~\ref{thm:1+j}.

Again, let
$\mathcal P$ denote the total number of network parameters and let
$\varepsilon>0$ be the prescribed approximation accuracy. 

\paragraph{Tanh networks.}
Define
\[
D_{\gamma,p,s}
:=
5p+s(4\gamma-p-1),
\qquad
\alpha
:=
\frac{s^2}{D_{\gamma,p,s}}.
\]
The approximation bound in Theorem~\ref{thm:1+j}(i) can then be written
as
\[
\sup_{f\in\mathcal K}
\left|
F(f)-\widehat G\bigl(f(\bar t)\bigr)
\right|
\leq
C_{\gamma,p,s,F}M^{-\alpha}.
\]
Thus, to make the approximation error at most $\varepsilon$, it is
sufficient to take
\[
M
\geq
\left(
\frac{C_{\gamma,p,s,F}}{\varepsilon}
\right)^{1/\alpha}.
\]
Up to integer rounding and the fixed lower bound on $M$ required by
Theorem~\ref{thm:1+j}(i), we may therefore choose $M$ of the size
\begin{equation}\label{eq:torus-poly-tanh-M}
M
=
\mathcal O\left(
\varepsilon^{-1/\alpha}
\right)
=
\mathcal O\left(
\varepsilon^{-\frac{D_{\gamma,p,s}}{s^2}}
\right).
\end{equation}

We next determine the resulting number of sampling points. Let
$\eta
:=
\frac{s}{
\frac{5p}{2}
+s\left(2\gamma-\frac p2-\frac12\right)
}$.
Because
$\frac{5p}{2}
+s\left(2\gamma-\frac p2-\frac12\right)
=
\frac{D_{\gamma,p,s}}{2}$,
we have
\[
\eta=\frac{2s}{D_{\gamma,p,s}},
\qquad \hbox{and} \qquad
\frac{\eta}{\alpha}=\frac{2}{s}.
\]
The choice $m=\lceil M^\eta\rceil$ consequently gives
$N=m^p
=
\mathcal O\left(M^{\eta p}\right)$.
Substituting \eqref{eq:torus-poly-tanh-M}, we obtain
\begin{equation}\label{eq:torus-poly-tanh-N}
N
=
\mathcal O\left(
\varepsilon^{-\frac{\eta p}{\alpha}}
\right)
=
\mathcal O\left(
\varepsilon^{-\frac{2p}{s}}
\right).
\end{equation}

Let
\[
W_1=N(M-1),
\qquad
W_2=
3\left\lceil\frac{N+1}{2}\right\rceil(5M)^N
\]
be the widths of the two hidden layers. Counting the weights and biases
of a fully connected realization gives
\[
\mathcal P_{\tanh}
\leq
(N+1)W_1+(W_1+1)W_2+(W_2+1).
\]
In particular,
$\mathcal P_{\tanh}
=
\mathcal O\left(
N^2(M+1)(5M)^N
\right)$,
and therefore
\begin{align*}
\log\mathcal P_{\tanh}
=
\mathcal O\left(
N\log(5M)+\log M+\log N
\right)
=
\mathcal O\left(N\log M\right).
\end{align*}
By \eqref{eq:torus-poly-tanh-M} and
\eqref{eq:torus-poly-tanh-N},
\[
N=\mathcal O\left(\varepsilon^{-\frac{2p}{s}}\right)
\quad\text{and}\quad
\log M=\mathcal O\left(\log\frac1\varepsilon\right).
\]
It follows that
\[
\log\mathcal P_{\tanh}
=
\mathcal O\left(
\varepsilon^{-\frac{2p}{s}}
\log\frac1\varepsilon
\right).
\]
Exponentiating both sides yields
$\mathcal P_{\tanh,\varepsilon}
=
\exp\left\{
\mathcal O\left(
\varepsilon^{-\frac{2p}{s}}
\log\frac1\varepsilon
\right)
\right\}$.

\paragraph{ReLU networks.}
Let
\[
q:=\log(ML),
\qquad
\kappa:=\frac{s}{2p},
\qquad
C:=C_{\gamma,p,s,F}.
\]
The approximation bound in Theorem~\ref{thm:1+j}(ii) becomes
\[
\sup_{f\in\mathcal K}
\left|
F(f)-\widehat G\bigl(f(\bar t)\bigr)
\right|
\leq
C\left(\frac{\log q}{q}\right)^\kappa.
\]
To make the right-hand side at most $\varepsilon$, it is sufficient
that
$\frac{\log q}{q}
\leq
\left(\frac{\varepsilon}{C}\right)^{1/\kappa}$.
Equivalently, defining
\[
U_\varepsilon
:=
\left(\frac{C}{\varepsilon}\right)^{1/\kappa}
=
\left(\frac{C}{\varepsilon}\right)^{\frac{2p}{s}},
\]
we require
\[
\frac{q}{\log q}\geq U_\varepsilon.
\]

For $U_\varepsilon$ sufficiently large, take
\[
Q_\varepsilon
:=
2U_\varepsilon\log U_\varepsilon.
\]
Since
\[
\log Q_\varepsilon
=
\log 2+\log U_\varepsilon+\log\log U_\varepsilon
\leq
2\log U_\varepsilon
\]
for all sufficiently large $U_\varepsilon$, it follows that
\[
\frac{Q_\varepsilon}{\log Q_\varepsilon}
\geq
U_\varepsilon.
\]
We may therefore choose, for example,
\[
M=L=
\left\lceil
\exp\left(\frac{Q_\varepsilon}{2}\right)
\right\rceil.
\]
With this choice,
\[
q=\log(ML)
=
\mathcal O\left(
U_\varepsilon\log U_\varepsilon
\right) \qquad \hbox{and} \qquad \frac{q}{\log q}
=
\mathcal O(U_\varepsilon).
\]
Because
$U_\varepsilon
=
\mathcal O\left(
\varepsilon^{-\frac{2p}{s}}
\right)$ and
$\log U_\varepsilon
=
\mathcal O\left(\log\frac1\varepsilon\right)$,
we obtain
\begin{equation}\label{eq:torus-poly-ReLU-q}
q
=
\mathcal O\left(
\varepsilon^{-\frac{2p}{s}}
\log\frac1\varepsilon
\right).
\end{equation}
For sufficiently small $\varepsilon$, this choice also satisfies the
lower bound $ML>e^{A_0}$ required by
Theorem~\ref{thm:1+j}(ii).

We next bound the number of sampling points. From
\[
m
=
\left\lfloor
C_0
\left(
\frac{q}{\log q}
\right)^{1/p}
\right\rfloor,
\]
we have
\[
N=m^p
=
\mathcal O\left(\frac{q}{\log q}\right)
=
\mathcal O(U_\varepsilon)
=
\mathcal O\left(
\varepsilon^{-\frac{2p}{s}}
\right).
\]

Let
\[
D=11L+2N+18
\]
and
\[
W=
3^{N+3}
\max\left\{
N\left\lfloor M^{1/N}\right\rfloor,
M+2
\right\}
\]
denote the depth and maximal width of the ReLU network, respectively.
The total number of weights and biases of a fully connected network
with depth $D$ and maximal width $W$ satisfies
\[
\mathcal P_{\mathrm{ReLU}}
\leq
C_1(D+1)(W+1)^2
\]
for some universal constant $C_1>0$. Moreover,
\[
\log D
=
\mathcal O\left(
\log L+\log(N+1)
\right)
=
\mathcal O(q).
\]
Using
$\left\lfloor M^{1/N}\right\rfloor
\leq M^{1/N}\leq M$
also gives
$\log W
=
\mathcal O\left(
N+\log N+\log(M+2)
\right)=
\mathcal O(N+q)$.
Since
\[
N
=
\mathcal O\left(
\varepsilon^{-\frac{2p}{s}}
\right)
\]
and, by \eqref{eq:torus-poly-ReLU-q},
\[
q
=
\mathcal O\left(
\varepsilon^{-\frac{2p}{s}}
\log\frac1\varepsilon
\right),
\]
we have $N=\mathcal O(q)$ and hence
\[
\log\mathcal P_{\mathrm{ReLU}}
=
\mathcal O(q).
\]
It follows that
\[
\log\mathcal P_{\mathrm{ReLU}}
=
\mathcal O\left(
\varepsilon^{-\frac{2p}{s}}
\log\frac1\varepsilon
\right).
\]
Therefore,
\[
\mathcal P_{\mathrm{ReLU},\varepsilon}
=
\exp\left\{
\mathcal O\left(
\varepsilon^{-\frac{2p}{s}}
\log\frac1\varepsilon
\right)
\right\}.
\]

Thus, the tanh and ReLU constructions require the same asymptotic
number of parameters to attain accuracy $\varepsilon$, although their
depth and width allocations are different.}

\vskip 0.2in

\bibliography{reference}
\end{document}